\documentclass[times, 5p]{elsarticle}
\usepackage{hyperref}
\journal{-}

\usepackage{amsmath}
\usepackage{diagbox}
\usepackage{subfig}
\usepackage{listings} 

\DeclareFixedFont{\ttb}{T1}{txtt}{bx}{n}{8} 
\DeclareFixedFont{\ttm}{T1}{txtt}{m}{n}{8}  

\usepackage{color}
\definecolor{deepblue}{rgb}{0,0,0.5}
\definecolor{deepred}{rgb}{0.6,0,0}
\definecolor{deepgreen}{rgb}{0,0.5,0}

\definecolor{dkgreen}{rgb}{0,0.6,0}
\definecolor{mauve}{rgb}{0.58,0,0.82}
\definecolor{desertpurple}{rgb}{0.525490,0.176470,0.525490}

\newcommand\pythonstyle{\lstset{
language=Python,
basicstyle=\ttm,
numberstyle={\tiny},
otherkeywords={self,as},             
keywordstyle=\ttb\color{deepblue},
commentstyle=\color{mauve},
emph={MyClass,__init__},          
emphstyle=\ttb\color{deepred},    
stringstyle=\color{deepgreen},
frame=tb,                         
showstringspaces=false            %
}}

\lstnewenvironment{python}[1][]
{
\pythonstyle
\lstset{#1}
}
{}

\usepackage{comment}
\newsavebox{\measurebox}

\def \FracTAL {\texttt{FracTAL} }

\newcommand{\ceecnet}{\texttt{CEECNet}}
\newcommand{\mantis}{\texttt{mantis}}

\def \tp {\mathrm{TP}}
\def \tn {\mathrm{TN}}
\def \fp {\mathrm{FP}}
\def \fn {\mathrm{FN}}

\def \tnmt {\mathcal{T}}
\def \ftnmt {\mathcal{FT}}

\hypersetup{
    colorlinks,
    linkcolor={red!50!black},
    citecolor={blue!50!black},
    urlcolor={blue!50!black}
}

\biboptions{authoryear}

\begin{document}

\begin{frontmatter}
\title{Looking for change? Roll the Dice and demand Attention}
\author[ICRAR,data61]{Foivos I. Diakogiannis\fnref{myfootnote1}}
\author[CSIROAF]{Fran\c{c}ois Waldner}
\author[data61]{Peter Caccetta}

\address[ICRAR]{ICRAR, the University of Western Australia}
\address[data61]{Data61, CSIRO, Floreat WA}
\fntext[myfootnote1]{foivos.diakogiannis@uwa.edu.au}
\address[CSIROAF]{CSIRO Agriculture \& Food, St Lucia, QLD, Australia}

\begin{abstract}
Change detection, i.e. identification per pixel of changes for some classes of interest from a set of bi-temporal co-registered
images, is a fundamental task in the field of remote sensing. It remains challenging due to unrelated forms of change that appear
at different times in input images. These are changes due to to different environmental conditions or simply changes of objects
that are not of interest.
%
Here, we propose a reliable deep learning framework for the task of  semantic change detection in very high-resolution aerial images.  Our framework consists of a new loss function,  new attention modules, new feature extraction building blocks, and a new backbone architecture that is tailored for the task of semantic change detection. 
Specifically,  we define a new form of set similarity, that is based on an iterative evaluation of a variant of the Dice coefficient. We use this similarity metric to define a new loss function as well as a new spatial and channel convolution Attention layer (the \FracTAL).  The new attention layer, designed specifically for vision tasks, is memory efficient, thus suitable for use in all levels of deep convolutional networks. 
Based on these, we introduce two new efficient self-contained feature extraction  convolution units.  We term these units  \ceecnet{} and  \FracTAL \texttt{ResNet}  units. We validate  the performance of these feature extraction building blocks on the CIFAR10 reference data  and compare the results with  standard ResNet modules. \textcolor{black}{We also compare the proposed \FracTAL attention layer against the Convolution Block Attention Module (\texttt{CBAM}), showing 1\% performance increase between two otherwise identical networks, that use different attention modules}. 
Further, we introduce a new encoder/decoder scheme, a network \emph{macro}-topology, that is tailored for the task of change detection.   
\textcolor{black}{The key insight in our approach is to facilitate the use of relative attention between two convolution layers in order to compare them.  Our network moves away from any notion of subtraction of feature layers for identifying change.}
We validate our approach by showing excellent performance  and achieving state of the art score (F1 and Intersection over Union - hereafter IoU) on two building change detection datasets, namely, the LEVIRCD (F1: 0.918, IoU: 0.848) and the  WHU (F1: 0.938, IoU: 0.882) datasets. 
\end{abstract}


\end{frontmatter}

\begin{figure*}[h!]
\centering
\includegraphics[clip, trim=7.25cm 4.40cm 4.25cm 4.1cm,width=\textwidth]{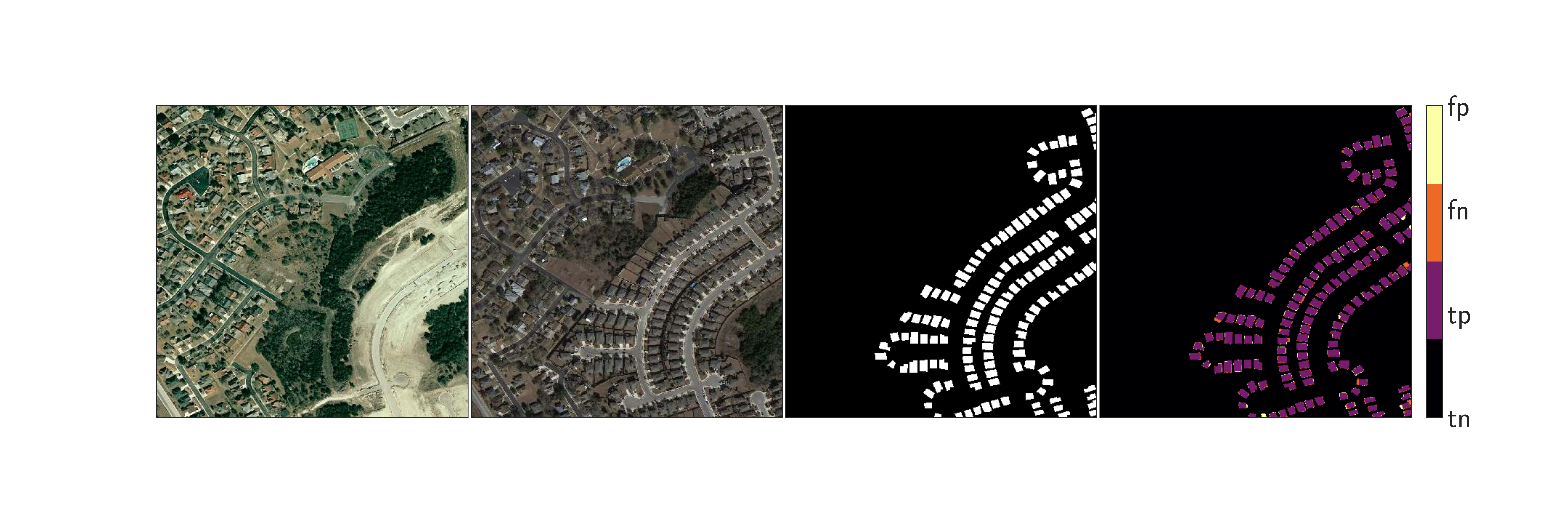}
\caption{Example of the proposed framework (architecture: \mantis{} \ceecnet{}V1) change detection performance on the LEVIRCD test set \citep{rs12101662}. From left to right:  input image at date 1,  input image at date 2, ground truth buildings change mask, and color coded the  true negative (\texttt{tn}), true positive (\texttt{tp}), false positive (\texttt{fp}) and false negative (\texttt{fn}) predictions.} 
\label{cover}
\end{figure*}

\section{Introduction}

Change detection is one of the core applications of remote sensing. 
The goal of change detection is to assign binary labels (``change'' or no ``change'') to every pixel in a study area based on at least two co-registered images taken at different times. The definition of ``change'' varies across applications and includes, for instance, urban expansion~\citep{rs12101662}, flood mapping~\citep{giustarini2012change}, deforestation~\citep{morton2005rapid}, and cropland abandonment~\citep{low2018mapping}. Changes of multiple land-cover classes, i.e. semantic change detection, can also be addressed  simultaneously~\citep{daudt2019multitask}. It remains a challenging task due to various forms of change owed to varying environmental conditions that do not constitute a change for the objects of interest \citep{Varghese_2018_ECCV_Workshops}.  

A plethora of change-detection algorithms has been devised and summarised in several reviews~\citep{lu2004change, coppin2004review, hussain2013change, tewkesbury2015critical}. In recent years, computer vision has further pushed the state of the art, especially in applications where the spatial context is paramount. The rise of computer vision, especially deep learning, is related to advances and democratisation of powerful computing systems, increasing amounts of available data, and the development of innovative ways to exploit data~\citep{daudt2019multitask}. 

Our starting point is  the hypothesis that human intelligence identifies differences in images by looking for change in objects of interest at a higher cognitive level \citep{Varghese_2018_ECCV_Workshops}. We understand this because the time required for identifying objects that changed between two images, increases  with time when the number of changed objects increases \citep{TREISMAN198097}. That is, there is strong correlation between processing time and number of individual objects that changed. In other words,  the higher the complexity of the changes the more time is required to accomplish it.  
Therefore, simply subtracting extracted features from images (which is a constant time operation) cannot account for the complexities of human perception. As a result, the deep convolutional neural networks proposed in this paper address change detection without using bespoke features subtraction.

In this work, we developed neural networks using attention mechanisms that emphasize areas of interest in two bi-temporal coregistered aerial images.
It is the network that learns what to emphasize, and how to extract features that describe change at a higher level. To this end, we propose a dual encoder -- single decoder scheme, that fuses information of corresponding layers with relative attention and extracts as a final layer  a segmentation mask. This mask designates change for classes of interest, and can also  be used for the dual problem of class attribution of change. As in previous work, we facilitate the use of conditioned multi-tasking\footnote{The algorithm predicts first the distance transform of the change mask, then it reuses this information and identifies the boundaries of the change mask, and, finally, re-uses both distance transform and boundaries to estimate the change segmentation mask.} \citep{DIAKOGIANNIS202094} that proves crucial for stabilizing the training process and improving performance. 
In summary, the main  contributions of this work are:
\begin{enumerate}
\item
We introduce a new set similarity metric that is a variant
of the Dice coefficient, the Fractal Tanimoto similarity measure (section \ref{section_fractal_tanimoto_similarity}). This similarity measure has the advantage
that it can be made steeper than the standard Tanimoto metric towards optimality, thus providing a finer-grained similarity metric between layers.  The level of steepness is controlled from a
depth recursion hyper-parameter. It can be used both as
a ``sharp'' loss function when fine-tuning a model at the
latest stages of training, as well as a set similarity metric
between feature layers in the attention mechanism.
\item Using the above set similarity as a loss function, we propose an evolving loss strategy for fine-tuning training of neural networks (section \ref{section_loss}). This strategy helps to avoid overfitting and improves performance.
\item We introduce the Fractal Tanimoto Attention Layer (hereafter \FracTAL), tailored for vision tasks (section \ref{section_fractal_tan_attention}). This layer uses the fractal Tanimoto similarity to compare queries with keys inside the Attention module. It is a form of spatial and channel attention combined. 
\item We introduce a feature extraction building block that is based on the Residual neural network and fractal Tanimoto Attention (section \ref{fractal_resnet}). The new \FracTAL \texttt{ResNet} converges faster to optimality than standard residual networks and enhances performance. 
\item We introduce two variants of a new feature extraction building block, the Compress-Expand / Expand-Compress unit (hereafter \ceecnet{} unit - section \ref{micro_ceecnet_ref}). This unit exhibits enhanced performance in comparison with standard residual units, and the \FracTAL \texttt{ResNet} unit.  
\item  Capitalizing on these findings,  we introduce  a new backbone encoder/decoder scheme, a \emph{macro}-topology - the   \texttt{mantis} - that is tailored for the task of change detection (section \ref{the_mantis_section}). The encoder part is a Siamese dual encoder, where the corresponding extracted features at each depth are fused together with  \FracTAL \textcolor{black}{relative} attention. In this way, information exchange between features extracted from bi-temporal images is enforced. There is no need for manual feature subtraction.  
\item Given the relative fusion operation between the encoder features at different levels, our algorithm achieves state of the art performance on the LEVIRCD and WHU datasets without requiring the use of contrastive loss learning during training (section \ref{section_results}). Therefore,  it is easier to implement with standard deep learning libraries and tools. 
\end{enumerate}
Networks integrating the above-mentioned contributions yielded state of the art performance for the task of building change detection in two
benchmark data sets for change detection: the WHU \citep{Ji2019FullyCN} and LEVIRCD \citep{rs12101662} datasets. 

In addition to the previously mentioned sections, the following  complete the works. In Section \ref{related_work} we present related work on Attention mechanism and change detection, specialised for the case of very high resolution (hereafter VHR) aerial images. In Section \ref{section_experimental_design} we describe the setup of our experiments.  In Section \ref{section_ablation_study} we perform an ablation study of the proposed schemes.   Finally, in  Section \ref{section_algorithms} we present in \textsc{mxnet/gluon} style pseudocode various key elements of our architecture\footnote{A software implementation of  the  models that relate to this work can be found on \href{https://github.com/feevos/ceecnet}{https://github.com/feevos/ceecnet}.}.

\section{Related Work}
\label{related_work}

\subsection{On attention}

The attention mechanism was first introduced by \cite{bahdanau2014neural} for the task of neural machine translation\footnote{That is, language to language translation of sentences, e.g. English to French.} (hereafter NMT).  This mechanism addressed the problem of translating very long sentences in encoder/decoder architectures. An encoder is a neural network that encodes a phrase to a fixed-length vector. Then the decoder operates on this output and produces a translated phrase (of variable length). It was observed that these types of architectures were not performing well when the input sentences were very long \citep{DBLP:journals/corr/ChoMBB14}.    
The attention mechanism provided a solution to this problem: instead of using all the elements of the encoder vector on equal footing for the decoder, the attention provided a weighted view of them. That is, it emphasized  the locations of encoder features that were more important than others for the translation, or stated another way,  it emphasized some input words that were more important for the meaning of the phrase. However, in NMT, the location of the translated words is not in direct correspondence with the input phrase, because of the syntax changes.  Therefore, \cite{bahdanau2014neural} introduced a relative alignment vector, $e_{ij}$, that was responsible for encoding the location dependences: in language, it is not only the meaning (value) of a word that is important but also its relative location in a particular syntax.   Hence, the attention mechanism that was devised  was comparing the emphasis of inputs at location $i$ with respect to output words at locations $j$.  
Later, \cite{DBLP:journals/corr/VaswaniSPUJGKP17} developed further this mechanism and introduced the scaled dot product self-attention mechanism as a fundamental constituent of their Transformer architecture. This allowed the dot product to be used as a similarity measure  between feature layers, including  feature vectors having large dimensionality. 

The idea of using attention for vision tasks soon passed to the community. \cite{DBLP:journals/corr/abs-1709-01507} introduced channel based attention, in their squeeze and excitation architecture. \cite{DBLP:journals/corr/abs-1711-07971} used spatial attention to facilitate non-local relationships across sequences of images. \cite{DBLP:journals/corr/ChenZXNSC16} combined both approaches by introducing joint spatial and channel wise attention in convolutional neural networks, demonstrating improved performance on image captioning datasets.   
\textcolor{black}{
\cite{10.1007/978-3-030-01234-2_1} introduced the 
 Convolution Block Attention Module (\texttt{CBAM}) which is also a form of spatial and channel attention, and showed improved performance on image classification and object detection tasks.}
 To the best of our knowledge, the most faithful implementation  of multi-head  attention \citep{DBLP:journals/corr/VaswaniSPUJGKP17} for convolution layers, is \cite{DBLP:journals/corr/abs-1904-09925} (spatial attention).

\subsection{On change detection} 

\cite{Sakurada2015ChangeDF} and \cite{Stent-RSS-16} (see also \citealt{DBLP:journals/corr/abs-1810-09111})  were some of the first to introduce fully convolutional networks for the task of scene change detection in computer vision, and they both introduced street view change detection datasets.  \cite{Sakurada2015ChangeDF} extracted features from a convolutional neural networks and combined them with super pixel segmentation to recover change labels in the original resolution. \cite{Stent-RSS-16} proposed an approach that chains  multi-sensor fusion simultaneous localization and mapping (SLAM) 
with a fast 3D reconstruction pipeline that provides coarsely registered image pairs to an encoder/decoder convolutional network. The output of their algorithm is a pixel-wise change detection binary mask. 
   
Researchers in the field of remote sensing 
 picked up and evolved this  knowledge and started using it for the task of land cover change detection. 
In the remote sensing community, the dual Siamese encoder and a single decoder is frequently adopted. The majority of different approaches then modifies how the different features extracted from the dual encoder are consumed (or compared) in order to produce a change detection prediction layer. 
 In the following we focus on approaches that follow this paradigm and are most relevant to our work. For a general overview of land cover change detection in the field of remote sensing interested readers can consult \cite{HUSSAIN201391} and \cite{6345345gdfGHTff}. For a  general review  on AI applications of change detection to the field of remote sensing \cite{Shi_2020}.  
 
 \cite{CAYEDAUDT2019102783} presented and evaluated various strategies for land cover change detection, establishing that their best algorithm was a joint  multitasking segmentation and change detection approach. That is, their algorithm predicted simultaneously the semantic  classes on each input image, as well as the binary mask of change between the two.

\begin{figure*}
\centering
\includegraphics[clip, trim=2.25cm 3.40cm 2.25cm 4.1cm,width=0.99\textwidth]{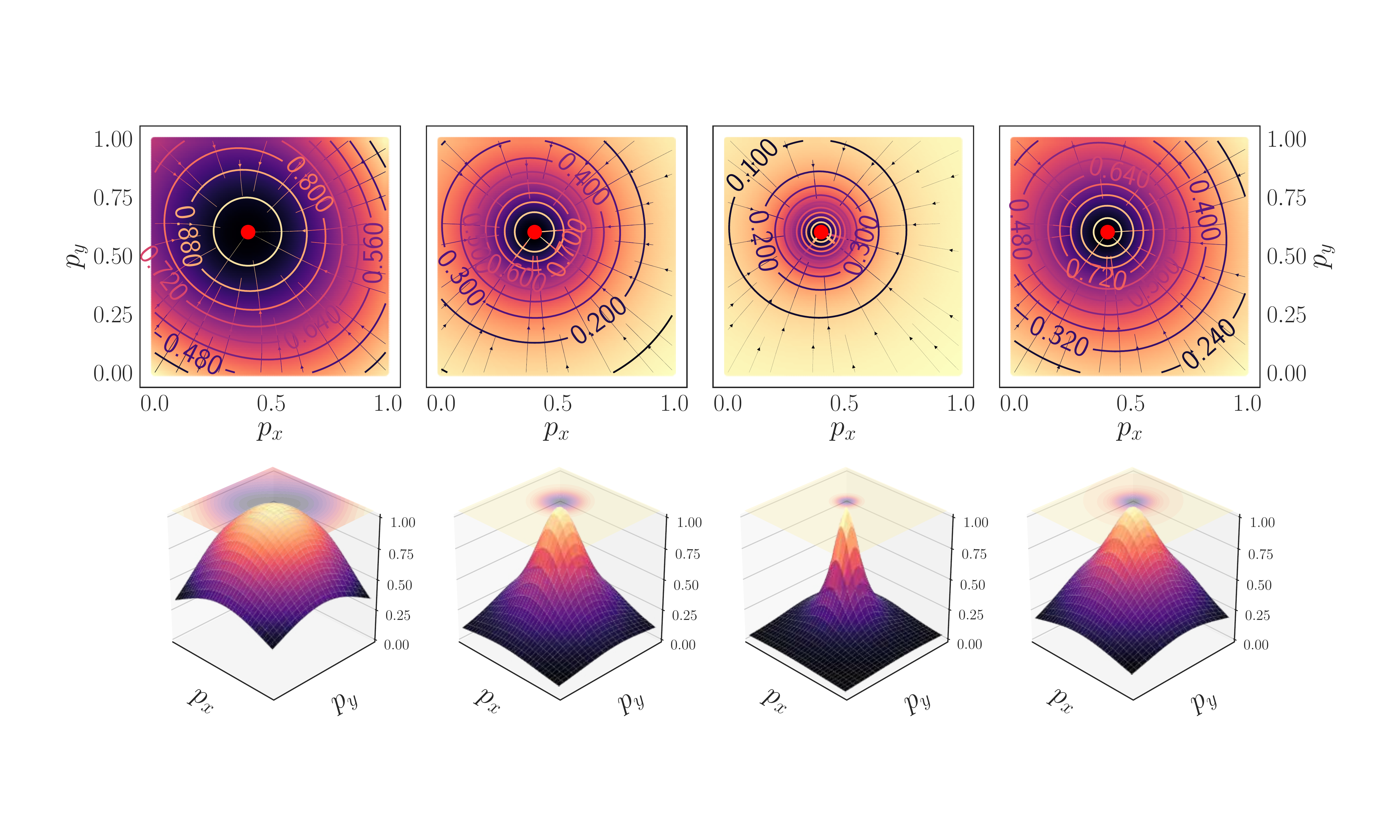}
\caption{Fractal Tanimoto similarity measure. In the top row we plot the two dimensional density maps for the $\ftnmt$ similarity coefficient. 
From left to right  the depths are $d\in \{0, 3, 5\}$. The last column corresponds to the average of values up to depth $d=5$, i.e. $\langle \ftnmt \rangle^{5} = (1/5)\sum_d \ftnmt^d$. In the bottom figure we represent in 3D the same values. The horizontal contour plot at $z=1$ corresponds to the Laplacian of the $\ftnmt$. It is observed that as the depth, $d$, of the iteration increases, the function becomes steeper towards optimality. } 
\label{FracTanmt2D3D}
\end{figure*}

For the task of buildings change detection,  
\cite{rs11111343} presented a methodology that is a two-stage process, wherein the first part they use a building extraction algorithm from single date input images. In the second part, the binary masks that are extracted are concatenated together and inserted into a different network that is responsible for identifying changes between the two binary layers. In order to evaluate the impact of the quality of the building extraction networks, the authors use two different architectures. The first, one of the most successful networks to date for instance segmentation, the Mask-RCNN \citep{DBLP:journals/corr/HeGDG17}, and the second the MS-FCN (multi scale fully convolutional network) that is based on the original UNet architecture \citep{DBLP:journals/corr/RonnebergerFB15}. The advantage of this approach, according to the authors is the fact that they could use unlimited synthetic data for training the second stage of the algorithm.  

\cite{chen2020dasnet} used a dual attentive convolutional neural network, i.e. the feature extractor was a siamese VGG16 pre-trained network. The attention module they used for vision, was both spatial and channel attention, and it was the one introduced in \cite{DBLP:journals/corr/VaswaniSPUJGKP17}, however with a single head.  Training was performed with a contrastive loss function. 

\cite{rs12101662} presented the STANet, which consists of a feature extractor based on ResNet18 \citep{DBLP:journals/corr/HeZRS15}, and two versions of spatio-temporal attention modules, the Basic spatial-temporal attention module (BAM) and the pyramid spatial-temporal attention module (PAM). The authors introduced the LEVIRCD change detection dataset and demonstrated excellent performance. Their training process facilitates a contrastive loss applied at the feature pixel level. Their algorithm predicts binary change labels.  

\cite{rs12030484} introduced the PGA-SiamNet that uses a dual Siamese encoder that extracts features from the two input networks. They used VGG16 for feature extraction. A key ingredient to their algorithm is the co-attention module \citep{Lu_2019_CVPR} that was initially developed for video object segmentation. The authors use it for fusing  the extracted features of each input image from the dual VGG16 encoder.

\section{Fractal Tanimoto similarity coefficient}
\label{section_fractal_tanimoto_similarity}

In \citep{DIAKOGIANNIS202094} we analyzed the performance of the various flavours of the Dice coefficient and introduced the Tanimoto with complement coefficient. Here, we expand further our analysis, and we present a new  functional form for this similarity metric. We use it both as a  
self-similarity measure between convolution layers in a new attention module, as well as a loss function for finetuning semantic segmentation models.

For two (fuzzy) binary vectors of equal dimension, $\mathbf{p}$, $\mathbf{l}$, whose elements lie in the range $[0,1]$  the Tanimoto similarity coefficient is defined:
\begin{equation}
T(\mathbf{p},\mathbf{l}) = \frac{\mathbf{p}\cdot \mathbf{l}}{\mathbf{p}^2 +  \mathbf{l}^2 - \mathbf{p}\cdot \mathbf{l}}
\end{equation}
 Interestingly, the dot product  between two fuzzy binary vectors  is another similarity measure of their agreement. This inspired us to introduce an iterative functional form of the Tanimoto:
\begin{align}
\label{Tnmt_fractal_def}
\tnmt^0 & \equiv T(\mathbf{p},\mathbf{l}) = \frac{\mathbf{p}\cdot \mathbf{l}}{\mathbf{p}^2 +  \mathbf{l}^2 - \mathbf{p}\cdot \mathbf{l}}\\
\tnmt^d &= \frac{\tnmt^{d-1}(\mathbf{p}, \mathbf{l})}{\tnmt^{d-1}(\mathbf{p},\mathbf{p})+  \tnmt^{d-1}(\mathbf{l},\mathbf{l}) - \tnmt^{d-1}(\mathbf{p},\mathbf{l})}
\end{align}
For example, expanding Eq. \eqref{Tnmt_fractal_def} for $d=2$, yields: 
\begin{equation}
\tnmt^2(\mathbf{p},\mathbf{l})=
\frac{\mathbf{p} \cdot \mathbf{l}}{(\mathbf{l}^2-\mathbf{p}\cdot \mathbf{l}+\mathbf{p}^2) \left(2-\frac{\mathbf{p}\cdot\mathbf{l}}{\mathbf{l}^2-\mathbf{p}\cdot\mathbf{l}+\mathbf{p}^2}\right) 
\left(2-\frac{\mathbf{p}\cdot \mathbf{l}}{(\mathbf{l}^2-\mathbf{p}.\mathbf{l}+\mathbf{p}^2) \left(2-\frac{\mathbf{p}\cdot \mathbf{l}}{\mathbf{l}^2-\mathbf{p}\cdot\mathbf{l}+\mathbf{p}^2}\right)}\right)}
\end{equation}
We can expand this for an arbitrary depth $d$ and then we get the following simplified version\footnote{The simplified formula was obtained with \textsc{Mathematica 11} Software.} of the fractal Tanimoto similarity measure:
\begin{equation}
\mathcal{T}^d(\mathbf{p},\mathbf{l}) = \frac{\mathbf{p}\cdot \mathbf{l}}{2^d(\mathbf{p}^2+\mathbf{l}^2)- (2^{d+1}-1) \mathbf{p}\cdot \mathbf{l}}
\end{equation}
This function takes values in the range $[0,1]$ and it becomes steeper as $d$ increases. At the limit $d\to \infty$ it behaves like the integral of the  Dirac $\delta$ function around point $\mathbf{l}$, $\int\delta (\mathbf{p} - \mathbf{l}) d\mathbf{p}$.  That is, the parameter $d$ is a form of annealing ``temperature''. Interestingly, although the iterative scheme was defined with $d$ being an integer, for continuous values $d \geq 0$, $\mathcal{T}^d$ remains bounded in the interval $[0,1]$. That is:   
\begin{equation}
\mathcal{T}^d: \Re^n\times \Re^n \to U \subseteq [0,1] 
\end{equation}
where $n=\dim (\mathbf{p})$ is the dimensionality of the fuzzy binary vectors $\mathbf{p}$, $\mathbf{l}$.

In the following we will use the functional form of the fractal Tanimoto with complement \citep{DIAKOGIANNIS202094}, i.e.:
\begin{equation}
\ftnmt^d(\mathbf{p},\mathbf{l}) \equiv \frac{\mathcal{T}^d(\mathbf{p},\mathbf{l}) +  \mathcal{T}^d(\mathbf{1}-\mathbf{p},\mathbf{1}-\mathbf{l})}{2}
\end{equation}

In Fig. \ref{FracTanmt2D3D} we provide a simple example for a ground truth vector $\mathbf{l} = \{0.4, 0.6 \}$ and a continuous vector of probabilities $\mathbf{p} = \{p_x, p_y\}$. On the top panel, we construct density plots of the Fractal Tanimoto function with complement, $\ftnmt^d$. Oveplotted are the gradient field lines that point to the ground truth. In the bottom pannels,  we plot the corresponding 3D representations.  From left to right, the first column corresponds to $d=0$, the second to $d=3$ and the third to $d=5$. It is apparent that the effect of the $d$ hyperparameter is to make the similarity metric steeper towards the ground truth. 
For all practical purposes (network architecture, evolving loss function) we use the average fractal Tanimoto loss (last column), due to having steeper gradients away from optimality\footnote{\textcolor{black}{The formula is valid for $d>=1$, for $d=0$ it reverts to the standard fractal Tanimoto loss, $\ftnmt^{d=0}(\mathbf{p},\mathbf{l})$.}}: 
\begin{equation}
\label{ftnmt_average}
\langle \ftnmt  \rangle^d(\mathbf{p},\mathbf{l})  \equiv \frac{1}{d} \sum_{i=0}^{d-1} \ftnmt^i(\mathbf{p},\mathbf{l})
\end{equation}

\section{Evolving loss strategy}
\label{section_loss}

In this section, we describe a training strategy that modifies the depth of the fractal Tanimoto similarity coefficient, when used as a loss function, on each learning rate reduction. For minimization problems, the fractal Tanimoto loss is defined through:   
$L = 1 - \langle \mathcal{FT}\rangle^d$. In the following, when we refer to the fractal Tanimoto loss function, it should be understood that this is defined trough the similarity coefficient, as described above. 

\begin{figure}
\centering
\includegraphics[clip, trim=2.25cm 3.40cm 2.25cm 4.1cm,width=0.99\columnwidth]{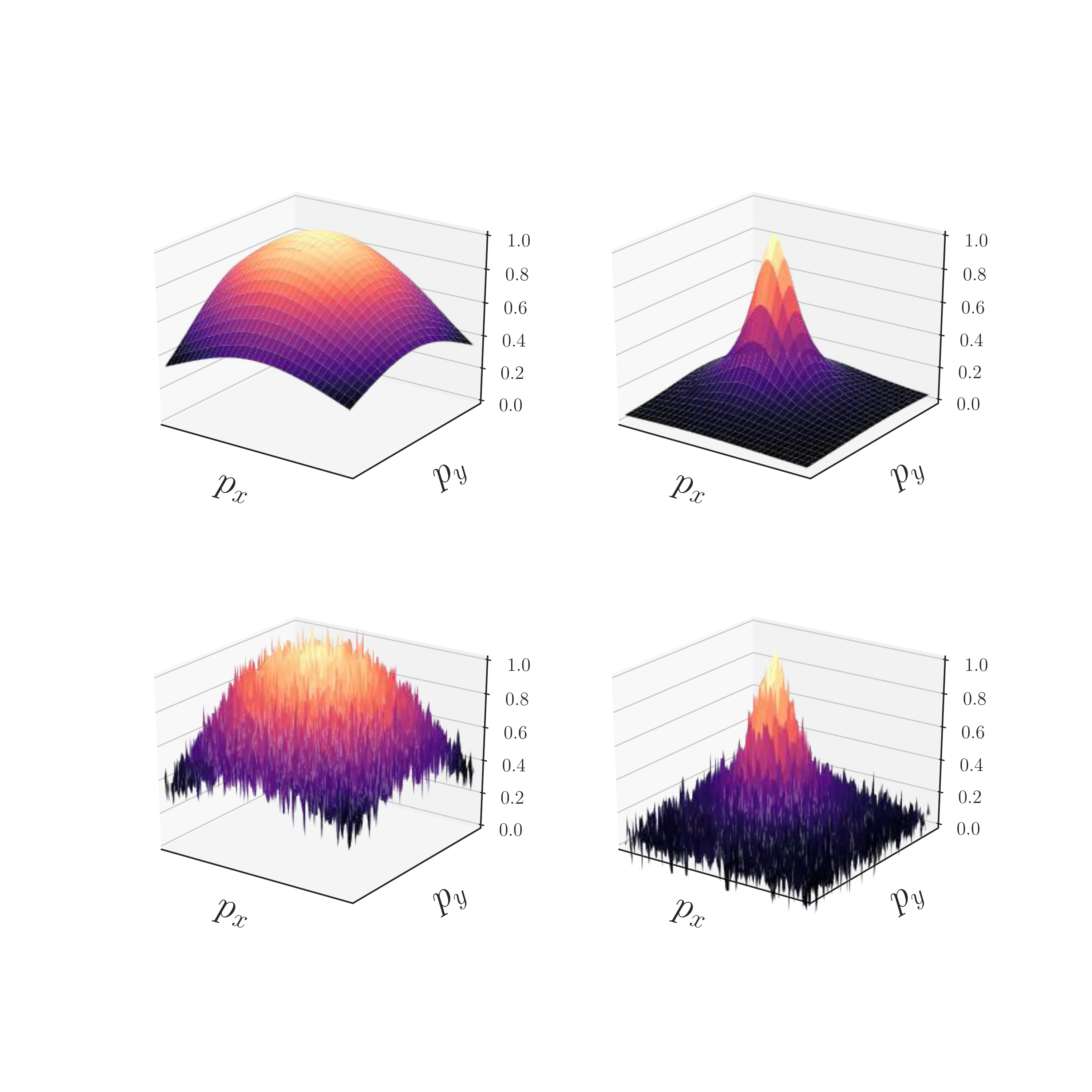}
\caption{Fractal Tanimoto similarity measure with noise. On the top row, from left to right is the $\ftnmt^{\textcolor{black}{0}}(\mathbf{p},\mathbf{l})$ and $(1/10)\sum_{d=0}^{9}(\ftnmt^d(\mathbf{p},\mathbf{l}))$. 
The bottom row is the same corresponding $\ftnmt^d(\mathbf{p},\mathbf{l})$ similarity measures, with Gaussian random noise added. When the algorithmic training approaches optimality with the standard Tanimoto, local noise gradients tend to dominate over the background average gradient. Increasing the slope of the background gradient at later stages of training is a remedy to this problem.}
\label{FracTanmtLoss3D_wthNoise}
\end{figure}

During training, and until the first learning rate reduction we use the standard Tanimoto with complement 
$\ftnmt^0(\mathbf{p},\mathbf{l})$. The reason for this is that for a random initialization of the weights (i.e. for an initial prediction point in the space of probabilities away from optimality), the gradients are steeper towards the best values for this particular loss function (in fact, for cross entropy are even steeper). This can be seen in Fig. \ref{FracTanmt2D3D} in the bottom row: clearly for an initial probability vector $\mathbf{p}=\{p_x,p_y\}$ away from the ground truth $\mathbf{l}=\{0.4,0.6 \}$ the gradients are steeper for $d=0$. As training evolves, and the value of the weights approaches optimality, the predictions approach the ground truth and the loss function flattens out. With batch gradient descent (and variants), we are not really calculating the true (global) loss function, but a noisy approximate version of it. This is because in each batch loss evaluation, we are not using all of the data for the gradients evaluation. 
In Fig. \ref{FracTanmtLoss3D_wthNoise} we represent a graphical representation of the true landscape and a noisy version of it for a toy 2D problem.  In the top row, we plot the value of the $\ftnmt^0$ similarity as well as the average value of the loss functions for $d=0,\ldots,9$ for the ground truth vector $\mathbf{l}=\{0.4,0.6 \}$. In the corresponding bottom rows, we have the same plot were  we also added random Gaussian noise. In the initial phases of training, the average gradients are greater than the local values dues to noise. As the network reaches optimality the average gradient towards optimality becomes smaller and smaller, and the gradients due to noise dominate the training. Once we reduce the learning rate, the step the optimizer takes is even smaller, therefore it cannot easily escape local optima (due to noise). What we propose is to ``shift gears'': once training stagnates, we change the loss function to a similar but steeper one towards optimality that can provide gradients (on average) that can dominate the noise.  Our choice during training is the following set of learning rates and depths of the fractal Tanimoto loss: 
$\{ (\texttt{lr}:10^{-3},d=0),\; (\texttt{lr}:10^{-4},d=10),\; (\texttt{lr}:10^{-5},d=20) \}$. In all evaluations of loss functions for $d>0$, we use the average value for all $d$ values (Eq.  \ref{ftnmt_average}).

\section{Fractal Tanimoto Attention}
\label{section_fractal_tan_attention}

Here, we present a novel convolutional attention layer based on the new similarity metric and a methodology of fusing  information from the output of the attention layer to features extracted from convolutions.

\subsection{Fractal Tanimoto Attention layer }

In the pioneering work of \cite{DBLP:journals/corr/VaswaniSPUJGKP17} the attention operator is defined through a scaled dot product operation. For images in particular, i.e. two dimensional features,  assuming that $\mathbf{q}\in \mathfrak{R}^{ C_q \times H \times W} $ is the query, $\mathbf{k} \in \mathfrak{R}^{ C \times H \times W}$ the key and $\mathbf{v} \in \mathfrak{R}^{ C \times H \times W}$ it's corresponding value, the (spatial) attention is defined as (see also \citealt{zhang2020dive}):
\begin{align}
\mathbf{o} &= \text{softmax} \left(\frac{\mathbf{q}\circ_1 \mathbf{k}}{\sqrt{d}} \right), 
& & \in\mathfrak{R}^{C_q \times C}\\
\text{Att}(\mathbf{q},\mathbf{k},\mathbf{v}) &= \mathbf{o}  \circ_2 \mathbf{v},
& &\in\mathfrak{R}^{C_q \times H \times W}
\end{align}  
Here $d$ is the dimension of the keys and the softmax operation is with respect to the first (channel) dimension. The term $\sqrt{d}$ is a scaling factor that ensures the Attention layer scales well even with a large number of dimensions  \citep{DBLP:journals/corr/VaswaniSPUJGKP17}. The operator $\circ_1$ corresponds to inner product with respect to  the spatial dimensions height, $H$, and width, $W$, while $\circ_2$ is a dot product with respect to channel dimensions\footnote{This is more apparent in index notation: 
\begin{align}
\label{attention_dot_product_traditional}
\mathbf{q} \circ_1 \mathbf{k} &\equiv  \sum_{\textcolor{magenta}{j}\textcolor{cyan}{k}} q_{i\textcolor{magenta}{j}\textcolor{cyan}{k}} k_{l\textcolor{magenta}{j}\textcolor{cyan}{k}} \in \mathfrak{R}^{C_q\times C}\\
\notag
\mathbf{o} & \equiv\text{softmax}\left(\mathbf{q} \circ_1 \mathbf{k} \right) \equiv  o_{il} \\
\notag
\text{Att}(\mathbf{q},\mathbf{k},\mathbf{v}) &\equiv  \text{Att}_{ikj} \equiv \mathbf{o} \circ_2 \mathbf{v} \equiv \sum_{\textcolor{magenta}{l}} o_{i\textcolor{magenta}{l}} v_{\textcolor{magenta}{l}kj} \in \mathfrak{R}^{C_q\times H \times W}
\end{align}}.
In this formalism each channel of the query features is compared with each of the channels of the key values. In addition there is a $1-1$ correspondence between keys and values, meaning that for each key corresponds a unique value.  
The point of the dot product is to emphasize the key-value pairs that are more relevant for the particular query. That is the dot product selects the keys  that are most similar to the particular query. It represents the projection of queries on the keys space. The softmax operator provides a weighted ``view'' of all the values for a particular set of queries, keys and values- or else a ``soft'' attention mechanism. In the multi-head attention paradigm, multiple attention heads that follow the principles described above are concatenated together. One of the key disadvantages of this formulation when used in vision tasks (i.e. two dimensional features) is the very large memory footprint that this layer exhibits. For 1D problems, such as Natural Language Processing, this is not - in general - an issue.

Here we follow a different approach.  We develop our formalism for the case where the number of query channels, $C_q$ is identical to the number of key channels, $C$. However, if desired, our formalism can work for the general case where $C_q \neq C$. 

Let $\mathbf{q} \in \mathfrak{R}^{C\times H \times W}$ be the query features, $\mathbf{k} \in \mathfrak{R}^{C\times H \times W}$ the keys and $\mathbf{v} \in \mathfrak{R}^{C\times H \times W}$ the values. In our formalism, it is a requirement for these operators to have values in $[0,1]$\footnote{This can be easily achieved by applying the sigmoid operator.}. Our approach is a joint spatial and channel attention mechanism. With the use of the Fractal Tanimoto similarity coefficient, we define the \emph{spatial}, $\boxtimes$, and \emph{channel}, $\boxdot$, similarity between the query, $\mathbf{q}$, and key, $\mathbf{k}$, features according to:
\begin{align}
\label{spatial_similarity}
\mathcal{T}^d_{\boxtimes}(\mathbf{q},\mathbf{k}) = \frac{\mathbf{q} \boxtimes \mathbf{k}}{2^d\left( \mathbf{q} \boxtimes \mathbf{q} + \mathbf{k} \boxtimes \mathbf{k} \right) - (2^{d+1}-1) \mathbf{q} \boxtimes \mathbf{k}} \in \mathfrak{R}^{C} \\ 
\label{channel_similarity}
\mathcal{T}^d_{\boxdot}(\mathbf{q},\mathbf{k}) = \frac{\mathbf{q} \boxdot \mathbf{k}}{2^d\left( \mathbf{q} \boxdot \mathbf{q} + \mathbf{k} \boxdot \mathbf{k} \right) - (2^{d+1}-1) \mathbf{q} \boxdot \mathbf{k}}
\in \mathfrak{R}^{H\times W}
\end{align}
where the spatial and channel products are defined as: 
\begin{align*}
\mathbf{q} \boxtimes \mathbf{k} &= \sum_{\textcolor{magenta}{j} \textcolor{cyan}{k}} q_{i\textcolor{magenta}{j}\textcolor{cyan}{k}} k_{i \textcolor{magenta}{j}  \textcolor{cyan}{k}} 
\in \mathfrak{R}^{C}\\ 
\mathbf{q} \boxdot \mathbf{k} &= \sum_{\textcolor{magenta}{i}} q_{\textcolor{magenta}{i}jk}k_{\textcolor{magenta}{i}jk} \in \mathfrak{R}^{H\times W}
\end{align*}
It is important to note that the output of these operators lies numerically within the range [0,1], where 1 indicates identical similarity and 0 indicates no correlation between the query and key. That is, there is no need for normalization or scaling as is the case for the traditional dot product similarity. 

In our approach the spatial and channel attention layers are  defined with element-wise multiplication\footnote{We use \textsc{python numpy} \citep{oliphant2006guide} semantics of broadcasting since the dimensions do not match.} (denoted by the symbol $\odot$): 
\begin{align*}
\text{Att}_{\boxtimes}(\mathbf{q},\mathbf{k},\mathbf{v}) &= \mathcal{T}^d_{\boxtimes}(\mathbf{q},\mathbf{k}) \odot \mathbf{v}\\
\text{Att}_{\boxdot}(\mathbf{q},\mathbf{k},\mathbf{v}) &= \mathcal{T}^d_{\boxdot}(\mathbf{q},\mathbf{k}) \odot \mathbf{v}
\end{align*}
It should be stressed that these operations do not consider that there is a $1-1$ mapping between keys and values. Instead, we consider a map of one-to-many, that is a single key can correspond to a set of values. Therefore, there is no need to use a softmax activation (see also \citealt{DBLP:journals/corr/KimDHR17}). The overall attention is defined as the average of the sum of these two operators: 
\begin{equation}
\text{Att}(\mathbf{q},\mathbf{k},\mathbf{v}) = 0.5(\text{Att}_{\boxtimes} + \text{Att}_{\boxdot})
\end{equation}
In practice we use the averaged fractal Tanimoto similarity coefficient with complement, $\langle \ftnmt \rangle^d_{\boxtimes/\boxdot}$,  both for spatial and channel wise attention.  

As stated previously, it is possible to extend the definitions of spatial and channel products in a way where we compare each of the channels (respectively, spatial pixels) of the query with each of the channels (respectively, spatial pixels) of the key. However, this imposes a heavy memory footprint, and makes deeper models, even for modern-day GPUs, prohibitive\footnote{\textcolor{black}{This will be made clear with an example of \FracTAL spatial similarity vs the dot product similarity that appears in SAGAN \citep{zhang2018selfattention} self-attention.  Let us assume that we have an input  feature layer of size $(B\times C\times H\times W)  = 32 \times  1024 \times 16 \times 16$ (e.g. this appears in the layer at depth 6 of UNet-like architectures, starting from 32 initial features). From this, three layers are produced of the same dimensionalty, the query, the key and value. With the  Fractal Tanimoto spatial similarity, $\mathcal{T}_{\boxtimes}$, the output of the similarity of query and keys is $B \times C\times 1\times 1 =  32\times 1024\times 1\times 1$ (Equation \ref{spatial_similarity}). The corresponding output of the dot similarity of spatial compoments in the self-attention is BxCxC → 32x1024x1024 (Equation \ref{attention_dot_product_traditional}), having $C$-times higher memory footprint. }}. In addition, we found that this approach did not improve performance  for the case of change detection and classification.  Indeed, one needs to question this for vision tasks: the initial definition of attention \citep{bahdanau2014neural} introduced a relative alignment vector, $e_{ij}$, that was necessary because, for the task of NMT, the syntax of phrases changes from one language to the other. That is, the relative emphasis with respect to  location between two vectors is meaningful.  When we compare two images (features) at the same depth of a network (created by two different inputs, as is the case for change detection), we anticipate that the channels (or spatial pixels) will be in correspondence. For example, the RGB (or hyperspectral) order of inputs, does not change. 
That is,  in vision, the situation can be different than NLP because we do not have a relative location change as it happens with words in phrases.

We propose the use of the Fractal Tanimoto Attention Layer (hereafter \FracTAL) for vision tasks as an improvement over the scaled dot product attention mechanism  \citep{DBLP:journals/corr/VaswaniSPUJGKP17}  for the following reasons:
\begin{enumerate}
\item The $\ftnmt$ similarity is automatically scaled in the region $[0,1]$, therefore it does not require normalization, or activation to be applied. This simplifies  the design and implementation of Attention layers and enables training without ad-hoc normalization operations.   
\item The dot product does not have an upper or lower bound, therefore a positive value cannot be a quantified measure of similarity. In contrast $\ftnmt$ has a bounded range of values in $[0,1]$. The lowest value indicates no correlation, and the maximum value perfect similarity. It is thus easier to interpret.  
\item Iteration $d$ is a form of hyperparameter, like ``temperature'' in annealing.  Therefore, the $\ftnmt$ can become as steep as we desire (by modification of the temperature parameter $d$), steeper than the dot product similarity. This can translate to finer query and key similarity.
\item Finally, it is efficient in terms of GPU memory footprint (when one considers that it does both channel and spatial attention), thus allowing the design of more complex convolution building blocks.  
\end{enumerate}
The implementation of the \FracTAL is given in Listing \ref{FTAttentionCODE}. The multihead attention is achieved using group convolutions for the evaluation of queries, keys and values. 

\subsection{Attention fusion}

A critical part in the design of convolution building blocks enhanced with attention is the way the information from attention is passed to convolution layers. To this aim
we propose fusion methodologies of feature layers with the \texttt{FracTAL} for two cases:  self attention fusion, and a relative attention fusion where information from two layers are combined. 

\subsubsection{Self attention fusion}
\label{fractal_resnet}

\begin{figure}
\centering
\includegraphics[width=0.8\columnwidth]{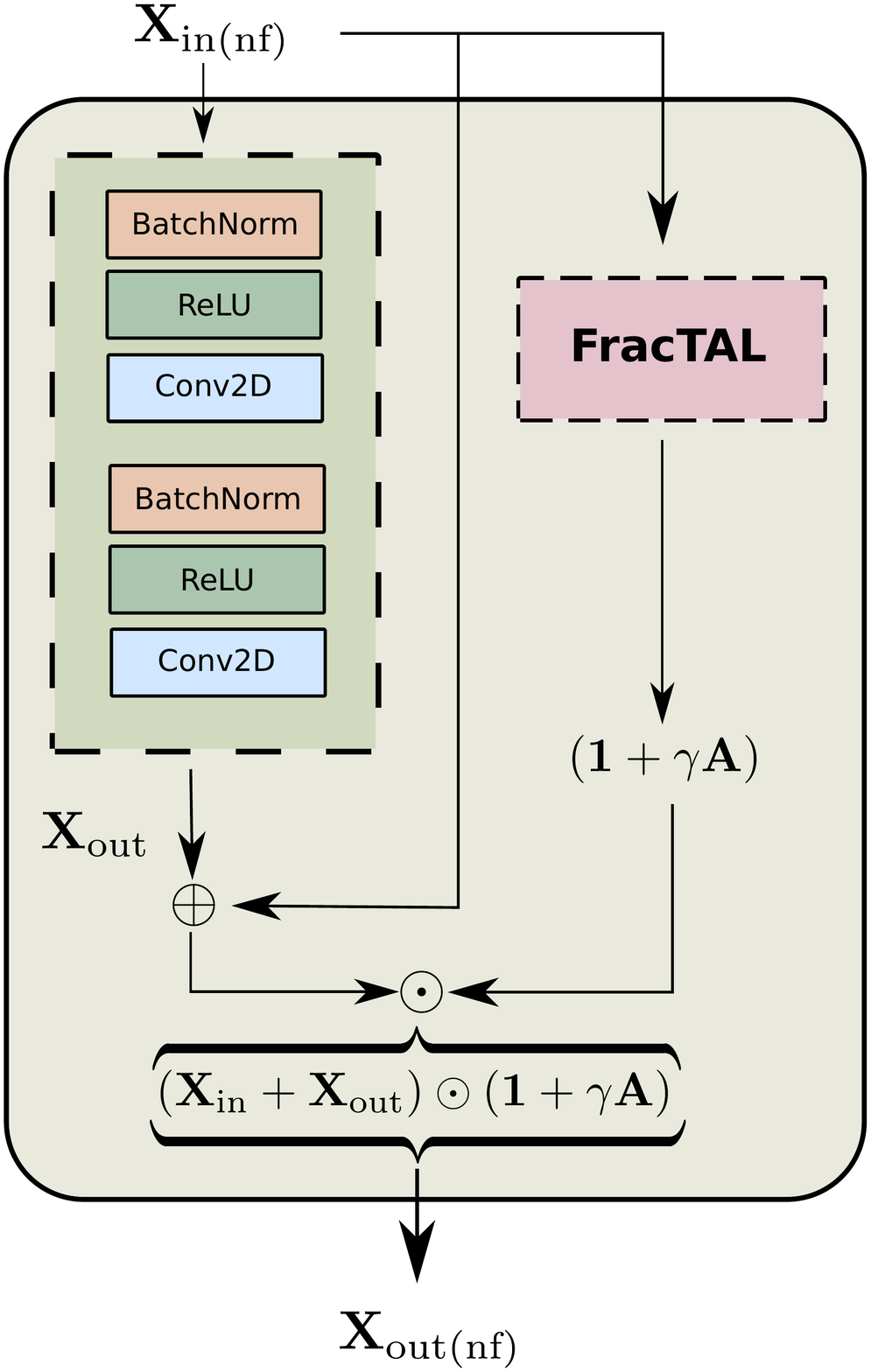}
\caption{The \FracTAL Residual unit. This  building block demonstrates  the fusion of the residual block with self \FracTAL evaluated from the input features.}
\label{ResNetFusion}
\end{figure}

We propose the following fusion methodology between a feature layer, $\mathbf{L}$, and its corresponding \FracTAL  self-attention layer, $\mathbf{A}$:
\begin{equation}
\mathbf{F} = \mathbf{L} +  \gamma \mathbf{L} \odot \mathbf{A}= \mathbf{L}   \odot(\mathbf{1}+\gamma \mathbf{A})
\end{equation}
Here $\mathbf{F}$ is the output layer produced from the fusion of $\mathbf{L}$ and the Attention layer, $\mathbf{A}$, $\odot$ describes element wise multiplication, $\mathbf{1}$ is a layer of ones like $L$, and $\gamma$ a trainable parameter initiated at zero. We next describe  the reasons why we propose this type of fusion. 

 The Attention output is maximal (i.e. close to 1) in areas on the features where it must ``attend'' and minimal otherwise (i.e. close to zero). Multiplying element-wise directly the \FracTAL attention layer $\mathbf{A}$ with the features, $\mathbf{L}$, effectively lowers the values of features in areas that are not ``interesting''. It does not alter the value of areas that ``are interesting''. 
This can produce loss  of information in areas where $\mathbf{A}$ ``does not attend'' (i.e. it does not emphasize), that would be otherwise valuable at a later stage. Indeed, areas of the image that the algorithm ``does not attend'' should not be perceived as empty space \citep{TREISMAN198097}. For this reason the ``emphasized'' features, $\mathbf{L} \odot \mathbf{A}$ are added to the original input $\mathbf{L}$. That is 
$\mathbf{L}+\mathbf{L} \odot \mathbf{A}$ is identical to $\mathbf{L}$ in spatial areas where $\mathbf{A}$ tends to zero, and is emphasized in areas where $\mathbf{A}$ is maximal. 

 In the initial stages of training, the attention layer, $\mathbf{A}$, does not contribute to $\mathbf{L}$, due to the initial value of the trainable parameter $\gamma_0=0$. Therefore it does not add complexity during  the initial phase of training and it allows for an annealing process of Attention  contribution \citep[][see also \citealt{chen2020dasnet}]{zhang2018selfattention}. This property is particularly important when $\mathbf{L}$ is produced from a known performant recipe (e.g. residual building blocks). 

In Fig \ref{ResNetFusion} we present this fusion mechanism for the case of a Residual unit \citep{DBLP:journals/corr/HeZR016,DBLP:journals/corr/HeZRS15}. Here the input layer, $\mathbf{X}_{\text{in}}$, is subject to the residual block sequence of Batch normalization, convolutions, and \texttt{ReLU} activations, and produces the $\mathbf{X}_{\text{out}}$ layer. A separate branch uses the $\mathbf{X}_{\text{in}}$ input to produce the self attention layer $\mathbf{A}$ (see Listing \ref{FTAttentionCODE}). Then we multiply element wise the standard output of the residual unit, $\mathbf{X}_{\text{in}}+\mathbf{X}_{\text{out}}$, with the $\mathbf{1}+\gamma \mathbf{A}$ layer. In this way, at the beginning of training, this layer behaves as a residual layer, which has the excellent convergent properties of resnet at initial stages, and at later stages of training the Attention becomes gradually more active and allows for greater performance. A software routine of this fusion for the residual unit, in particular, can be seen in Listing \ref{FTAttResUnitCODE} in the Appendix.

\subsubsection{Relative attention fusion}
\label{relative_attention_section}

\begin{figure}
\centering
\includegraphics[width=0.82\columnwidth]{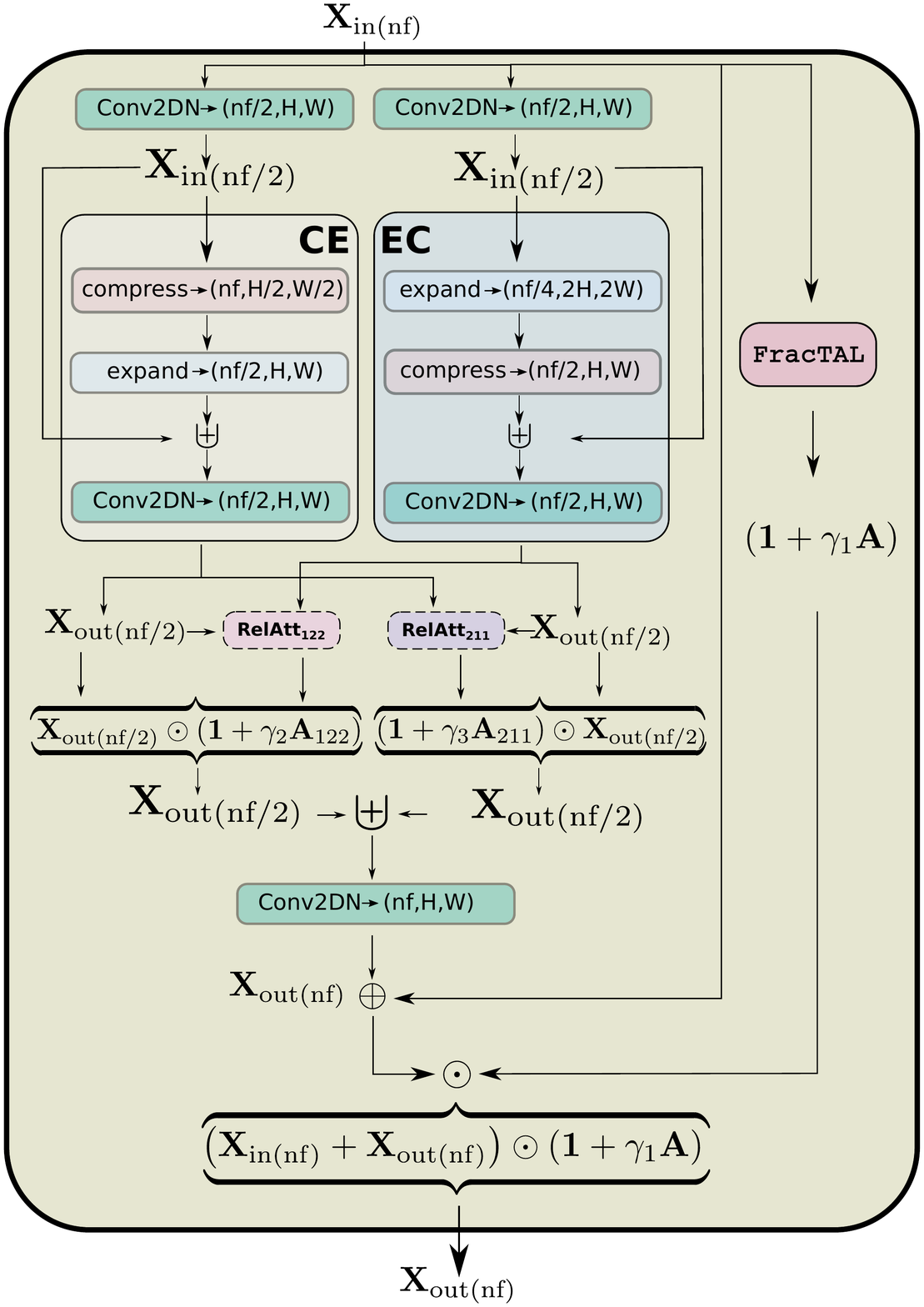}
\caption{Compress Expand Expand Compress unit (CEECNet). The symbol $\uplus$ represents concatenation of features  along the channel dimension (for V1).  For version V2, we replace the concatenation, $\uplus$ followed by the normalized convolution layer with a relative fusion attention, as described in Section \ref{relative_attention_section}} 
\label{ceecnet_unit_v4}
\end{figure}

Assuming we have two input layers, $\mathbf{L}_1, \mathbf{L}_2$, we can calculate the relative attention of each with respect to the other. This is achieved by using as query the layer we want to ``attend to'' and as a key and value the layer we want to use as information for attention. 
In practical implementations, the query, the key, and the value layers result in after the application of a convolution layer to some input.
\begin{align}
\label{rel_fusion_12}
\mathbf{F}_1 &=  \mathbf{L}_1   \odot \left[\mathbf{1}+\gamma_1 \mathbf{A}_{122}(\mathbf{q} (\mathbf{L}_1),\mathbf{k}\left(\mathbf{L}_2\right),\mathbf{v} (\mathbf{L}_2)) \right]\\
\label{rel_fusion_21}
\mathbf{F}_2 &=  \mathbf{L}_2   \odot[\mathbf{1}+\gamma_2 \mathbf{A}_{211}(\mathbf{q} (\mathbf{L}_2),\mathbf{k}\left(\mathbf{L}_1\right),\mathbf{v} (\mathbf{L}_1))]\\
\label{rel_fusion_final}
\mathbf{F} &=  \text{Conv2DN} \left(\text{concat}([\mathbf{F}_1,\mathbf{F}_2]) \right)
\end{align}
Here, the $\gamma_{1,2}$ parameters are initialized at zero, and the concatenation operations are performed along the channel dimension. Conv2DN is a two dimensional convolution operation followed by a normalization layer, e.g. BatchNorm \citep{DBLP:journals/corr/IoffeS15}). An implementation of this process in \textsc{mxnet/gluon} pseudocode style can be found in Listing \ref{FusionCODE}.

The relative attention fusion presented here can be used as  a direct replacement of concatenation followed by a convolution layer in any network design.

\section{Architecture}
\label{section_architecture}

We break down the network architecture into three component parts: the \emph{micro}-topology of the building blocks, which represents the fundamental constituents of the architecture; the \emph{macro}-topology of the network, which describes how building blocks are connected to one another to maximize performance; and the multitasking head, which is responsible for transforming the features produced by the \texttt{micro} and \texttt{macro}-topologies into the final prediction layers where change is identified.
Each of the choices of \texttt{micro} and \texttt{macro} topology has a different impact on the GPU memory footprint.
Usually, selecting very deep \texttt{macro}-topology improves performance, but then this increases the overall memory footprint and does not leave enough space for using an adequate number of filters (channels) in each \texttt{micro}-topology. 
There is obviously a trade off between the \texttt{micro}-topology feature extraction capacity and overall network depth. Guided by this, we seek to maximize the feature expression capacity of the \texttt{micro}-topology  for a given number of filters, perhaps at the expense of consuming computational resources.

\subsection{Micro-topology: the \texttt{CEECNet} unit}
\label{micro_ceecnet_ref}

The basic intuition behind the construction of the \texttt{CEEC} building block, is that it provides two different, yet complementary, views for the same input. The first view (the \textbf{CE} block - see Fig. \ref{ceecnet_unit_v4})  is a ``summary understanding'' operation (performed in lower resolution than the input - see also \citealt{DBLP:journals/corr/NewellYD16,doi:10.1080/01431161.2020.1734251} and \citealt{Qin_2020}). 
The second view (the \textbf{EC} block)  is an ``analysis of detail'' operation (performed in higher spatial resolution than the input). It then exchanges information between these two views using relative attention, and it finally fuses them together, by emphasizing the most important parts using the \FracTAL. 

Our hypothesis and motivation for this approach is quite similar to the scale-space analysis in computer vision \citep{10.5555/528688}: viewing input features at different scales, allows the algorithm to focus on different aspects of the inputs, and thus perform more efficiently. The fact that by merely increasing the resolution of an image does not increase its content information is not relevant here: guided by the loss function the algorithm can learn to represent at higher resolution features that otherwise would not be possible in lower resolutions. We know this from the successful application of convolutional networks in super-resolution problems \citep{DBLP:journals/corr/abs-1902-06068} as well as (variational) autoencoders \citep{DBLP:journals/corr/abs-1812-05069,DBLP:journals/corr/abs-1906-02691}: in both of these paradigms deep learning approaches manage to increase meaningfully the resolution of features that exists in lower spatial dimension layers.

\begin{figure}
\centering
\includegraphics[clip, trim=.5cm 0.30cm 0.4cm 0.3cm,width=0.825\columnwidth]{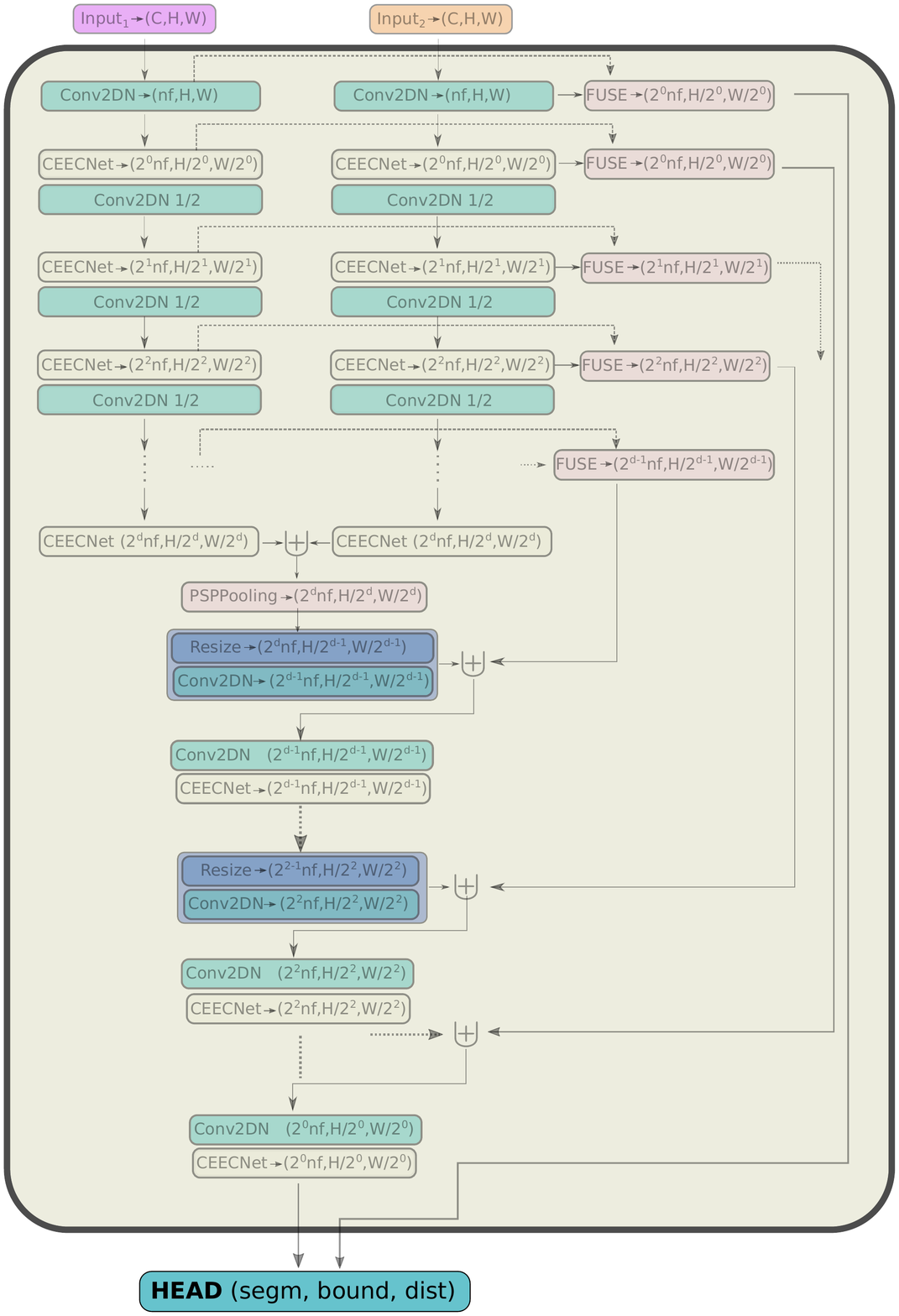}
\caption{The \mantis{} \ceecnet V1 architecture for the task of change detection.  The Fusion  operation (\texttt{FUSE}) is described with \textsc{mxnet/gluon} style pseudocode in detail on Listing \ref{FusionCODE}.} 
\label{mantis_t11_architecture}
\end{figure}

In the following we define the volume $V$ of  features of dimension $(C,H,W)$\footnote{Here, $C$ is the number of channels, $H$ and $W$ the spatial dimensions, height and width respectively.}, as the product of the number of their channels (or filters), $C$ (or $nf$), with their spatial dimensions, height, $H$, and width, $W$, i.e. $V=nf\cdot H \cdot W$\footnote{For example, for each batch dimension, the output volume of a  layer of size $(C,H,W) = (32,256,256)$ is $V=32\cdot256^2 = 2097152$.}. 
The two branches consist of: a ``mini $\cup$-Net'' operation (\textbf{CE} block), that is responsible for summarizing information from the input features by first compressing the total volume of features into half its original size and then restoring it. The second branch,  a ``mini $\cap$-Net'' operation (\textbf{EC} block), is responsible for analyzing in higher detail the input features: it initially doubles the volume of the input features, by halving the number of features and doubling each spatial dimension.  It subsequently compresses this expanded volume to its original size. The input to both layers is concatenated with the output, and then a normed convolution restores the number of channels to their original input value. Note that the mini $\cap$-Net is nothing more than the symmetric (or dual) operation of the mini $\cup$-Net.

The outputs of the \textbf{EC} and \textbf{CE} blocks are fused together with relative attention fusion (section \ref{relative_attention_section}). In this way, exchange of information between the layers is encouraged. The final emphasized outputs are concatened together, thus restoring the initial number of filters, and the produced layer is passed through a normed convolution in order to bind the relative channels.  The operation is concluded with a \FracTAL residual operation and fusion (similar to Fig. \ref{ResNetFusion}), where the input is added to the final output and emphasized by the self attention on the original input. 
The \ceecnet{} building block is described schematically in Fig. \ref{ceecnet_unit_v4}. 

The compression operation, \textbf{C}, is achieved by applying a normed convolution layer of stride equal to 2 (\texttt{k}=3, \texttt{p}=1, \texttt{s}=2) followed by another convolution layer that is identical in every aspect except the stride that is now \texttt{s}=1. The purpose of the first convolution is to both resize the layer and extract features. The purpose of the second  convolution layer is to extract features. The expansion operation, \textbf{E}, is achieved by first resizing the spatial dimensions of the input layer using Bilinear interpolation, and then the number of channels is brought to the desired size by the application of a convolution layer (\texttt{k}=3,\texttt{p}=1,\texttt{s}=1). Another identical convolution layer is applied to extract further features. The full details of the convolution operations used in the \textbf{EC} and \textbf{CE} blocks  can be found on Listing \ref{CEECNetUnitCODE}.

\subsection{Macro-topology: dual encoder, symmetric decoder}
\label{the_mantis_section}

In this section we present the \texttt{macro}-topology (i.e. backbone) of the  architecture that uses as building blocks either the \ceecnet{} or the \FracTAL \texttt{ResNet} units. We start by stating the intuition behind our choices and continue with a detailed description of the \texttt{macro}-topology. Our architecture is heavily influenced from the \texttt{ResUNet-a} model \citep{DIAKOGIANNIS202094}. We will refer to this macro-topology as the \texttt{mantis} topology\footnote{For no particular reason, other than that the mantis shrimp is an \href{https://theoatmeal.com/comics/mantis_shrimp}{amazing sea creature} that resides in the waters of Australia.}. 

In designing this backbone, a key question we tried to address is how can we facilitate exchange of information between features extracted from images at different dates.   The following two observations guided us:
\begin{enumerate}
\item  
We make the hypothesis that the process of change detection between two images requires a mechanism similar to human attention. We base this hypothesis on the fact that  the time required for identifying objects that changed in an image correlates directly with the number of changed objects. That is, the more objects a human needs to identify between two pictures, the more time is required. This is in accordance with the feature-integration theory of Attention   \citep{TREISMAN198097}. In contrast, subtracting features extracted from two different input images is a process that is constant in time, independent of the complexity of the changed features. Therefore, we avoid using adhoc feature subtraction in all parts of the network. 
\item  In order to identify change, a human needs to look and compare two images multiple times, back and forth.  We need \emph{things to emphasize on image at date 1, based on information on image at date 2} (Eq. \ref{rel_fusion_12}), and, vice versa (Eq. \ref{rel_fusion_21}). And then combine both of these information together (Eq. \ref{rel_fusion_final}). That is, exchange information, with relative attention (section \ref{relative_attention_section})  between the two, at multiple levels. A different way of stating this as a question is: what is \emph{important} on input image 1 based on \emph{information} that exists on image 2, and vice versa? 
\end{enumerate}
\textcolor{black}{
Given the above, we now proceed in detailing the \mantis{} macro-topology (with \ceecnet V1 building blocks, see Fig. \ref{mantis_t11_architecture}). 
The encoder part is a series of building blocks, where the size of the features is downscaled between the application of each subsequent building block. Downscaling is achieved with a normed convolution with stride, \texttt{s}=2 without using activations.  There exist two encoder branches that share identical  parameters in their convolution layers. The input to each branch is an image from a different date and the role of the encoder is to extract features at different levels from each input image. During the feature extraction by each branch, each of the two inputs is treated as an independent entity. At successive depths, the  outputs of the corresponding building block are fused together with the relative attention methodology as described in section \ref{relative_attention_section}, but they are not used until later, in the decoder part. Crucially, this fusion operation, suggests to the network that the \emph{important parts of the first layer, will be defined by what exists on the second layer} (and vice versa),  but it does not dictate how exactly  the network should compare the extracted features (e.g. by demanding  the features to be similar for unchanged areas, and maximally different for changed areas\footnote{We tried this approach and it was not successful.}). This is something that the network will have to discover in order to match its predictions with the ground truth. Finally, the last encoder layers are concatenated and inserted to the pyramid scene pooling layer (\texttt{PSPPooling}  --    \citealt{DIAKOGIANNIS202094,zhao2017pspnet}).}

\textcolor{black}{
In the (single) decoder part is where the network extracts features based on the relative information that exist in the two inputs. Starting from the output of the \texttt{PSPPooling} layer (middle of network), we upscale lower resolution features
with bilinear interpolation and combine them with the fused outputs of the decoder with a concatenation operation followed by a normed convolution layer, in a way similar to the \texttt{ResUNet-a} \citep{DIAKOGIANNIS202094} model. The \mantis{} \ceecnet V2 model replaces all concatenation operations followed by a normed convolution, with a \texttt{Fusion} operation as described in Listing \ref{FusionCODE}.} 

The final features extracted from this macro-topology architecture is the final layer from the \ceecnet{} unit that has the same spatial dimensions as the first input layers, as well as the Fused layers from the \emph{first} \ceecnet{} unit operation. Both of these layers are inserted in the segmentation \texttt{HEAD}.
\begin{figure}
\centering
\includegraphics[clip, trim=0.25cm 0.20cm 0.25cm 0.1cm,width=0.7\columnwidth]{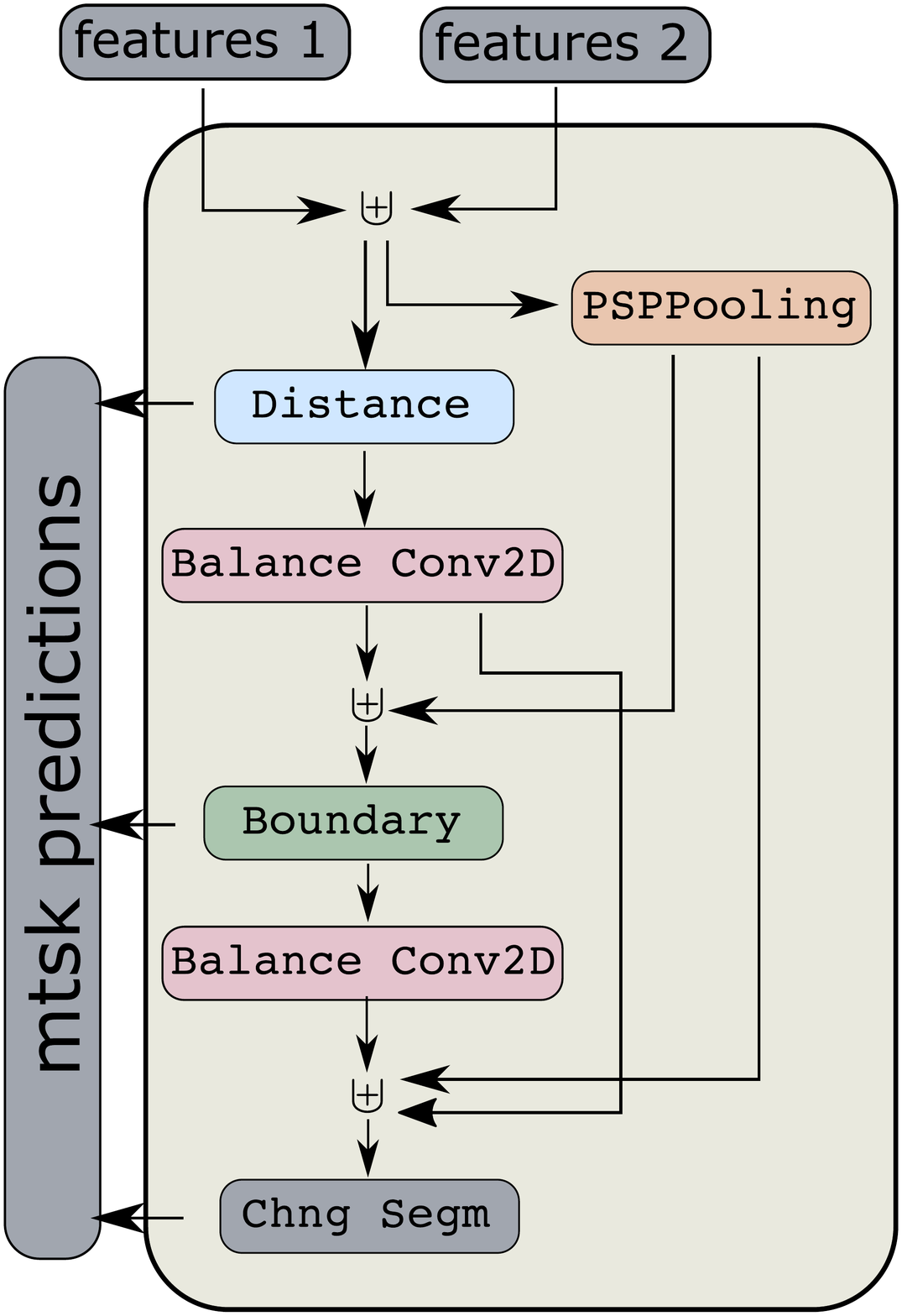}
\caption{Conditioned multitasking segmentation \texttt{HEAD}. Here, features 1 and 2 are the outputs of the \texttt{mantis} \ceecnet{} features extractor. The symbol $\uplus$ represents concatenation along the channels dimension. The algorithm first predicts the distance transform of the classes (regression), then re-uses this information to estimate the boundaries and finally both of these predictions are re-used for the change prediction layer. \textcolor{black}{Here, \texttt{Chng Segm} stands for change segmentation layer, and \texttt{mtsk} for multitasking predictions.}}
\label{classification_HEAD}
\end{figure}
\subsection{Segmentation  \texttt{HEAD}}

The features extracted from the features extractor (Fig. \ref{mantis_t11_architecture}) are  inserted to a conditioned multitasking segmentation head (Fig. \ref{classification_HEAD}) that produces three layers: a segmentation mask, a boundary mask and a distance transform mask. This is identical with the   \texttt{ResUNet-a} ``causal'' segmentation head, that has shown great performance in a variety of segmentation tasks \citep{DIAKOGIANNIS202094,WALDNER2020111741},  with two modifications.

The first modification relates to the evaluation of boundaries: instead of using a standard sigmoid activation for the boundaries layer, we are inserting a scaling parameter, $\gamma$, that controls how sharp the transition from 0 to 1 takes place, i.e. 
\begin{equation}
\texttt{sigmoid}_{\text{crisp}}(x) = \texttt{sigmoid}(x/\gamma),\quad \gamma \in [\epsilon,1]
\end{equation} 
Here $\epsilon = 10^{-2}$ is a smoothing parameter. The $\gamma$ coefficient is learned during training. We inserted this scaling after noticing in initial experiments that the algorithm needed improvement close to the boundaries of objects. In other words, the algorithm was having difficulty  separating nearby pixels. Numerically, we anticipate that the distance between the  values of activations of neighbouring pixels is small, due to the patch-wise nature of convolutions. Therefore, a remedy to this problem is making the transition boundary sharper.  
 We initialize training with $\gamma = 1$.

The second modification to the segmentation \texttt{HEAD} relates to balancing the number of channels of the boundaries and distance transform predictions before re-using them in the final prediction of segmentation change detection. This is achieved by passing them through a convolution layer that brings the number of channels to the desired number. Balancing the number of channels treats the input features and the intermediate predictions as equal contributions to the final output. 
In Fig. \ref{classification_HEAD} we present schematically the conditioned multitasking head, and the various dependencies between layers. Interested users can refer to \cite{DIAKOGIANNIS202094} for details of the conditioned multitasking head.

\section{Experimental Design}
\label{section_experimental_design}

In this section, we describe the setup of our experiments for the evaluation of the proposed algorithms on the task of change detection.  
We start by describing the two datasets we used (LEVIRCD \citealt{rs12101662} and WHU \citealt{Ji2019FullyCN}) as well as the data augmentation methodology we followed. Then we proceed in describing  the metrics used for performance evaluation and the inference methodology. All models \mantis{} \ceecnet V1, \textcolor{black}{V2} and \mantis{} \FracTAL{} \texttt{ResNet} have an initial number of filters equal to \texttt{nf}=32, and the depth of the encoder branches was equal to 6. We designate these models with \texttt{D6nf32}. 

\subsection{LEVIRCD Dataset}

The LEVIR-CD change detection dataset \citep{rs12101662} consists of 637 pairs of  VHR aerial images of resolution 0.5m per pixel. It covers various types of buildings, such as villa residences, small garages, apartments, and warehouses. It contains 31,333 individual building changes. 
The authors provide a train/validation/test split, which standardizes the performance process. We used a different split for training and validation, however, we used the test set the authors provide for reporting performance.  
\textcolor{black}{ For each tile from the training and validation set, we used $\sim$47\% of the area  for training and the remaining $\sim$53\% for validation.
For a rectangle area with sides of length $a$ and $b$, this is achieved by using as training area the rectangle with sides $a'=0.6838\, a$ and $b'=0.6838 \, b$, i.e.  $\texttt{training area}=0.6838^2\, ab \approx 0.47 \, ab$. Then $\texttt{val area} = 1. - \texttt{train area} \approx 0.53 \texttt{total area}$.  
 From each of these areas, we extracted chips of size $256\times 256$. These are overlapping in each dimension with stride equal to $256/2=128$ pixels. }

\begin{figure}
\centering
\includegraphics[clip, trim=.8cm .1cm .5cm 0.1cm,width=\columnwidth]{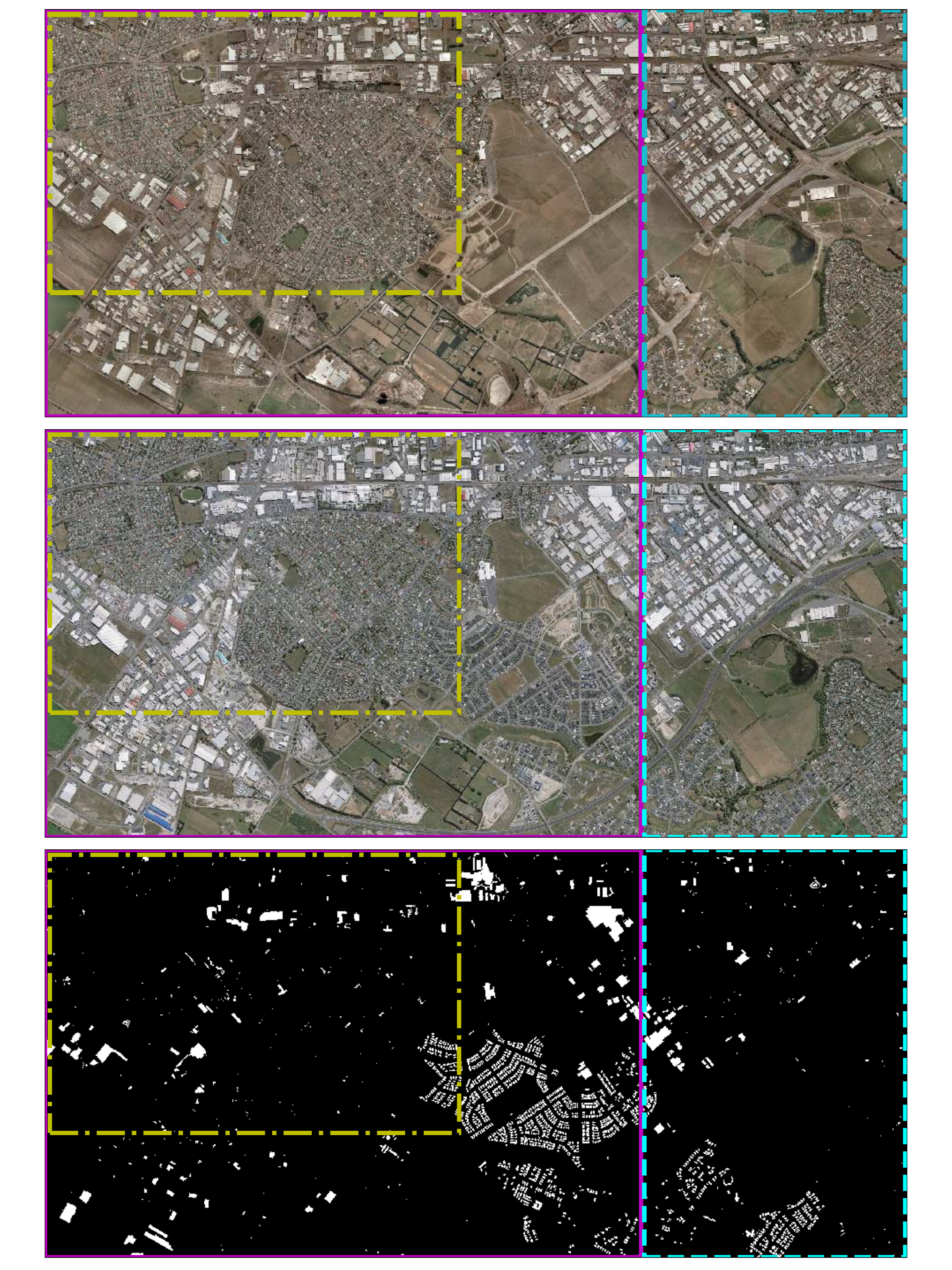}
\caption{Train - validation - test split of the WHU dataset. The yellow (dash-dot line) rectangle represents the training data. The area between the magenta rectangle (solid line) and the yellow (dash-dot) represents the validation data. Finally, the cyan rectangle (dashed) is the test data. The reasoning for our split is to include in the validation data both industrial and residential areas and isolate (spatially) the training area from the test area in order to avoid spurious spatial correlation between training/test sites. \textcolor{black}{The train/validation/test ratio split is $\texttt{train:val:test}\approx 33:37:30$}.} 
\label{NZBLDGCD_TRAIN_TEST_SPLIT}
\end{figure}

\subsection{WHU Building Change Detection}

The WHU building change dataset \citep{Ji2019FullyCN} consists of two  aerial images (2011 and 2016) that 
 cover an area of $\sim20$km${}^2$, which was changed from 2011 (earthquake) to 2016.
 The images resolution is 0.3m spatial resolution. The dataset contains 12796 buildings. We split the triplets of images and ground truth change labels, in three areas with ratio 70\% for training and validation and 30\% for testing. 
 \textcolor{black}{We further split the 70\% part in $\sim$47\% area for training and $\sim$53\% area for validation, in a way similar to the split we followed for each tile of the LEVIRCD dataset}. 
The splitting can be seen in Fig. \ref{NZBLDGCD_TRAIN_TEST_SPLIT}. 
Note that the training area is spatially separated from the test area (the validation area is in between the two). \textcolor{black}{The reason for the rather large train/validation ratio is for us to ensure there is adequate spatial separation between training and test areas, thus minimize spatial correlation effects.}

\subsection{Data preprocessing and augmentation}

We split the original tiles in training chips of size $F^2=256^2$ by using a sliding window methodology with stride $\texttt{s}=F/2=128$ pixels (the chips are overlapping in half the size of the sliding window). This is the maximum size we can fit to our architecture due to GPU memory limitations that we had at our disposal (NVIDIA P100 16GB). With this batch size we managed to fit a batch size of 3 per GPU for each of the architectures we trained. Due to the small batch size, we used \texttt{GroupNorm} \citep{DBLP:journals/corr/abs-1803-08494} for all normalisation layers.

The data augmentation methodology we used during training our network was the one used for semantic segmentation tasks as described in \cite{DIAKOGIANNIS202094}. That is, random rotations with respect to a random center with a (random) zoom in/out operation. We also implemented random brightness and  random polygon shadows. In order to help the algorithm explicitly on the task of the change detection, we implemented time reversal (reversing the order of the input images should not affect the binary change mask) and random identity (we randomly gave as input one of the two images, i.e. null change mask). These latter transformations were implemented at a rate of 50\%.

\subsection{Metrics}
In this section, we present the metrics we used for quantifying the performance of our algorithms. With the exception of the Intersection over Union (IoU) metric, for the evaluation of all other metrics we used the \textsc{Python} library \textsc{pycm} as described in \cite{Haghighi2018}. The statistical measures we used in order to evaluate the performance of our modelling approach are pixel-wise precision, recall, F1 score, Matthews Correlation Coefficient (MCC) \citep{MATTHEWS1975442} and the Intersection over union. These are defined through: 
\begin{align*}
\text{precision} &= \frac{\tp}{\tp+\fp}\\
\text{recall} &= \frac{\tp}{\tp+\fn}\\
\text{F1} &= 2\frac{\text{precision} \times \text{recall}}{\text{precision}+\text{recall}}\\
\text{MCC} &= \frac{\tp \times \tn - \fp \times \fn}{\sqrt{(\tp+\fp)(\tp+\fn)(\tn+\fp)(\tn+\fn)}}\\
\mathrm{IoU} &= \frac{\tp}{\tp+\fn+\fp}
\end{align*}

\subsection{Inference}

In this section, we provide a brief description of the model selection after training (i.e. which epochs will perform best on the test set) as well as the inference methodology we followed for large raster images that exceed the memory capacity of modern-day GPUs.

\subsubsection{Inference on large rasters}

Our approach is identical to the one used in  \cite{DIAKOGIANNIS202094}, with the difference that now we are processing two input images. Interested readers that want to know the full details can refer to Section 3.4 of \cite{DIAKOGIANNIS202094}. 

During inference on test images, we extract multiple overlapping windows of size $256 \times 256$ with a step (stride) size of $256/4=64$ pixels. The final prediction ``probability'', per pixel, is evaluated as the average ``probability'' over all inference windows that overlap on the given pixel. In this definition, we refer to ``probability'' as the output of the softmax final classification layer, which is a continuous value in the range $[0,1]$. It is not a true probability, in the statistical sense, however, it does express the confidence of the algorithm in obtaining the inference result.  

With this overlapping approach, we make sure that the pixels that are closer to the edges and correspond to boundary areas for some inference windows, appear closer to the center area of subsequent inference windows. For the boundary pixels of the large raster, we apply reflect padding before performing inference \citep{DBLP:journals/corr/RonnebergerFB15}.

\begin{figure}
\centering
\includegraphics[width=\columnwidth]{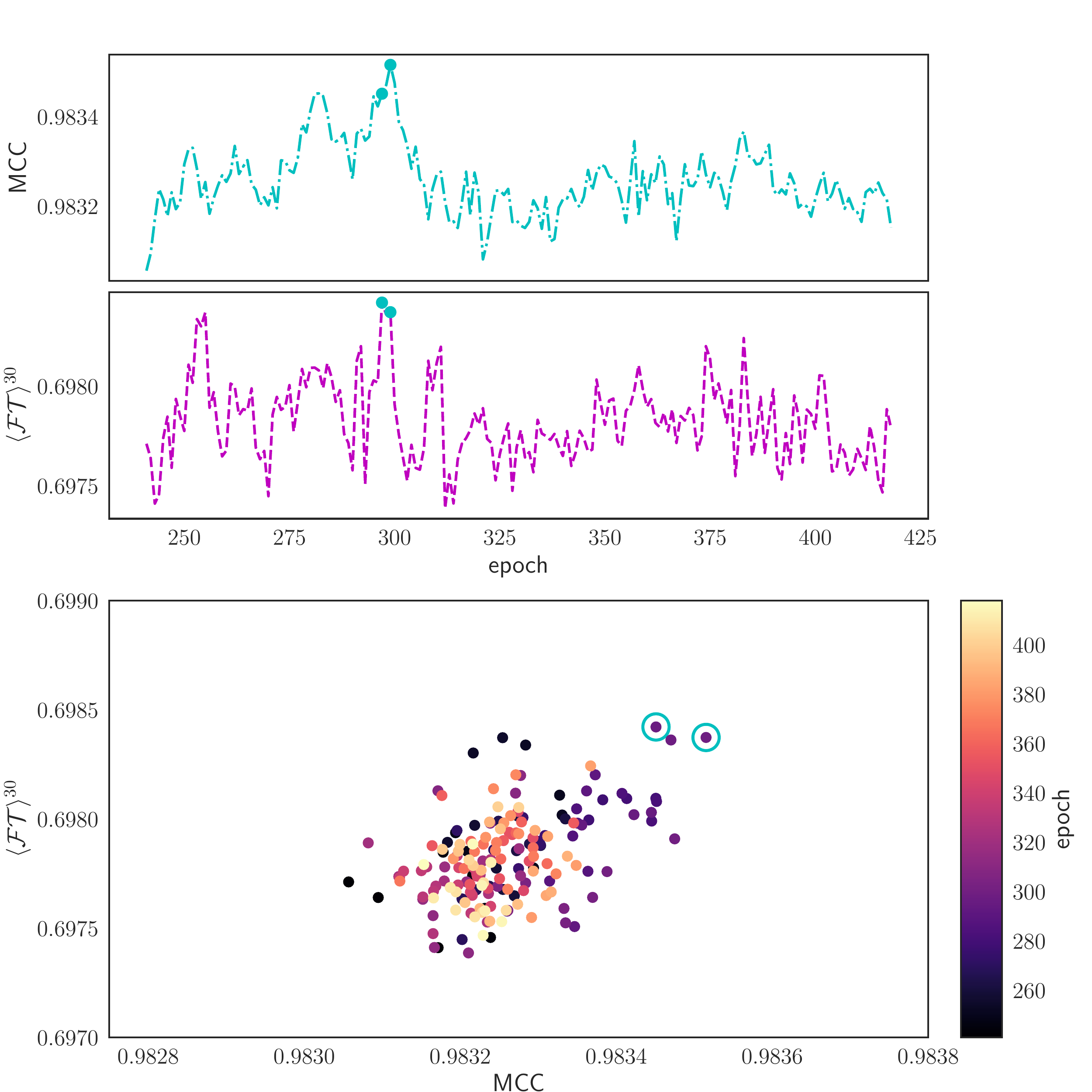}
\caption{Pareto front selection after the last reduction of learning rate. The bottom panel designates with open cyan circles the two points that are equivalent in terms of quality prediction when both MCC and $\langle \ftnmt \rangle$ are taken into account. The top two panels show the corresponding evolutions of these measures during training. There, the Pareto optimal points are designated with full circle dots (cyan). } 
\label{pareto_selection}
\end{figure}

\subsubsection{Model selection using Pareto efficiency}

For monitoring the performance of our modelling approach, we usually rely on the MCC metric on the validation dataset. We observed, however, that when we perform simultaneously learning rate reduction and $\langle\ftnmt\rangle^{d}$ depth increase, initially the MCC decreases (indicating performance drop), while the $\langle\ftnmt\rangle^{d}$ similarity  is (initially) strictly increasing. After training starts to stabilize around some optimality region (with the standard noise oscillations), there are various cases where the MCC metric and $\langle\ftnmt\rangle^{d}$ similarity coefficient do not agree on which is the best model. To account for this effect and avoid losing good candidate solutions, we evaluate the average of the inference output of a set of best candidate models. These best candidate models are selected according to the models that belong to the Pareto front of the most evolved solutions. We use all the Pareto front \citep{10.1007/s11047-018-9685-y} model weights as acceptable solutions for inference. A similar approach was followed for the selection of hyper parameters for optimal solutions in \cite{WALDNER2020111741}.

In Fig. \ref{pareto_selection} we plot on the top panel the evolution of the MCC, and $\langle\ftnmt\rangle^{d}$ for $d=30$. Clearly, these two performance metrics do not always agree. For example, the $\langle\ftnmt\rangle^{30}$ is close to optimality in approximate epoch $\sim$250, while the MCC is clearly suboptimal. We highlight with filled circles (cyan dots) the two solutions that belong to the pareto front. In the bottom panel we plot the correspondence of the MCC values with the $\langle\ftnmt\rangle^{30}$ similarity metric. The two circles show the corresponding non-dominated Pareto solutions (i.e. best candidates).

\section{\FracTAL units and evolving loss ablation study}
\label{section_ablation_study}

\textcolor{black}{
In this section    we present the performance of the \FracTAL{} \texttt{ResNet} \citep{DBLP:journals/corr/HeZRS15,DBLP:journals/corr/HeZR016} and \ceecnet{} units we introduced against ResNet and CBAM \citep{10.1007/978-3-030-01234-2_1} baselines as well as  
 the effect of the evolving $\langle \ftnmt\rangle ^d$ loss function on training a neural network. We also present a qualitative and quantitative analysis on the effect of the depth parameter in the \FracTAL based on the \mantis{} \FracTAL \texttt{ResNet} network.}

\subsection{\FracTAL building blocks performance}

 We construct three identical networks in \texttt{macro}-topological graph (backbone), but different in \texttt{micro}-topology (building blocks). The first two networks are equipped  with two different versions of \ceecnet: the first is identical with the one presented in Fig.  \ref{ceecnet_unit_v4}. The second is similar to the one in Fig.  \ref{ceecnet_unit_v4} with all concatenation operations that are followed by normed convolutions being replaced with Fusion operations, as described in Listing \ref{FusionCODE}. 
The third network uses as building blocks the \FracTAL ResNet building blocks (Fig.  \ref{ResNetFusion}). 
 Finally, the fourth network uses as building blocks standard residual units  as described in \cite{DBLP:journals/corr/HeZRS15,DBLP:journals/corr/HeZR016} (ResNet V2). All building blocks have the same dimensionality of input and output features. However, each type of building block has a different number of parameters. By keeping  the dimensionality of input and output layers  identical to all layers, we believe, the performance differences of the networks will reflect  the feature expression capabilities of the building blocks we compare.

\begin{figure}
\centering
\includegraphics[width=\columnwidth]{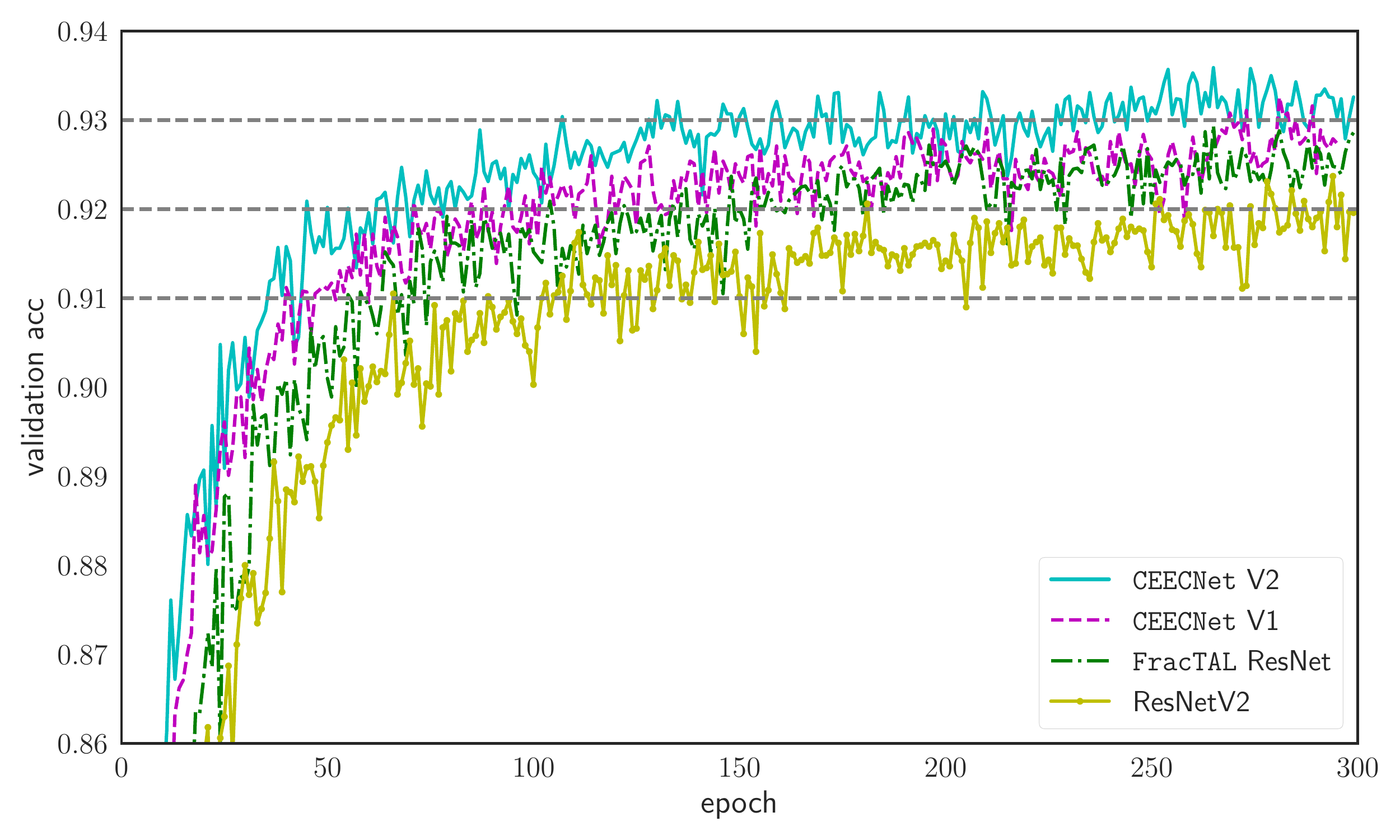}
\caption{Comparison of the V1 and V2 versions of  \ceecnet{} building blocks with  a \FracTAL ResNet implementation and a standard ResNet V2 building blocks. The models were trained for 300 epochs on CIFAR10 with standard cross entropy loss.} 
\label{CEECNet_vs_ResNet_comparison}
\end{figure}

 In Fig. \ref{CEECNet_vs_ResNet_comparison} we plot the validation loss for 300 epochs of training on CIFAR10  dataset \citep{Krizhevsky09learningmultiple} without learning rate reduction, We use cross entropy loss and Adam optimizer \citep{DBLP:journals/corr/KingmaB14}. The backbone of each of the networks is described in Table \ref{ceecnet_vs_resnet}. 
It can be seen  that the convergence and performance of all building blocks equipped with the \FracTAL outperform standard Residual units. In particular we find that the performance and convergence properties of the networks follow:
\texttt{ResNet} $\langle$ \FracTAL{}\texttt{ResNet} $\langle$ \ceecnet V1 $\langle$ \ceecnet V2. The performance difference between \FracTAL{}\texttt{ResNet} and \ceecnet V1 will become more clearly apparent in the change detection datasets.  The V2 version of \ceecnet{} that uses \texttt{Fusion} with relative attention (cyan solid line) instead of concatenation (V1 - magenta dashed line), for combining layers in the Compress-Expand and Expand-Compress branches, has superiority over V1. However, it is a computationally more  intensive unit.

\begin{figure}
\centering
\includegraphics[clip, trim=.8cm .1cm .5cm 0.1cm,width=\columnwidth]{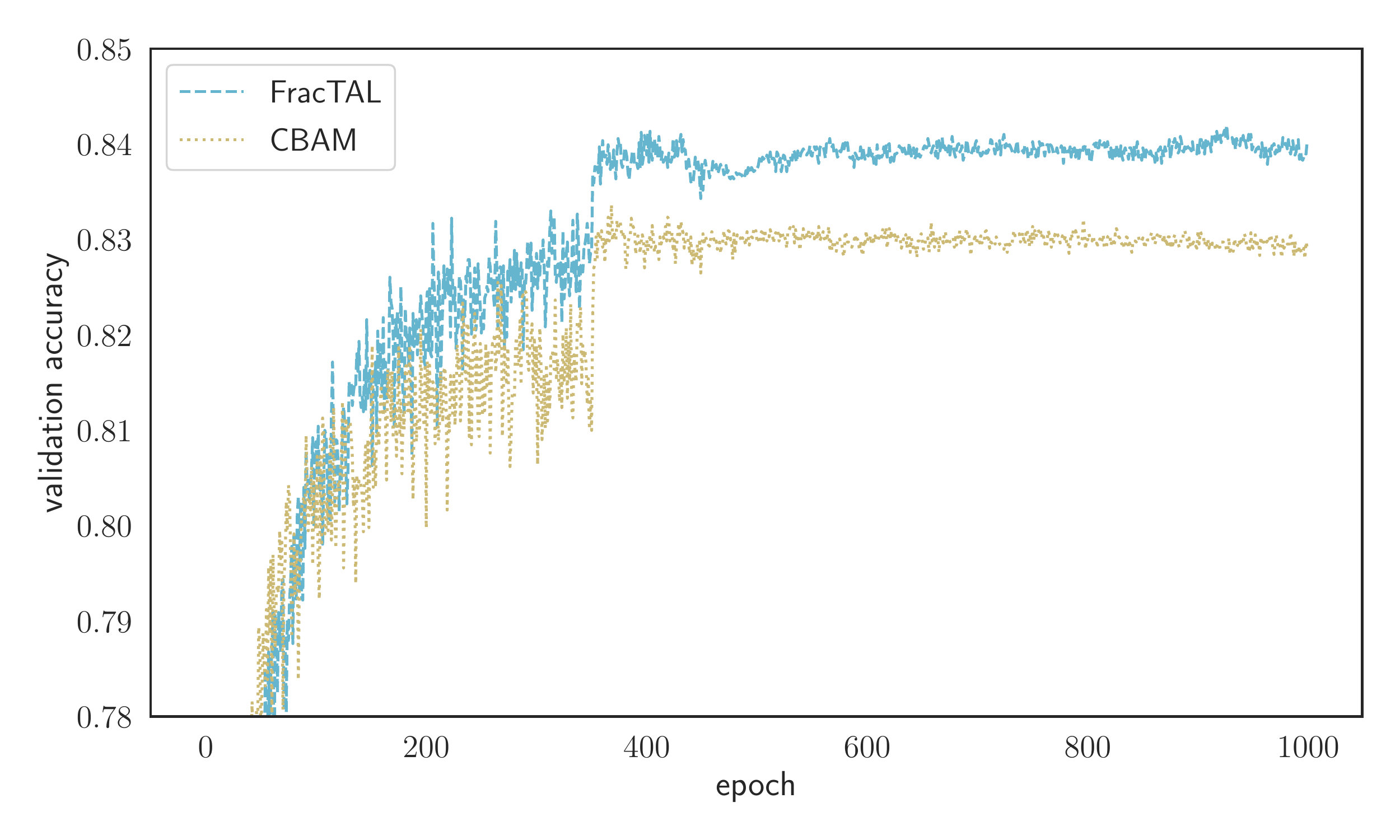}
\caption{\textcolor{black}{Performance improvement of the \texttt{FracTAL}-{\texttt{resnet34}} over \texttt{CBAM-resnet34}: replacing the \texttt{CBAM} attention layers, with \FracTAL ones, for two otherwise identical networks, results in 1\% performance improvement.}} 
\label{fractal_vs_cbam}
\end{figure}

\subsection{\textcolor{black}{Comparing \FracTAL with CBAM}}

Having shown the performance improvement over the residual unit,  we proceed in comparing the \FracTAL proposed attention with a modern attention module, and in particular the    Convolution Block Attention Module (\texttt{CBAM}) \citep{10.1007/978-3-030-01234-2_1}. 
We construct two networks that are identical in all aspects except the implementation of the attention used.  We base our implementation on a publicly available repository that reproduces the results of \cite{10.1007/978-3-030-01234-2_1} - written in \textsc{Pytorch}\footnote{\textcolor{black}{\href{https://github.com/luuuyi/CBAM.PyTorch}{https://github.com/luuuyi/CBAM.PyTorch}.}} - that we translated  into the \textsc{mxnet} framework. From this implementation we use the \texttt{CBAM-resnet34} model and we compare it with a \texttt{FracTAL-resnet34} model, i.e. a model which is identical to the previous one, with the exception that we replaced the \texttt{CBAM} attention with the \FracTAL (attention). Our results can be seen on Fig. \ref{fractal_vs_cbam}, where a clear performance improvement is evident merely by changing the attention layer used. The  improvement is of the order of 1\%, from 83.37\% (\texttt{CBAM}) to 84.20\% (\texttt{FracTAL}), suggesting that the \FracTAL has better feature extraction capacity than the \texttt{CBAM} layer.

\subsection{\textcolor{black}{Evolving loss}}

\begin{figure}
\centering
\includegraphics[width=\columnwidth]{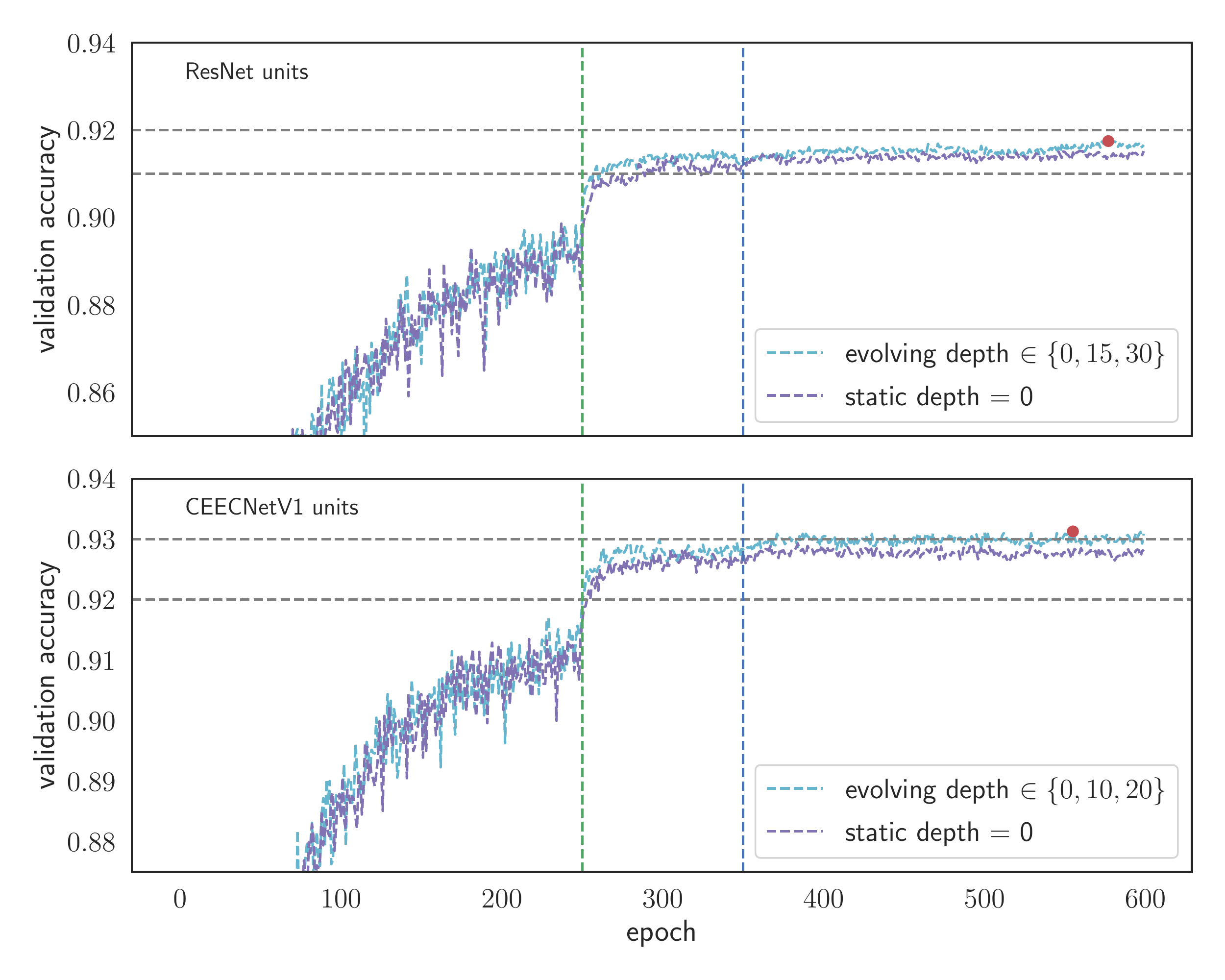}
\caption{\textcolor{black}{Training on CIFAR10 of two classification networks with static and evolving loss strategies. The two networks have  identical \texttt{macro}-topologies, but  different \texttt{micro}-topologies. The first network (top) uses standard Residual units for its building blocks, while the second (bottom) \texttt{CEECNetV1} units. The networks are trained with a static $\ftnmt$ ($d=0$) loss stragety, and an evolving one. We increase the depth $d$ of the $\ftnmt^{d}(\mathbf{p},\mathbf{l})$  loss function with each learning rate reduction. The vertical dashed lines designate epochs where the learning rate was scaled to 1/10th of its original value. The validation accuracy is mildly increased, although there is a clear difference.}}
\label{EvolvingLoss_cifar10}
\end{figure}

\begin{figure*}
\centering
\includegraphics[clip, trim=4.5cm 6.1cm 3.cm 6.15cm,width=\columnwidth]{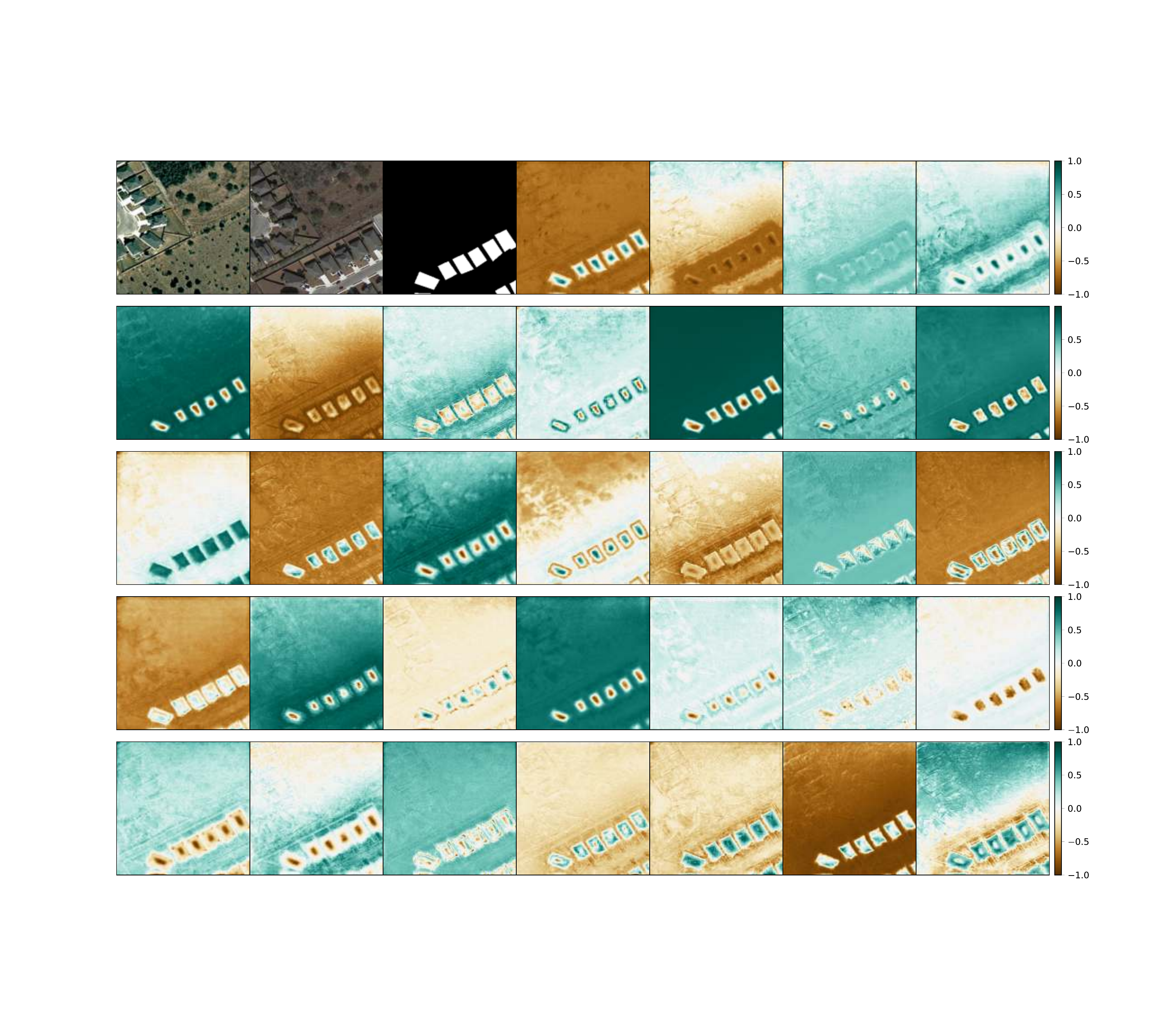}
\includegraphics[clip, trim=4.5cm 6.1cm 3.cm 6.15cm,width=\columnwidth]{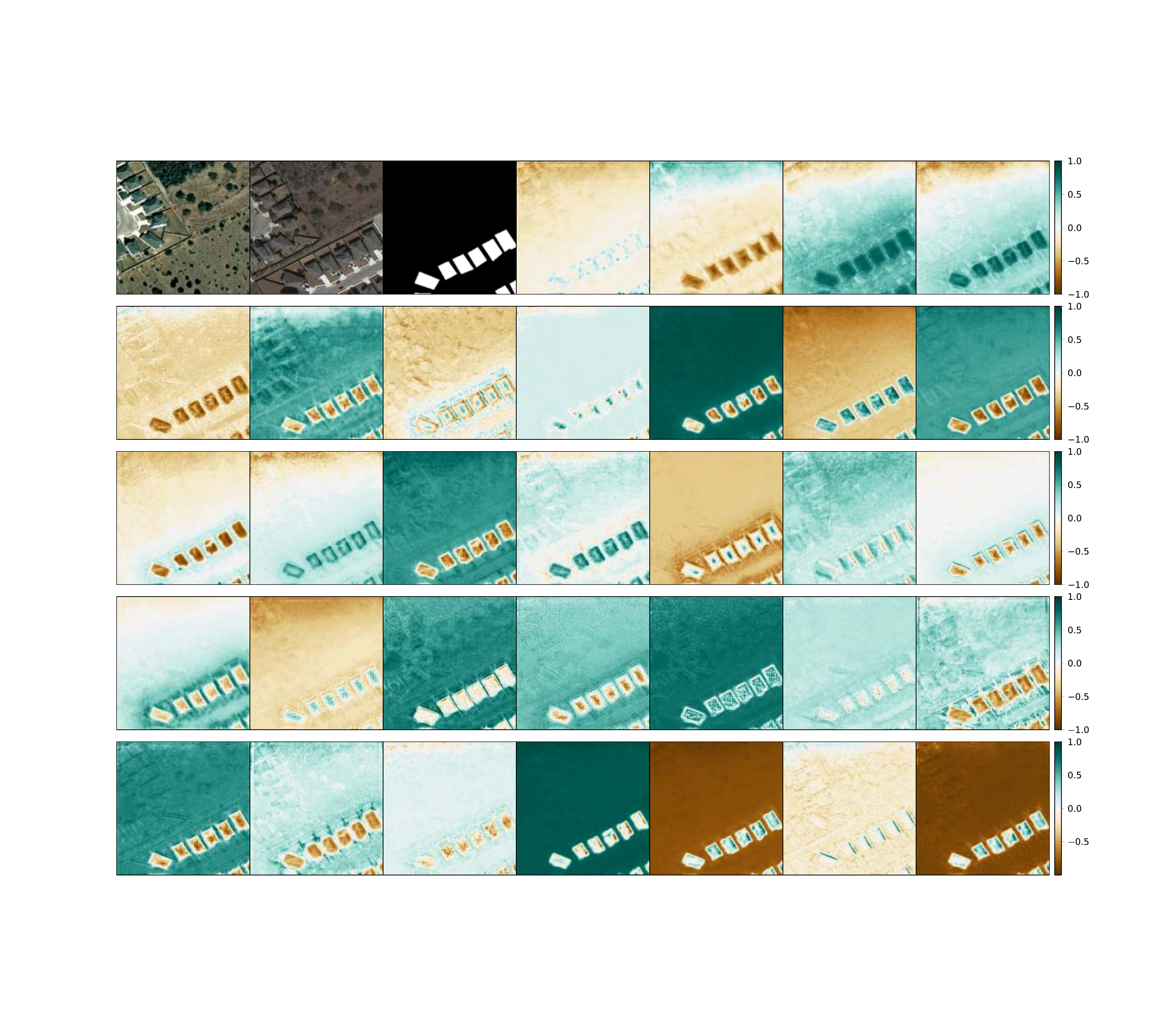}
\caption{\textcolor{black}{Visualization of the last features (before the multitasking head) for the \mantis{}\FracTAL \texttt{ResNet} models of \FracTAL depth $d=0$ (left pannel) and $d=10$ (right pannel). The features appear similar. For each panel the top left first three images are the input image at date $t_1$, the input image at date $t_2$ and the ground truth mask.}} 
\label{FracTAL_d0_vs_d10_bfr_head}
\end{figure*}

\textcolor{black}{
We continue by presenting experimental results on the performance of the evolving  loss strategy on \textsc{CIFAR10} using two networks, one with  standard  ResNet building blocks and one with  \ceecnet \texttt{V1} units. The macro topology of the  networks is identical to the one in Table \ref{ceecnet_vs_resnet}. In addition, we also demonstrate performance differences on the change detection task, by training the \mantis {} \ceecnet V1 model on the LEVIRCD dataset, with static and evolving loss strategies for \FracTAL depth, $d=5$.} 

\textcolor{black}{
In Fig. \ref{EvolvingLoss_cifar10} we demonstrate the effect of this approach: we train the network on \textsc{CIFAR10} with standard residual blocks \citep[top panel][]{DBLP:journals/corr/HeZR016,DBLP:journals/corr/HeZRS15} under the two different loss strategies. In both strategies, we reduce the initial  learning rate by a factor of 10 at epochs 250 and 350. In the first strategy, we train the networks with $\ftnmt^0$. In the second strategy, we evolve the depth of the fractal Tanimoto  loss function: we start training with   $\ftnmt^0$ and on the two subsequent learning rate reductions we use $\langle\ftnmt \rangle^{15}$ and $\langle\ftnmt \rangle^{30}$.  In the top panel, we plot the validation accuracy for the two strategies. The performance gain following the evolving depth loss is $\sim 0.25\%$ in validation accuracy.   
In the bottom panel we plot the validation accuracy for the \ceecnet{}V1  based models. Here, the evolution strategy is same as above with the difference that  we use different depths for the $\ftnmt$ loss (to observe potential differences). These are $d \in \{0,10,20\}$. Again, the difference in the validation accuracy is $\sim +0.22\%$ for the evolving loss strategy. }

\begin{figure*}
\centering
\includegraphics[clip, trim=7.cm 12.8cm 5.2cm 12.75cm,width=\textwidth]{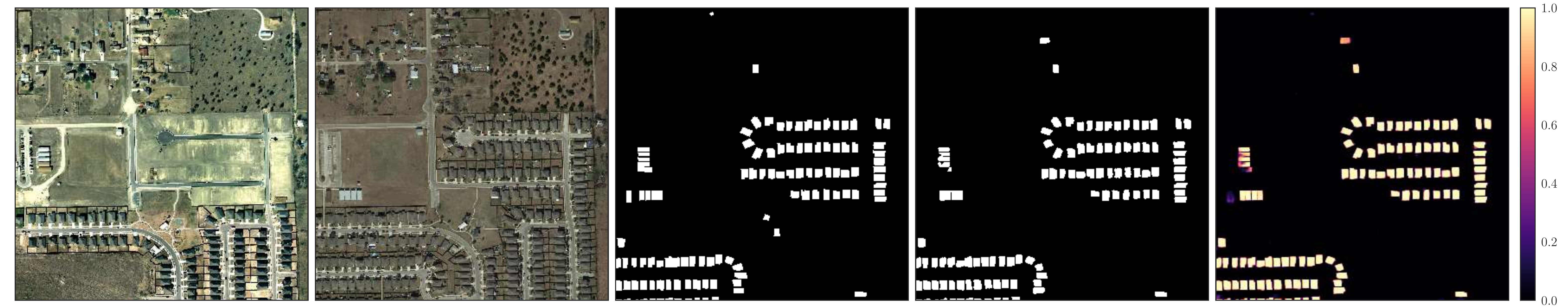}
\includegraphics[clip, trim=7.cm 12.8cm 5.2cm 12.75cm,width=\textwidth]{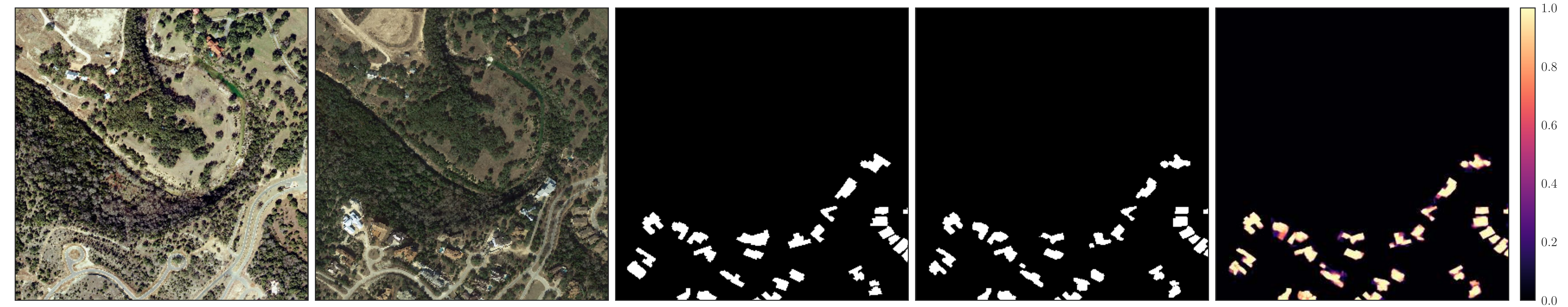}
\includegraphics[clip, trim=7.cm 12.8cm 5.2cm 12.75cm,width=\textwidth]{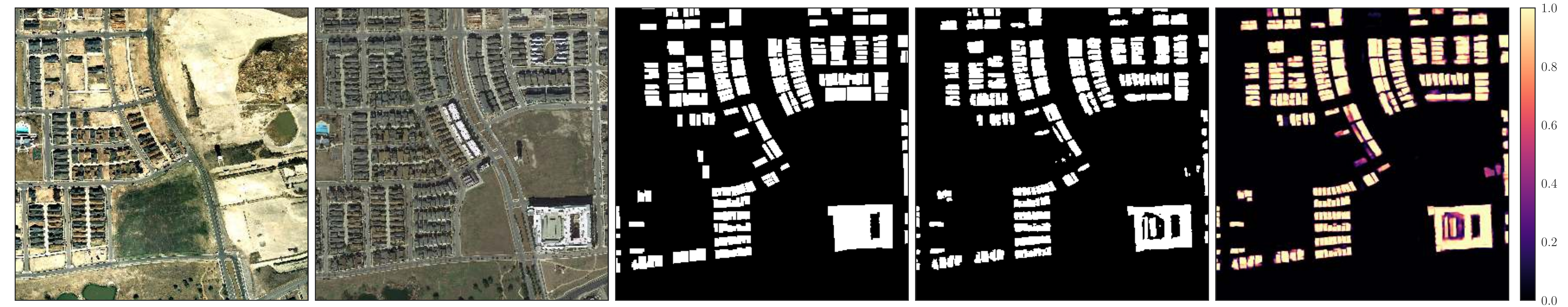}
\includegraphics[clip, trim=7.cm 12.8cm 5.2cm 12.75cm,width=\textwidth]{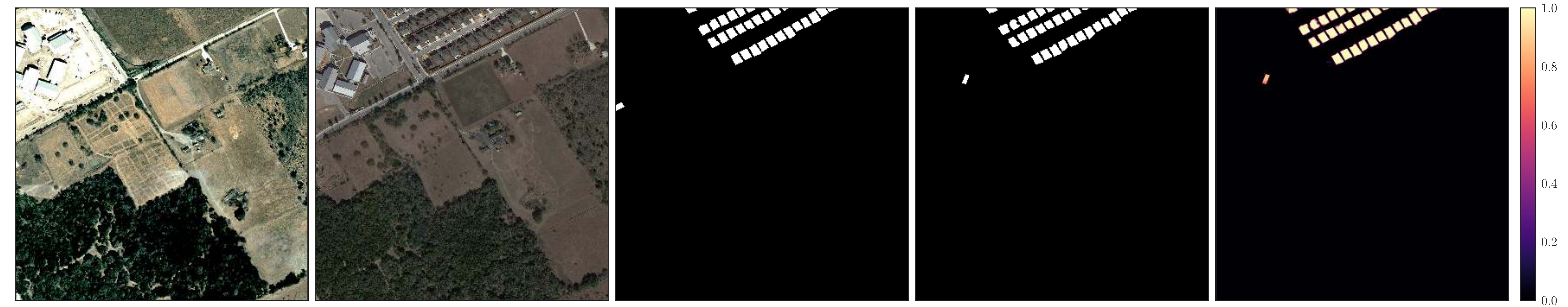}
\includegraphics[clip, trim=7.cm 12.8cm 5.2cm 12.75cm,width=\textwidth]{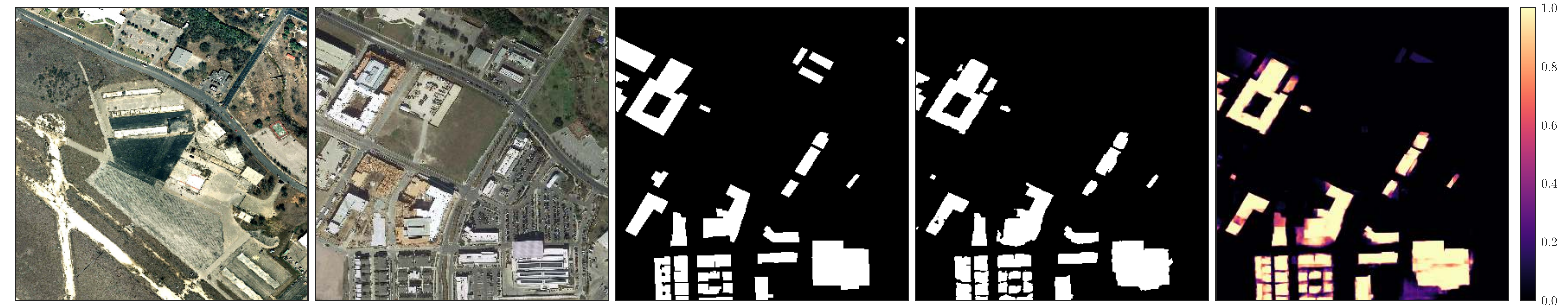}
\includegraphics[clip, trim=7.cm 12.8cm 5.2cm 12.75cm,width=\textwidth]{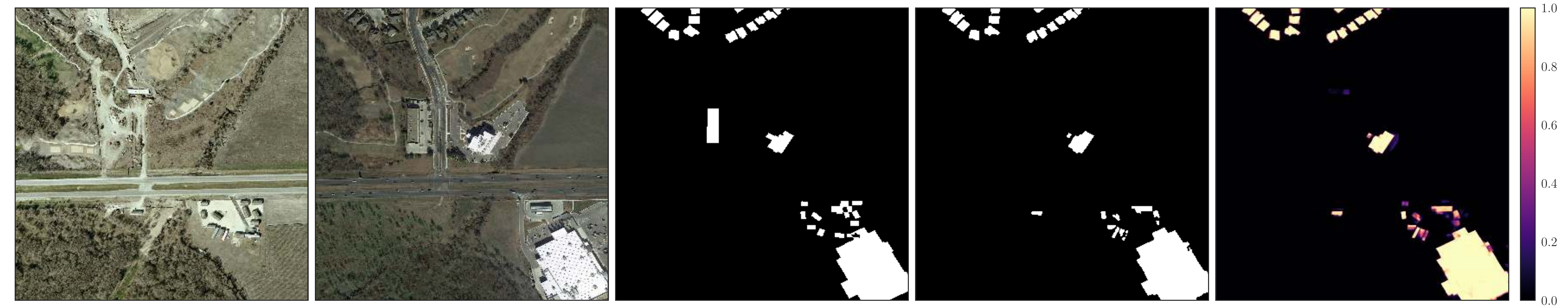}
\caption{Examples of inferred change detection on some test tiles from the LEVIRCD dataset of the \mantis{} \ceecnet V1 model \textcolor{black}{(evolving loss strategy, \FracTAL depth $d=5$)}. For each row, from left to right input image date 1, input image date 2, ground truth,  change prediction (threshold 0.5) and confidence heat map.} 
\label{LEVIRCD_show}
\end{figure*}

\begin{figure}
\centering
\includegraphics[width=\columnwidth]{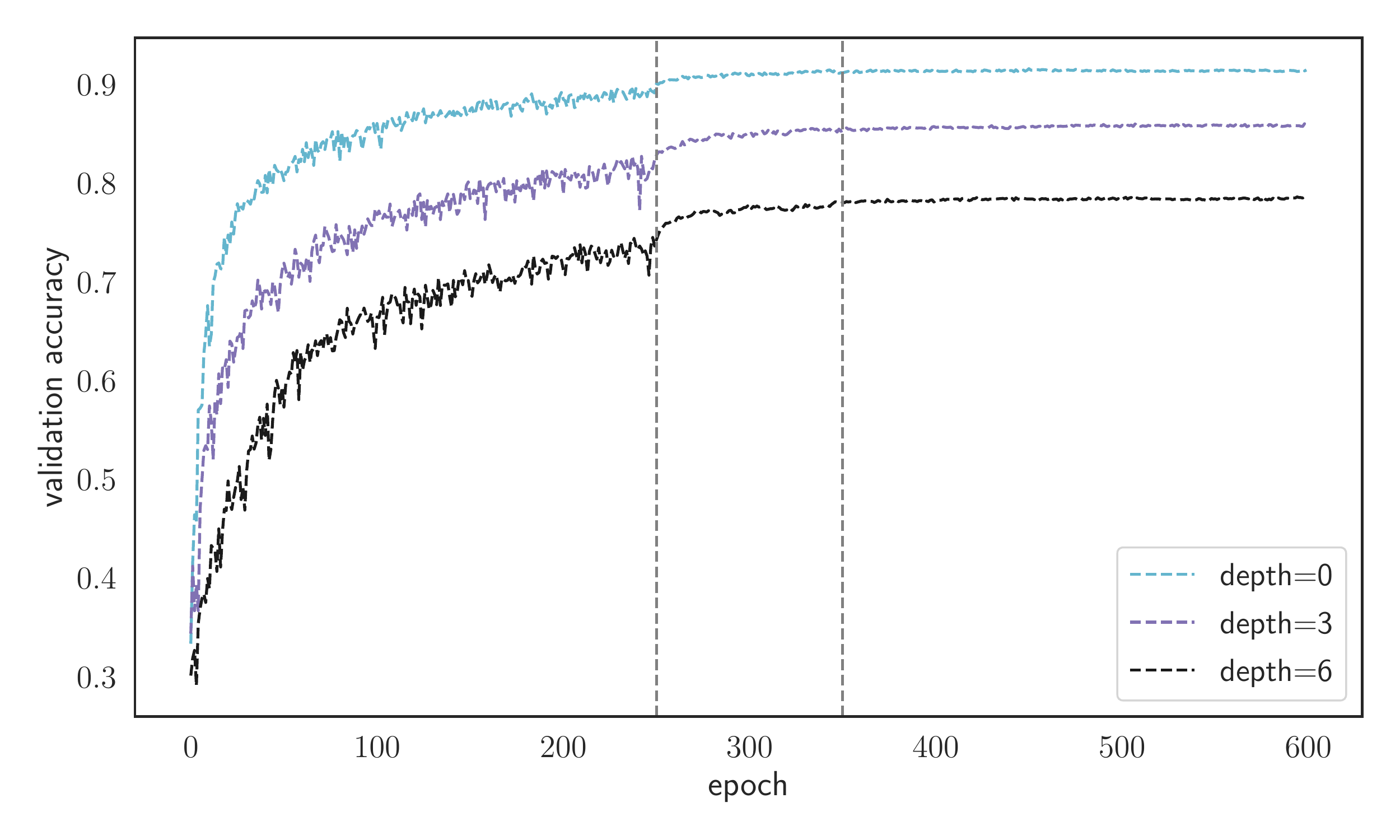}
\caption{Training on CIFAR10 of a network with standard ResNet building blocks and fixed depth, $d$, of the  $\langle \ftnmt^d \rangle$ loss. The vertical dashed lines designate epochs where the learning rate was scaled to 1/10th of its original value. As the depth of iteration, $d$, increases ($d$ remains constant for each training)  the convergence speed of the validation accuracy degrades.} 
\label{cifar10_ft_start_rand}
\end{figure}

We should note that we observed performance degradation by using for training (from random weights) the $\langle\ftnmt\rangle^d$ \textcolor{black}{loss} for $d>1$. This is evident in Fig. \ref{cifar10_ft_start_rand} where we train from scratch on CIFAR10 three identical models with different depth for the $\ftnmt^d$ function: $d=[0,3,6]$. It is seen that as the hyperparameter $d$ increases, the performance of the validation accuracy degrades. 
  We consider that this happens due to the low value of the gradients away from optimality, as it requires the network to train longer to reach the same level of validation accuracy. In contrast, the greatest benefit we observed by using this training strategy is that the network can avoid overfitting after learning rate reduction (provided that the slope created by the choice of depth $d$ is significant) and has the potential to reach higher performance. 

\textcolor{black}{Next we perform a test on evolving vs static loss strategy on the LEVIR CD change detection dataset, using the \ceecnet V1 units, as it can be seen in Table \ref{LEVIRCD_performance}. The \ceecnet V1 unit, trained with the evolving loss strategy, demonstrates +0.856\% performance increase on the Interesection over Union (IoU) and +0.484\% increase in MCC. 
Note that, for the same \FracTAL depth, $d=5$, the \FracTAL ResNet network, trained with the evolving loss strategy performs better than the \ceecnet V1 that is trained with the static loss strategy, while it falls behind the \ceecnet V1 trained with the evolving loss strategy. We should also note that performance increment is larger, in comparison with the classification task on \textsc{CIFAR10}, reaching almost $\sim$1\% for the IoU.}

\subsection{\textcolor{black}{Performance dependence on \FracTAL depth}}

\textcolor{black}{
In order to understand how the \FracTAL layer behaves with respect to different depths, we train three identical networks, the \mantis{} \FracTAL \texttt{ResNet} (\texttt{D6nf32}), using \FracTAL depths in the range $d \in \{0, 5, 10 \}$. The performance results on the LEVIRCD dataset can be seen on Table \ref{LEVIRCD_performance}. It seems the three networks perform similarly (they all achieve SOTA performance on the LEVIRCD dataset), with the $d=10$ having top performance (+0.724\% IoU), followed by the $d=0$ (+0.332\%IoU) and, lastly the $d=5$ network 
(baseline). We conclude that the depth $d$ is a hyper parameter dependent on the problem at task that users of our method can choose to optimize against. Given that all models have competitive performance, it seems also that the proposed depth $d=5$ is a sensible choice. }

\textcolor{black}{
In Fig. \ref{FracTAL_d0_vs_d10_bfr_head} we visualize the features of the last convolution, before the multitasking segmentation head for \FracTAL depth $d=0$ (left panel) and $d=10$ (right panel). The features at different depths appear similar, all identifying the regions of interest clearly. To the human eye, according to our opinion, the features for depth $d=10$ appear slightly more refined in comparison with the features corresponding to depth $d=0$ (e.g. by comparing the images in the corresponding bottom rows). The entropy of the features for $d=0$  (entropy: 15.9982) is negligibly higher (+0.00625 \%) than for  the case $d=10$ (entropy: 15.9972), suggesting both features have the same information content for these two models. We note that, from the perspective of information compression (assuming no loss of information), lower entropy values are favoured over higher values, as they indicate a better compression level.}

\section{Results}
\label{section_results}

In this section, we report the quantitative and qualitative  performance of the models we developed for the task of change detection on the LEVIRCD \citep{rs12101662} and WHU \citep{Ji2019FullyCN} datasets.
 \textcolor{black}{All of the inference visualizations are performed with models having the proposed \FracTAL depth $d=5$, although this is not always the best performant network.}

\subsection{Performance on LEVIRCD}
\label{levircd_performance_section}

For this particular dataset, a fixed test set is provided  and a comparison with methods that other authors followed is possible. Both \FracTAL{} \texttt{ResNet} and \ceecnet{} (V1, V2) outperform the baseline \citep{rs12101662} with respect to the F1 score by $\sim$5\%.

In Fig. \ref{LEVIRCD_show} we present the inference of the \ceecnet{} V1 algorithm for various images from the test set. For each row, from left to right we have input image at date 1, input image at date 2, ground truth mask, inference (threshold = 0.5), and algorithm's confidence\footnote{This should not be confused with statistical confidence.} heat map. It is interesting to note that the algorithm has zero doubt in areas where buildings exist in both input images. That is, it is clear our algorithm identifies change in areas covered by buildings, and not building footprints.  In Table \ref{LEVIRCD_performance} we present numerical performance results of both \FracTAL \texttt{ResNet} as well as \ceecnet{} V1\& V2. All metrics, precision, recall, F1, MCC and IoU are excellent. The \mantis{} \ceecnet{} \textcolor{black}{for \FracTAL depth $d=5$}, outperforms the \mantis{} \FracTAL{} \texttt{ResNet} by a small numerical margin, however the difference is clear. This difference can also be seen in the bottom panel of Fig. \ref{ceecnet_vs_fractal_resunet}. We should also note that the numerical difference on, say, F1 score, does not translate to equal portions of quality difference in images. That is, a 1\% difference in F1 score, may have a significant impact on the quality of inference. 
We further discuss  this on Section \ref{ceecnet_vs_resnet_qualititative}. 
\textcolor{black}{Overall the best model is \mantis{} \ceecnet{} V2 with \FracTAL depth $d=5$. Second best is the \mantis{} \FracTAL \texttt{ResNet} with \FracTAL depth $d=10$.} Among the same set of models (\mantis{} \FracTAL {}\texttt{ResNet}), it seems that depth $d=10$ performs best, however we do not know if this generalizes to all models and datasets. We consider that \FracTAL depth $d$ is a hyperparameter that needs to be finetuned for optimal performance, and, as we've shown, the choice $d=5$ is a sensible one as in this particular dataset it provided us with state of the art results.  

\begin{table*}
\footnotesize
\caption{\textcolor{black}{Model comparison on the LEVIR building change detection dataset. We designate with \textbf{bold} font the best values,  with \underline{underline} the second best, and with square brackets, $[\;]$ the third best model. All  of our frameworks (\texttt{D6nf32}) use the \mantis{} macro-topology and achieve state of the art performance. Here \texttt{evo} represents evolving loss strategy, \texttt{sta}, static loss strategy and the depth $d$ refers to the $\ftnmt$ similarity metric of the \FracTAL (attention) layer.}}
\label{LEVIRCD_performance}
\begin{center}
\begin{tabular}{|l |c|c|c|c|c| }
\hline
Model & Precision & Recall & F1 & MCC & IoU \\\hline
\cite{rs12101662} & 83.80 & \textbf{91.00} & 87.30 & - & - \\\hline\hline 
 \ceecnet V1 ($d=5$, \texttt{sta}) & 93.36 & 89.46 & 91.37& 90.94 & 84.10 \\[2pt] 
 \ceecnet V1 ($d=5$, \texttt{evo}) & \underline{93.73} & [89.93] &  [91.79] & [91.38] & [84.82] \\[2pt]
 \ceecnet V2 ($d=5$, \texttt{evo}) & \textbf{93.81} & 89.92 &  \textbf{91.83} & \textbf{91.42} & \textbf{84.89}\\[2pt] 
\hline\hline
\FracTAL \texttt{ResNet} ($d=0$ \texttt{evo}) & 93.50&89.79 &91.61 &91.20 & 84.51\\[2pt]
\FracTAL \texttt{ResNet} \textcolor{black}{($d=5$, \texttt{evo})}    & 93.60 & 89.38  & 91.44  & 91.02 & 84.23 \\[2pt]
\FracTAL \texttt{ResNet} ($d=10$ \texttt{evo}) & [93.63] & \underline{90.04} &\underline{91.80} & \underline{91.39} & \underline{84.84}\\[2pt]\hline
\end{tabular}
\end{center}
\end{table*}

\begin{table*}
\footnotesize
\caption{Model comparison on the WHU building change detection dataset.  
We designate with \textbf{bold} font the best values,  with \underline{underline} the second best, and with square brackets, $[\;]$ the third best model. \cite{rs11111343} presented two models for extracting buildings prior estimating the change mask. These where the Mask-RCNN (in table: \texttt{M1}) and MS-FCN (in table: \texttt{M2}). Our models consume input images of  size of $256\times 256$ pixels. With the exception of \cite{liu2019building} that uses the same size, all other results consume  inputs of size of $512\times 512$ pixels. }
\label{whu_ceecnet_vs_resuneta_comparison}
\begin{center}
\begin{tabular}{|l |c|c|c|c|c| }
\hline
Model & Precision & Recall & F1 & MCC & IoU \\\hline
\cite{rs11111343} \texttt{M1} & 93.100 & 89.200 & [91.108] & - & [83.70] \\ 
\cite{rs11111343} \texttt{M2} & 93.800 & 87.800 & 90.700 & - & 83.00 \\
\cite{chen2020dasnet}   & 89.2 & [90.5] & 89.80 &- & - \\ 
\cite{Cao_2020}   & [94.00] & 79.37 & 86.07 & - & -  \\
\cite{liu2019building}  & 90.15 & 89.35 & 89.75 & - & 81.40
\\\hline\hline 
\FracTAL \texttt{ResNet} \textcolor{black}{($d=5$, \texttt{evo})}     & \underline{95.350} & \underline{90.873} &  \underline{93.058} &  \underline{92.892}  & \underline{87.02}\\[2pt] 
 \ceecnet V1 \textcolor{black}{($d=5$, \texttt{evo})}  &  \textbf{95.571}&  \textbf{92.043} &  \textbf{93.774} & \textbf{93.616}  & \textbf{88.23}\\[2pt]
\hline
\end{tabular}
\end{center}
\end{table*}

\subsection{Performance on WHU}
\label{whu_performance_section}

\textcolor{black}{
In Table \ref{whu_ceecnet_vs_resuneta_comparison} we present the results of training the \mantis{} network with \FracTAL ResNet and \ceecnet V1 building blocks. Both of our proposed architectures outperform all other modeling frameworks, although we need to stress that each of the other authors followed a different splitting strategy of the data. However, with our splitting strategy, we used only the 32.9\% of the total area for training. This is significantly less than the majority of all other methods we report here, and we should anticipate a significant performance degradation in comparison with other methods. In contrast, despite the relatively smaller training set, our method outperforms other approaches.} 
\textcolor{black}{
In particular, 
\cite{rs11111343}, used 50\% of the raster for training, and the other half for testing (Fig. 10 in their manuscript).   
In addition, there is no spatial separation between training and test sites, as it exists in our case, and this should work in their advantage. Also, the usage of a larger window for training (their extracted chips are of spatial dimension $512\times 512$) increases in principle the performance because it includes more context information. There is a tradeoff here though, in that using a larger window size reduces the number of available training chips, therefore the model sees a smaller number of chips during training. 
\cite{chen2020dasnet} split randomly their training and validation chips. This should improve performance, because there is a tight spatial correlation for two extracted chips that are in geospatial proximity.   \cite{Cao_2020} used as a test set $\sim$ 20\% of the total area of the WHU dataset, however, they do not specify the splitting strategy they followed for the training and validation sets. Finally, \cite{liu2019building} used approximately $\sim$ 10\% of the total area for reporting test score performance. They also do not mention their splitting strategy. 
}

\begin{figure}
\centering
\includegraphics[clip, trim=1.5cm 0.040cm 5.25cm .1cm,width=\columnwidth]{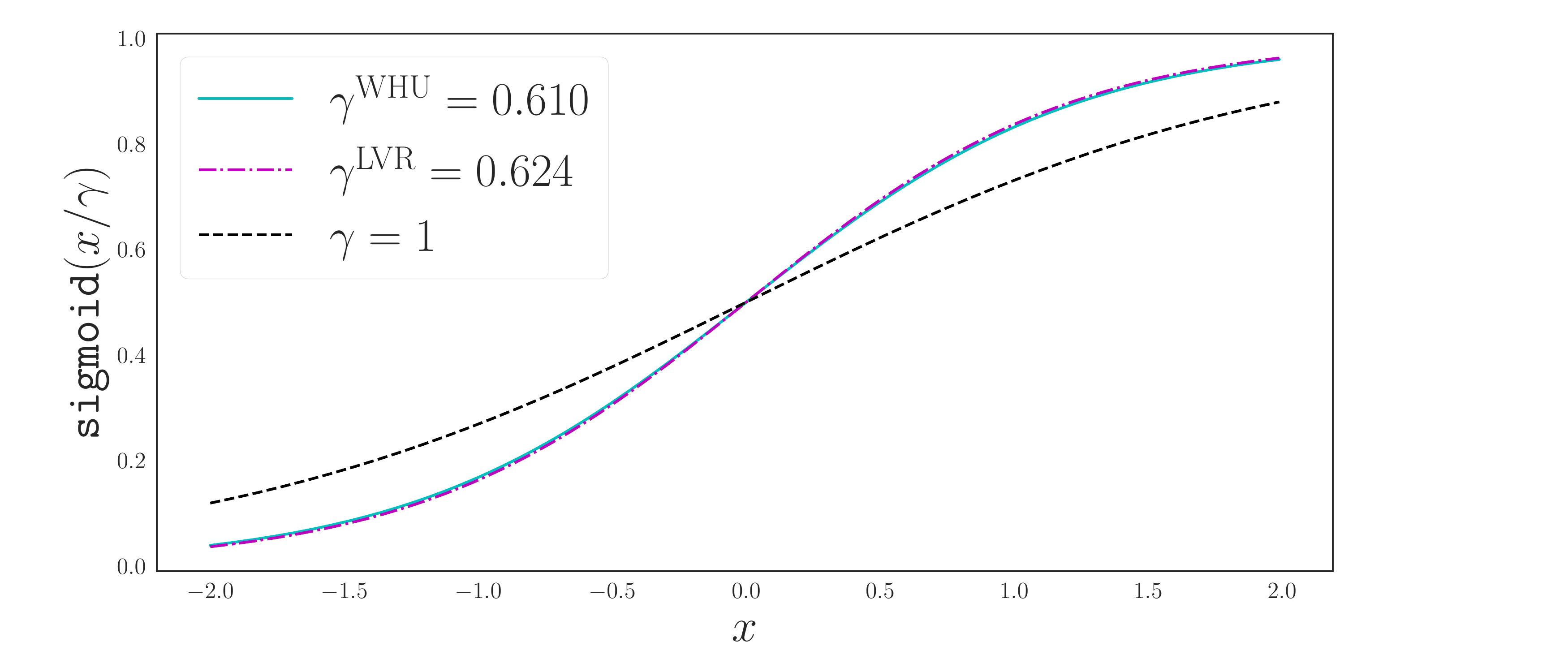}
\caption{Trainable scaling parameters, $\gamma$, for the sigmoid activation, i.e. $\texttt{sigmoid}(x/\gamma)$, that are used in the prediction of change mask boundary layers.} 
\label{ceecnet_crisp_sigmoid}
\end{figure}
In this table we could not include \cite[][PGA-SiamNet]{rs12030484} that report performance results evaluated only on the changed pixels, and not the complete test images. Thus, they are missing out all false positive predictions that can have a dire impact on the performance metrics. They report precision: 97.840, recall: 97.01,  F1: 97.29 and IoU: 97.38.

In Fig. \ref{NZBLDGCD_TEST_AREA_show} we plot from left to right, the test area on date 1, the test area on date 2, the ground truth mask, and the confidence heat map of these predictions. 
In Fig. \ref{NZBLDGCD_show} we plot a set of examples of inference on the WHU dataset. The correspondence of the images in each row is identical to Fig. \ref{LEVIRCD_show}, with the addition that we denote with blue rectangles the locations of changed buildings (true positive predictions), and with red squares missed changes from our model (false negative). It can be seen that  the most difficult areas are the ones that are heavily populated/heavily built up, and the changes are small area buildings.

\begin{figure}
\centering
\includegraphics[clip, trim=.1cm 1.6cm 4.5cm 0.1cm,width=\columnwidth]{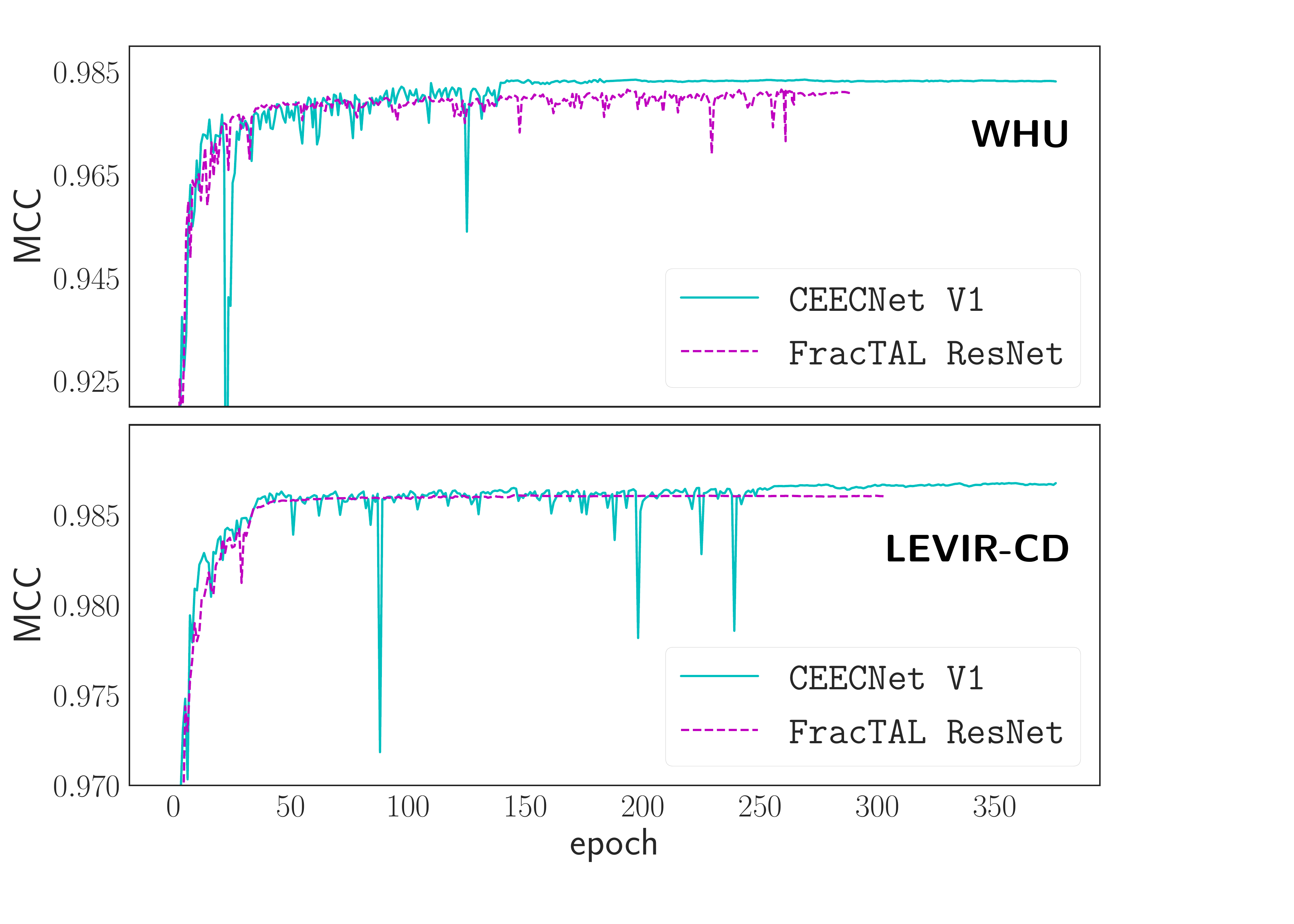}
\caption{\mantis{} \ceecnet{}V1 vs \mantis{} \FracTAL \texttt{ResNet}  \textcolor{black}{(\FracTAL depth, $d=5$)} evolution performance on change detection validation datasets. The top panel corresponds to the LEVIRCD dataset. The bottom panel to the WHU dataset. For each network  we followed the evolving loss strategy: there are two learning rate reductions followed by two scaling ups of the $\langle \ftnmt \rangle^d$ loss function. All four training histories avoid overfitting, thanks to making the loss function sharper towards optimality. } 
\label{ceecnet_vs_fractal_resunet}
\end{figure}

\begin{figure*}
\centering
\includegraphics[clip, trim=7.cm 12.8cm 5.2cm 12.5cm,width=\textwidth]{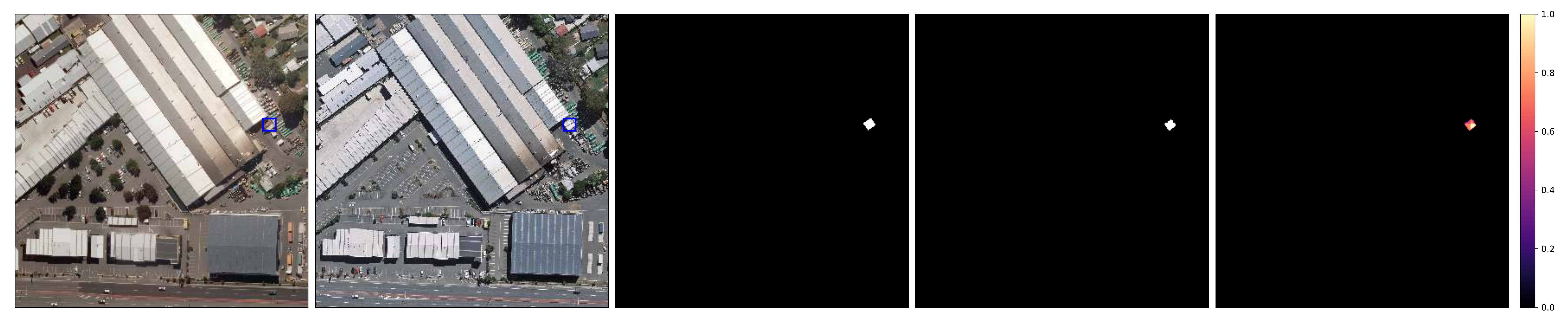}
\includegraphics[clip, trim=7.cm 12.8cm 5.2cm 12.5cm,width=\textwidth]{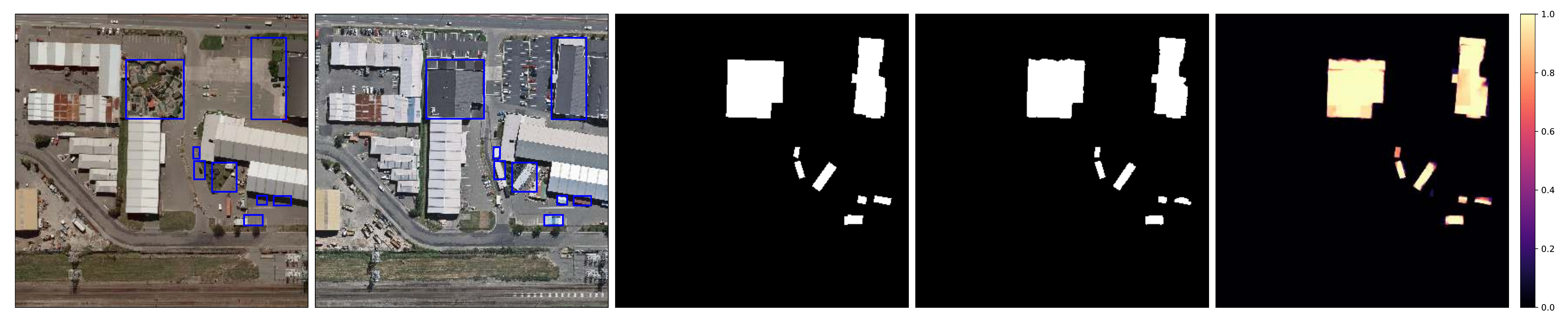}
\includegraphics[clip, trim=7.cm 12.8cm 5.2cm 12.5cm,width=\textwidth]{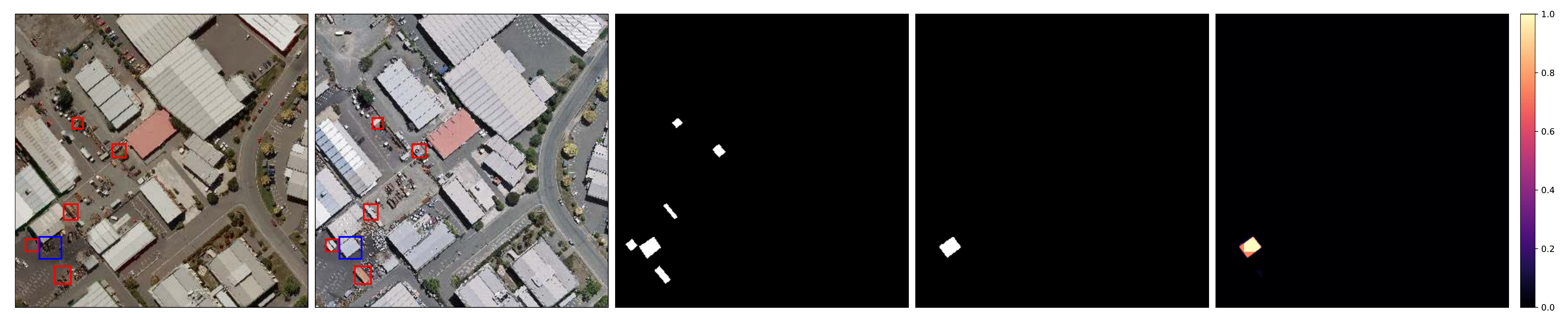}
\includegraphics[clip, trim=7.cm 12.8cm 5.2cm 12.5cm,width=\textwidth]{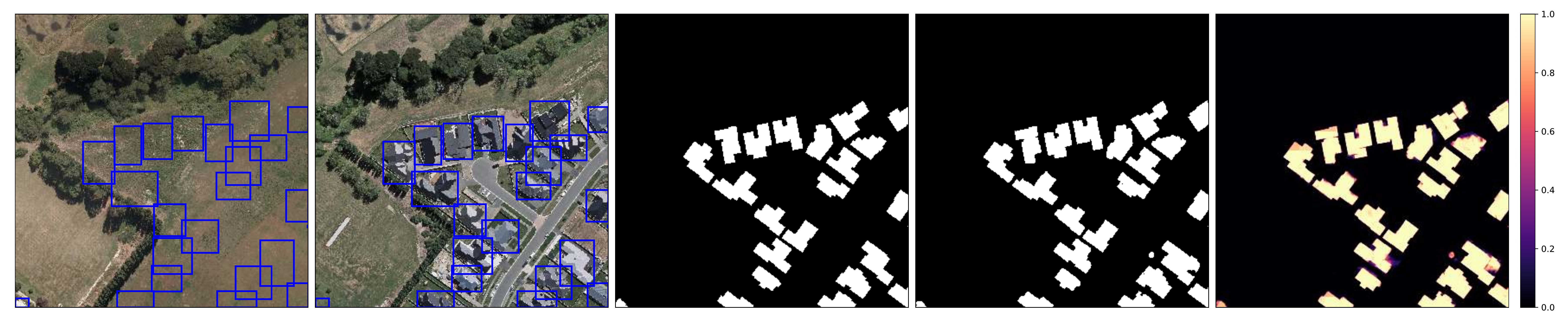}
\includegraphics[clip, trim=7.cm 12.8cm 5.2cm 12.5cm,width=\textwidth]{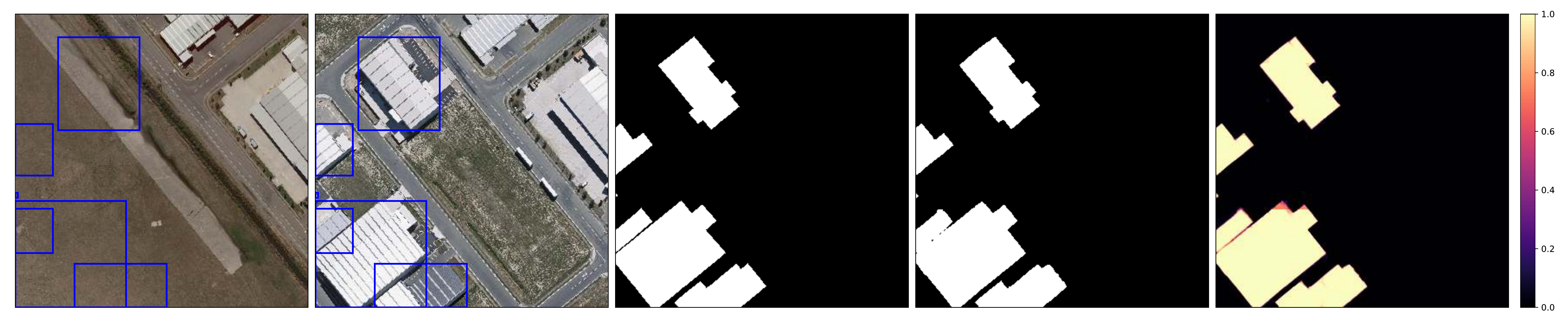}
\includegraphics[clip, trim=7.cm 12.8cm 5.2cm 12.5cm,width=\textwidth]{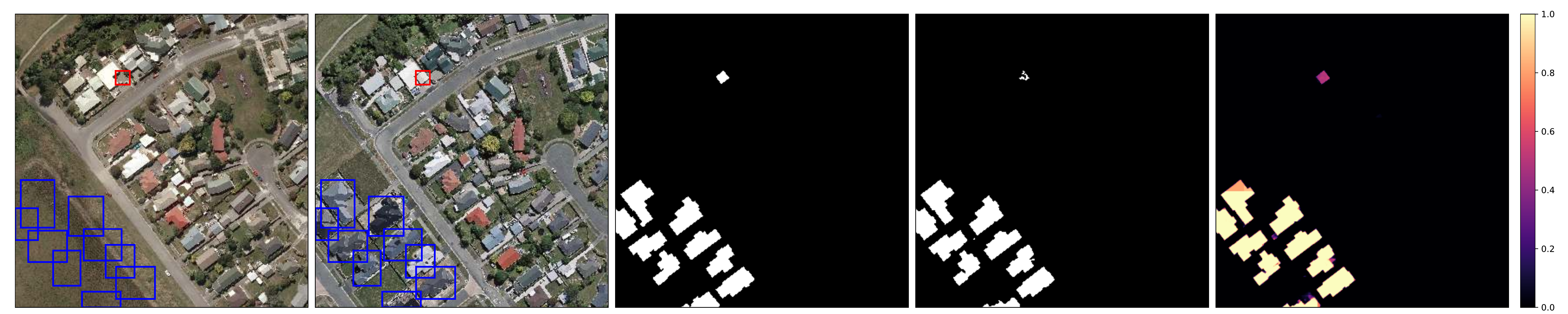}
\caption{Sample of change detection on windows of size $1024\times 1024$ from the WHU dataset. Inference is with the \mantis{} \ceecnet V1  model. The ordering of the inputs, for each row, is as in Fig. \ref{LEVIRCD_show}. We indicate with blue boxes successful findings and with red boxes missed changes on buildings.} 
\label{NZBLDGCD_show}
\end{figure*}

\begin{figure*}
\centering
\includegraphics[clip, trim=4.5cm 6.1cm 3.cm 6.15cm,width=\columnwidth]{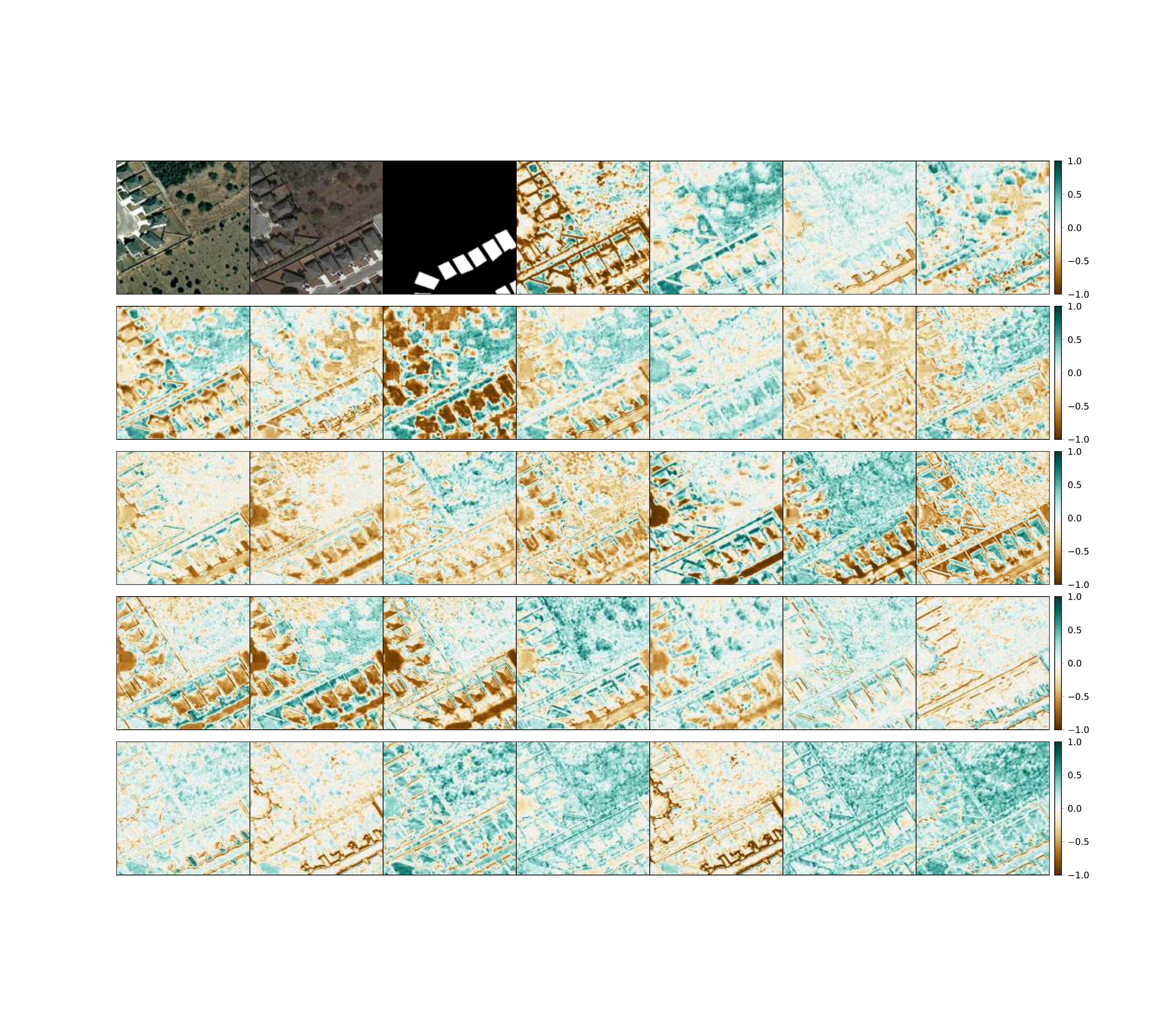}
\includegraphics[clip, trim=4.5cm 6.1cm 3.cm 6.15cm,width=\columnwidth]{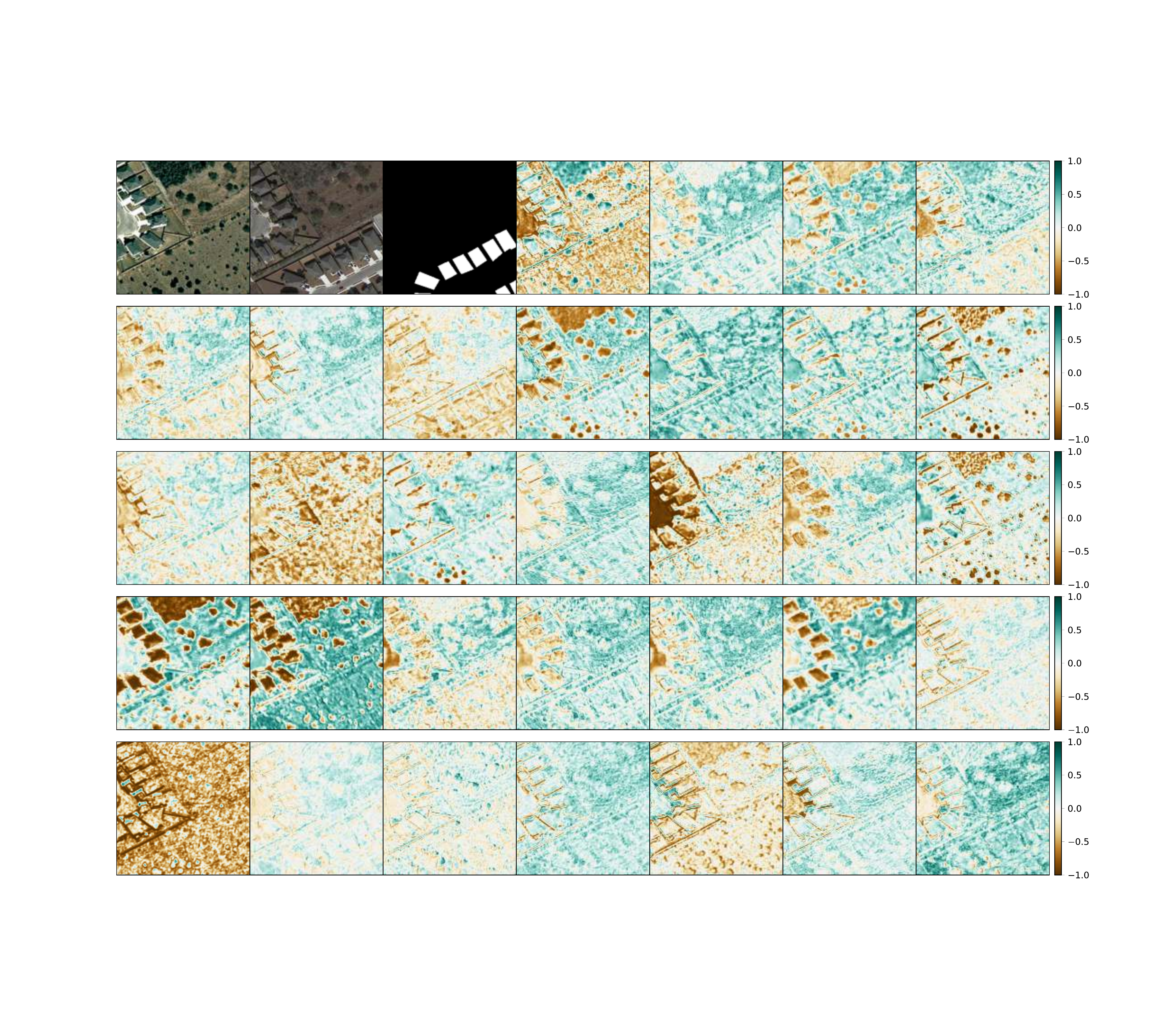}
\caption{\textcolor{black}{Visualization of the relative attention units, \texttt{ratt12} (left pannel) and \texttt{ratt21} (right pannel), for the \texttt{mantis} \FracTAL \texttt{ResNet} with \FracTAL depth, $d=10$. These come from the first feature extractors (channels=32, filter spatial size $256\times 256$). Here, \texttt{ratt12} is the relative attention where for query we use input at date $t_1$, and the key/value filters are created from input at date $t_2$. In the top left rows for each pannel we have input image at date $t_1$, input image at date $t_2$, and ground truth building change labels, followed by the visualization of each of the 32 channels of the features.}} 
\label{FracTAL_d10_ratt12_n_21}
\end{figure*}

\subsection{The effect of scaled sigmoid on the segmentation HEAD}

Starting from an initial value $\gamma = 1$ of the scaled sigmoid boundary layer,  
 the fully trained model \mantis{} \ceecnet V1 learns the following parameters that control how ``crisp'' the boundaries should be, or else, how sharp the decision boundary should be:
\begin{align*}
\gamma_{\texttt{sigmoid}}^{\texttt{LVR}} &= 0.610 \\
\gamma_{\texttt{sigmoid}}^{\texttt{WHU}} &= 0.625 
\end{align*}
The deviation of these coefficients from their initial values, demonstrates that indeed the network finds useful to modify the decision boundary. In Fig. \ref{ceecnet_crisp_sigmoid} we plot the standard sigmoid function ($\gamma=1$) and the sigmoid functions recovered after training on the LEVIRCD and WHU datasets. 

The algorithm in both cases learns to modify the decision boundary, by making it sharper. This means that for two nearby pixels, one belonging to a boundary, the other to a background class, the numerical distance between them needs to be smaller to achieve class separation, in comparison with standard sigmoid. Or else, a small $\delta x$ change is sufficient to transition between boundary and no-boundary class.

\subsection{\textcolor{black}{Qualitative \ceecnet{}  and \FracTAL performance}}
\label{ceecnet_vs_resnet_qualititative}

\begin{figure*}
\centering
\includegraphics[clip, trim=7.cm 12.8cm 5.2cm 12.cm,width=\textwidth]{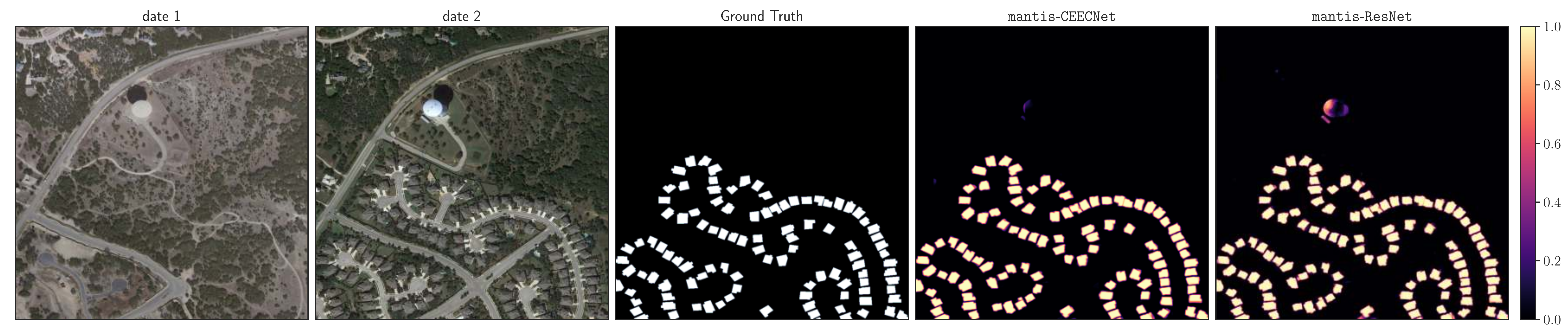}
\includegraphics[clip, trim=7.cm 12.8cm 5.2cm 12.cm,width=\textwidth]{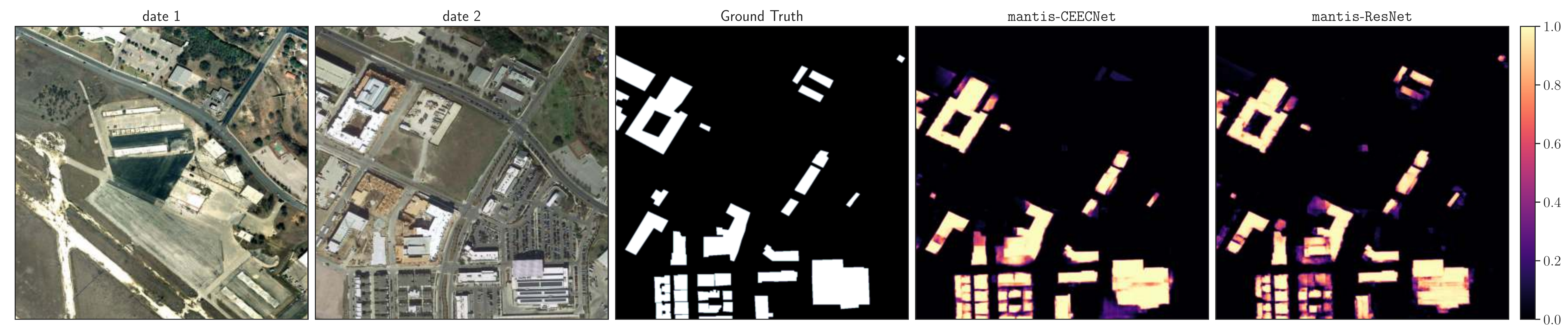}
\includegraphics[clip, trim=7.cm 12.8cm 5.2cm 12.cm,width=\textwidth]{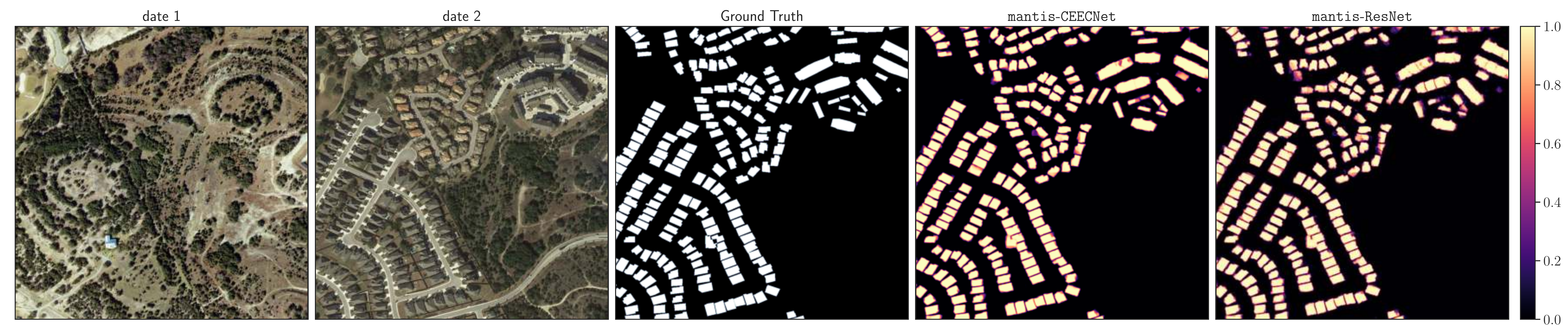}
\includegraphics[clip, trim=7.cm 12.8cm 5.2cm 12.cm,width=\textwidth]{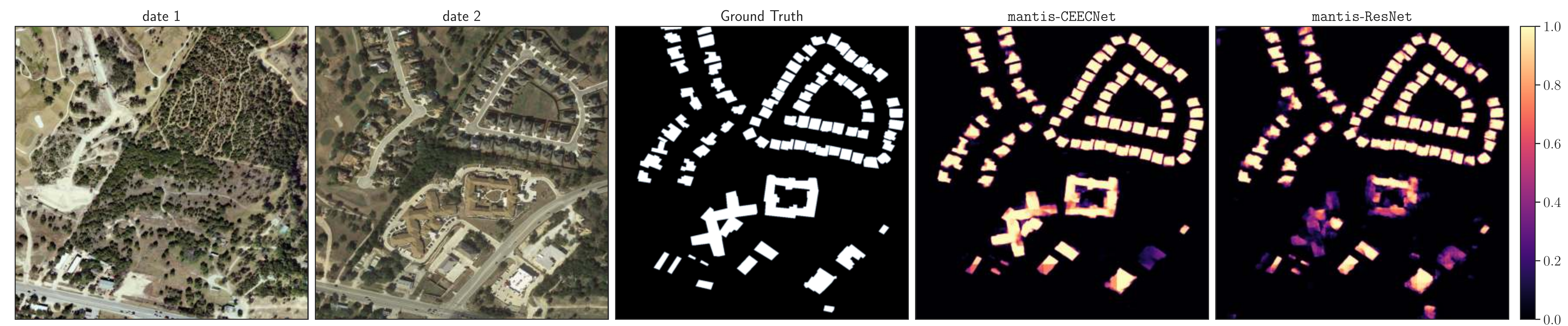}
\includegraphics[clip, trim=7.cm 12.8cm 5.2cm 12.cm,width=\textwidth]{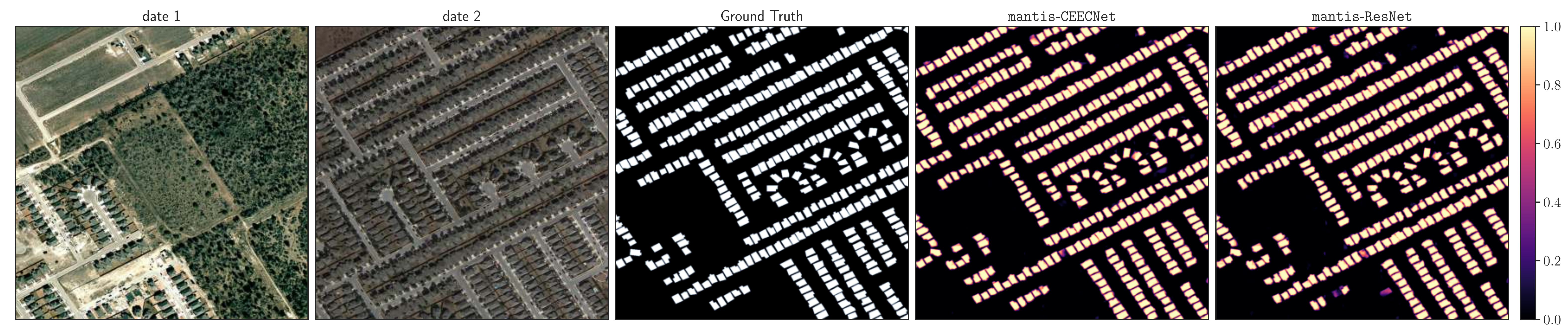}
\caption{Samples of relative quality change detection on test tiles of size $1024\times 1024$ from the LEVIRCD dataset. For each row from left to right: input image date 1, input image date 2, ground truth, confidence heat maps of \mantis{} \ceecnet V1 and \mantis{} \FracTAL \texttt{ResNet} respectively. } 
\label{ceecnet_vs_resnet_lvrcd}
\end{figure*}

\begin{figure*}
\centering
\includegraphics[clip, trim=7.cm 12.8cm 5.2cm 12.cm,width=\textwidth]{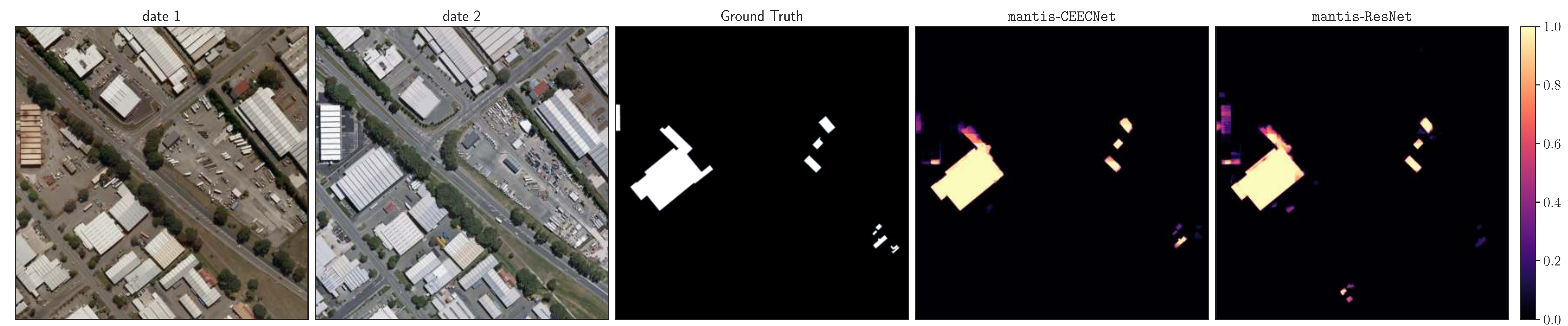}
\includegraphics[clip, trim=7.cm 12.8cm 5.2cm 12.cm,width=\textwidth]{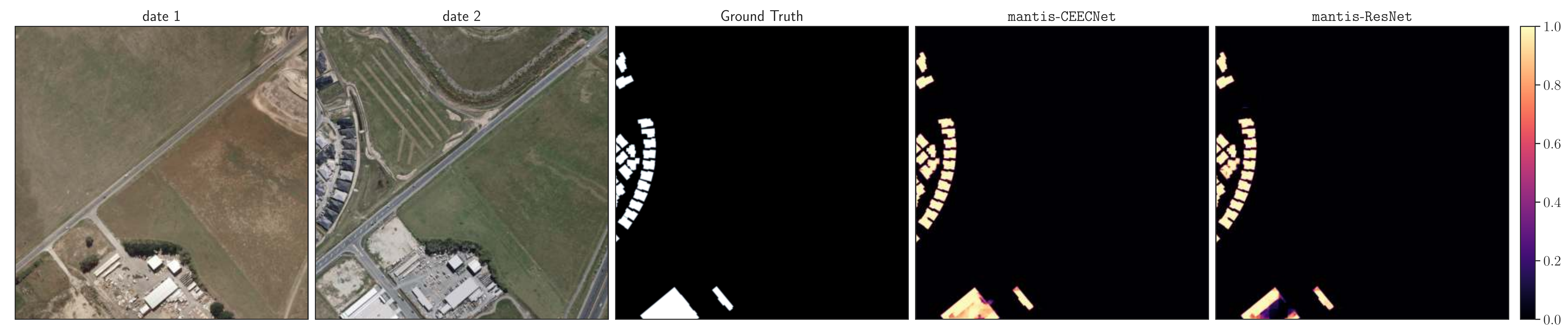}
\includegraphics[clip, trim=7.cm 12.8cm 5.2cm 12.cm,width=\textwidth]{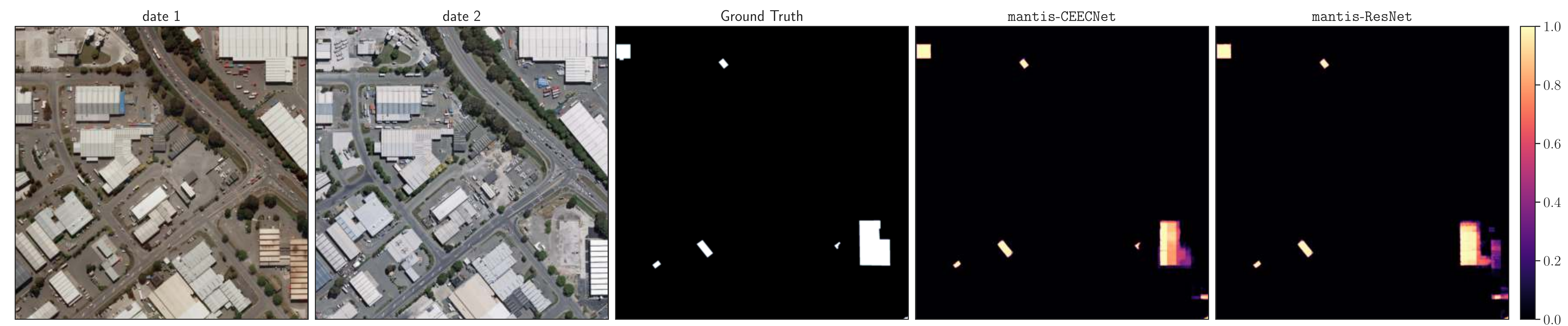}
\includegraphics[clip, trim=7.cm 12.8cm 5.2cm 12.cm,width=\textwidth]{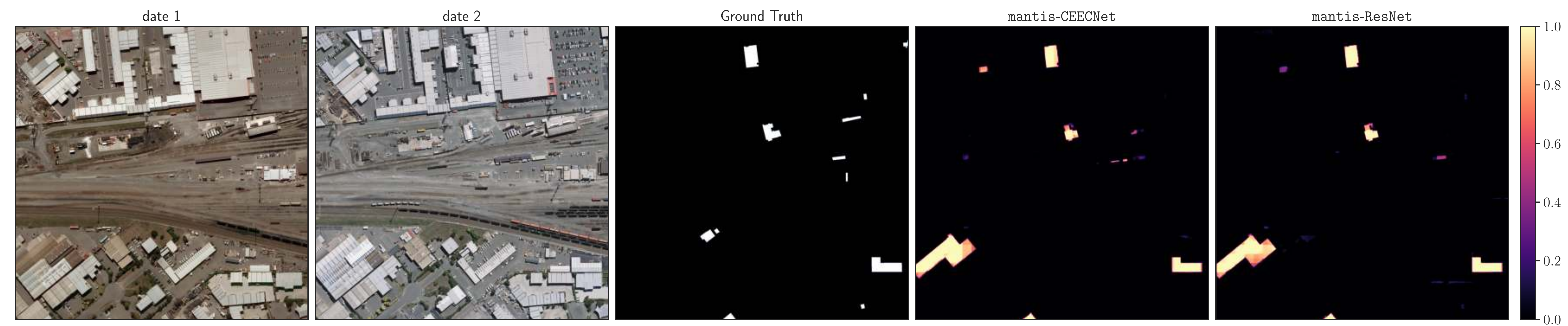}
\includegraphics[clip, trim=7.cm 12.8cm 5.2cm 12.5cm,width=\textwidth]{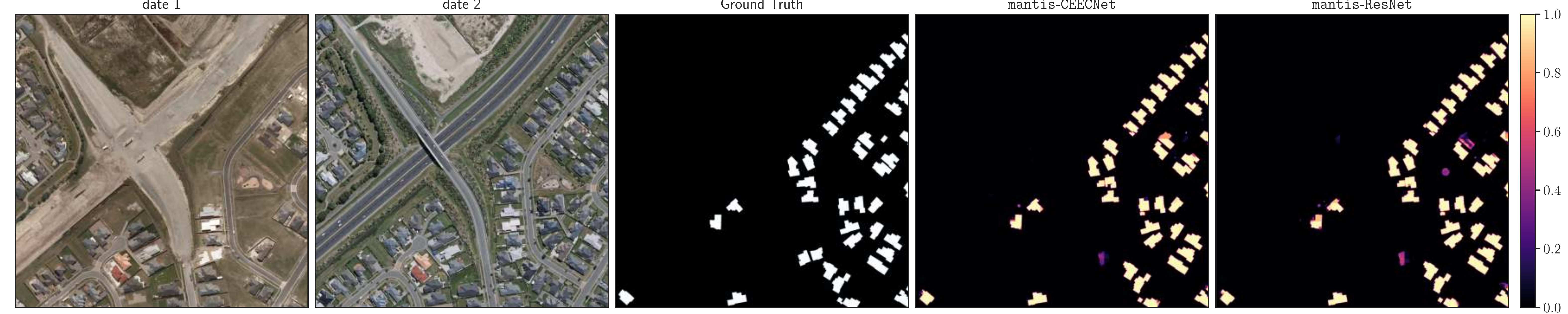}
\includegraphics[clip, trim=7.cm 12.8cm 5.2cm 12.cm,width=\textwidth]{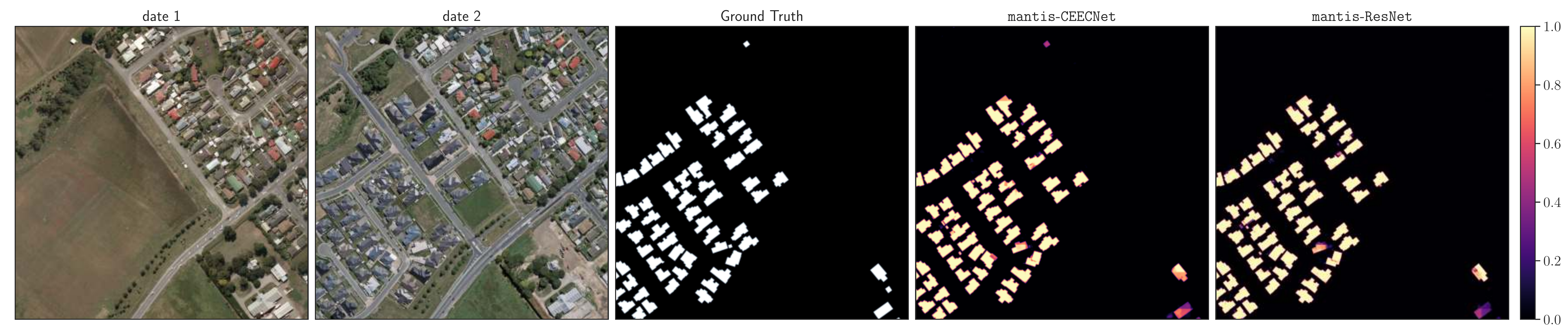}
\caption{As in Fig. \ref{ceecnet_vs_resnet_lvrcd} for sample windows of size $2048\times 2048$ from the WHU dataset.} 
\label{ceecnet_vs_resnet_whu}
\end{figure*}

In this section we base our comparison on \ceecnet{} V1 and \FracTAL \texttt{ResNet} models with \FracTAL depth $d=5$. 
Although both  \ceecnet{} V1  and  \FracTAL \texttt{ResNet} achieve a very high MCC (Fig. \ref{ceecnet_vs_fractal_resunet}), the superiority of \ceecnet{}, \textcolor{black}{for the same \FracTAL depth $d=5$},  is evident in the inference maps in both  the LEVIRCD (Fig. \ref{ceecnet_vs_resnet_lvrcd})  and WHU (Fig. \ref{ceecnet_vs_resnet_whu}) datasets. This confirms their relative scores (Tables \ref{LEVIRCD_performance} and \ref{whu_ceecnet_vs_resuneta_comparison}) and the faster convergence of \ceecnet{} V1 (Fig \ref{CEECNet_vs_ResNet_comparison}). 
 Interestingly, \ceecnet{} V1 predicts change with more confidence than \FracTAL \texttt{ResNet} (Figures \ref{ceecnet_vs_resnet_lvrcd} and \ref{ceecnet_vs_resnet_whu}), even when it errs, as can be seen from the corresponding confidence heat maps. 
\textcolor{black}{The decision on which of the models one should use is a decision to be made with respect to the relative ``cost'' of training each model,  available hardware resources and performance target goal.}

\subsection{\textcolor{black}{Qualitative assesment of the \mantis{} \texttt{macro}-topology}}

A key ingredient of our approach on the task of change detection is that we emphasize on the importance of avoiding using the difference of features to identify change. Instead, we propose the exchange of information between features extracted from images at different dates with the concept of relative attention (section \ref{the_mantis_section}) and fusion (Listing \ref{FusionCODE}). In this section our aim is to get insight on the behaviour of the relative attention and fusion layers, and compare them with the features obtained by the difference of the outputs of convolution layers of images at different dates. We use the outputs of layers of a trained \mantis{} \FracTAL \texttt{ResNet} model, trained on LEVIRCD with \FracTAL depth $d=10$.

\begin{figure*}
\centering
\includegraphics[clip, trim=4.5cm 6.1cm 3.cm 6.15cm,width=\columnwidth]{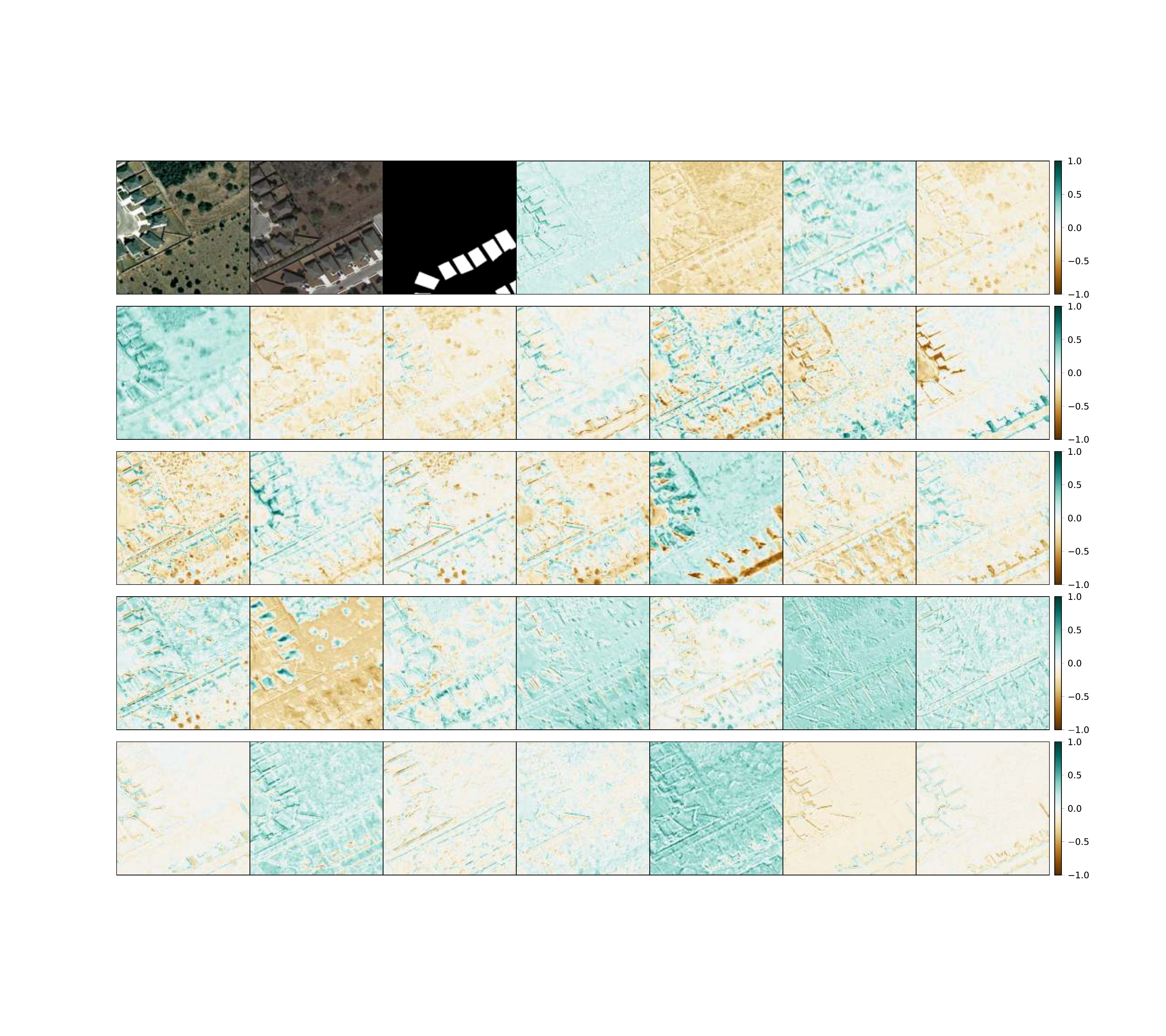}
\includegraphics[clip, trim=4.5cm 6.1cm 3.cm 6.15cm,width=\columnwidth]{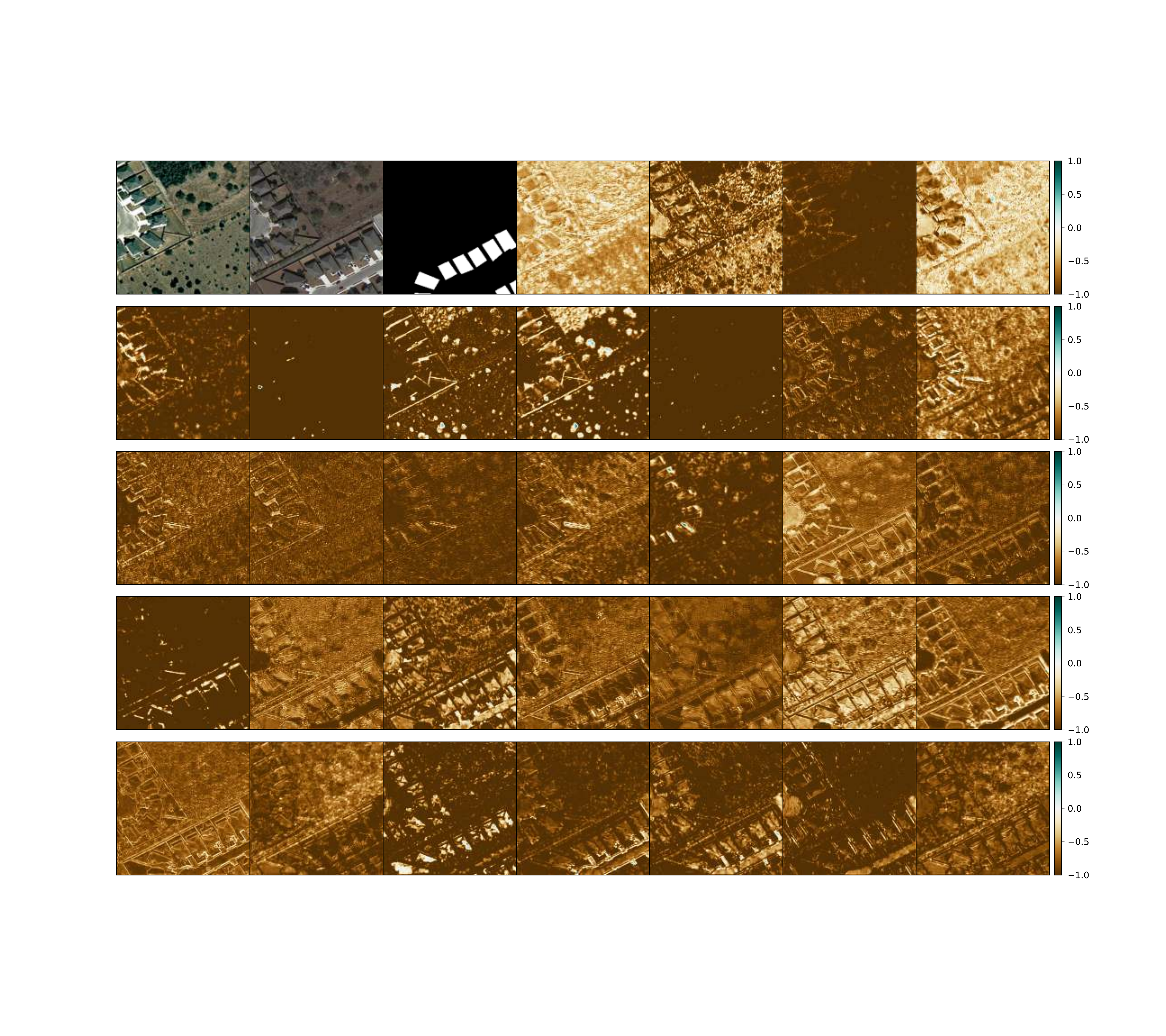}
\caption{\textcolor{black}{For the same model as in Fig. \ref{FracTAL_d10_ratt12_n_21}  we plot the  difference of the first feature extractor blocks (left pannel) vs the first Fusion feature extraction block.  The entropy of the fusion features is half that of the difference channels. This means there is less ``surprise'' in the fusion filters, in comparison with the difference of filters, for the same trained network. }} 
\label{FracTAL_d10_diff_n_fusions}
\end{figure*}

In Fig. \ref{FracTAL_d10_ratt12_n_21}  we visualize the features of the first relative attention layers (channels=32, spatial size $256\times256$, \texttt{ratt12} (left pannel) and \texttt{ratt21} (right pannel) for a set of image patches belonging to the test set (size: $3\times 256\times 256$).
Here, the notation 
\texttt{ratt\textcolor{magenta}{1}\textcolor{cyan}{2}} indicates that the query features come from the input image at date $t_{\textcolor{magenta}{1}}$, while the key/value features are extracted from the input image at date $t_{\textcolor{cyan}{2}}$. Similar notation is applied for the relative attention,  \texttt{ratt\textcolor{cyan}{2}\textcolor{magenta}{1}}.
 Starting from the top left corner we provide the input image at date $t_1$, the input image at date $t_2$ and the ground truth mask of change and after that we visualize the features as single channel images. 
Each feature (i.e. image per channel)  is normalized in the range $[-1,1]$ for visualization purposes. 
It can be seen that the algorithm emphasizes from the early stages (i.e. first layers) to structures containing buildings and boundaries of these. In particular the \texttt{ratt12} (left pannel) has emphasis on boundaries of buildings that exist on both images. It also seems to represent all buildings that exist in both images. The \texttt{ratt21} layer (right pannel) seems to emphasize more the buildings that exist on date 1, but not on date 2. In addition, in both relative attention layers, emphasis is given on roads and pavements.

In Fig. \ref{FracTAL_d10_diff_n_fusions} we visualize the difference of features of the first convolution layers (channels=32, spatial size $256\times256$ - left pannel) and the fused features (right pannel) obtained using the relative attention and fusion methodology (Listing \ref{FusionCODE}). Some key differences between the two is that we observe that there is less variability within channels in the output of the fusion layer, 
in comparison with the difference of features. 
 In order to quantify the information content of the features,  
we calculated the Shanon entropy of the features for each case and we found that the fusion features have half the entropy (11.027) in comparison with the entropy of the difference features (20.97). Similar entropy ratio was found for all images belonging to the test set. This means that the fusion features are less ``surprising'', than the difference features. This may suggest that the fusion provides a better compression of information in comparison with the difference of layers, assuming both layers have the same information content. It may also mean that the fusion layers have less information content than the difference features, i.e. they are harmful for the change detection process. However, if  this was the case, our approach would fail to achieve state of the art performance on the change detection datasets. Therefore, we conclude that the lower entropy value translates to better encoding of information, in comparison with the difference of layers.

\section{Conclusions}

In this work, we propose a new deep learning framework for the task of semantic change detection on very high resolution aerial images, presented here for the case of changes in buildings. This framework is built on top of several novel contributions that can be used independently in computer vision tasks. Our contributions are: 
\begin{enumerate}
\item A novel set similarity coefficient, the fractal Tanimoto coefficient, that is derived from a variant of the Dice coefficient. This coefficient can provide finer detail of similarity, at a desired level (up to a delta function), and this is regulated by a temperature-like hyper-parameter, $d$ (Fig. \ref{FracTanmt2D3D}). 
\item \textcolor{black}{A novel training loss scheme, where we use an evolving loss function, that changes according to learning rate reductions. This helps avoid overfitting and allows for a small increase in performance (Figures \ref{EvolvingLoss_cifar10} \& \ref{cifar10_ft_start_rand}). In particular, this scheme provided $\sim$0.25\% performance increase in validation accuracy on CIFAR10 tests, and  performance increase of $\sim$0.9\%  on IoU and $\sim$0.5\% on MCC on the LEVIRCD dataset.}
\item A novel spatial and channel attention layer, the fractal Tanimoto Attention Layer (\FracTAL - see Listing \ref{FTAttentionCODE}), that uses the fractal Tanimoto similarity coefficient as a means of quantifying the similarity between query and key entries. This layer is memory efficient and scales well with the size of input features. 
\item A novel building block, the \FracTAL \texttt{ResNet} (Fig \ref{ResNetFusion}), that has a small memory footprint and excellent convergent  and performance properties that outperform standard ResNet building blocks. 
\item A novel building block, the Compress/Expand - Expand/Compress (\ceecnet{}) unit (Fig. \ref{ceecnet_unit_v4}), that has better performance than the \FracTAL \texttt{ResNet} (Fig. \ref{ceecnet_vs_fractal_resunet}), that comes, however, at a higher computational cost. 
\item A corollary that follows from the introduced building blocks, is a novel fusion methodology of layers and their corresponding attentions, both for self and relative attention, that improves performance (Fig. \ref{ceecnet_vs_fractal_resunet}). This methodology can be used as a direct replacement of concatenation in convolution neural networks.     
\item A novel macro-topology (backbone) architecture, the \mantis{}      
 topology (Fig. \ref{mantis_t11_architecture}), that combines the building blocks we developed and is able to consume images from two different dates and produce a single change detection layer. It should be noted that the same topology can be used in general segmentation problems, where we have two input images to a network that are somehow correlated and produce a semantic map.   \textcolor{black}{That is, it can be used for fusion of features coming from different inputs (e.g. Digital Surface Maps and RGB images).} 
\end{enumerate}
Putting all things together, all of the proposed networks that presented in this contribution, \mantis{} \FracTAL \texttt{ResNet} and \mantis{} \ceecnet V1\&V2, outperform other proposed networks and achieve state of the art results on the LEVIRCD \citep{rs12101662} and the WHU\footnote{Note that there is not standardized test set for the WHU dataset, therefore relative performance is indicative but not absolute: it depends on the train/test split that other researchers have performed. \textcolor{black}{However, we only used 32.9\% of the area of the provided data for training (which is much less than what other methods we compared against have used) and this demonstrates the robustness and generalisation abilities of our algorithm.}} \citep{Ji2019FullyCN} building change detection datasets (Tables \ref{LEVIRCD_performance} \& \ref{whu_ceecnet_vs_resuneta_comparison}). 
In comparison with state of the art architectures that use atrous dilated convolutions, the proposed architectures do not require fine tuning of the dilation rates. Therefore, they are simpler, and easier to set up and train. 

In this work we did not experiment with deeper architectures, that would surely improve performance (e.g. D7nf32 models usually perform better), or with hyper parameter tuning.

\section*{Acknowledgments}
This project was supported by resources and expertise provided by CSIRO IMT Scientific Computing.   
The authors acknowledge  the support of the \textsc{mxnet} community. The authors would like to thank Pan Chen for careful reading of the manuscript and feedback.   
The authors acknowledge the contribution of the anonymous referees, whos questions helped to improve the quality of the manuscript.

\bibliography{AI_BIB}

\appendix 

\section{CIFAR10 comparison network characteristics}
In Table \ref{ceecnet_vs_resnet} we present in detail the characteristics of the layers that we used to compare on the CIFAR10 dataset. All building blocks use kernel size = 3 and padding = 1 (SAME). 

\begin{table}
\footnotesize
\caption{\ceecnet{}V1 vs \ceecnet{}V2 vs \FracTAL \texttt{ResNet} vs \texttt{ResNet} building blocks comparison. All Building Blocks use kernel size \texttt{k=3} and padding \texttt{p=1} (SAME) and stride \texttt{s=1}. The transition convolutions that half the size of the features use the same kernel size and padding, however the stride  is \texttt{s=2}. In the following we indicate with \texttt{nf} the number of output channels of the convolution layers, and with \texttt{nh} the number of heads in the multihead \FracTAL module.}
\label{ceecnet_vs_resnet}
\begin{center}
\begin{tabular}{|l |c|c| }
\hline
\texttt{Layers} & Proposed Models & \texttt{ResNet} \\\hline\hline
\texttt{Layer} 1 & \texttt{BBlock}[\texttt{nf=64,nh=8}]  & \texttt{BBlock}[\texttt{nf=64}]  \\\hline
\texttt{Layer} 2 & \texttt{BBlock}[\texttt{nf=64,nh=8} ]   & \texttt{BBlock}[\texttt{nf=64}] \\\hline
\texttt{Layer} 3 & \texttt{Conv2DN(nf=128,s=2)} & \texttt{Conv2DN(nf=128,s=2)} \\\hline
\texttt{Layer} 4 & \texttt{BBlock}[\texttt{nf=128,nh=16}]  & \texttt{BBlock}[\texttt{nf=128}] \\\hline
\texttt{Layer} 5 & \texttt{BBlock}[\texttt{nf=128,nh=16}]  & \texttt{BBlock}[\texttt{nf=128}]\\\hline
\texttt{Layer} 6 & \texttt{Conv2DN(nf=256,s=2)} & \texttt{Conv2DN(nf=256,s=2)} \\\hline
\texttt{Layer} 7 &\texttt{BBlock}[ \texttt{nf=256,nh=32}] & \texttt{BBlock}[\texttt{nf=256}]\\\hline
\texttt{Layer} 8 &\texttt{BBlock}[ \texttt{nf=256,nh=32}] &\texttt{BBlock}[\texttt{nf=256}]\\\hline
\texttt{Layer} 9 & \texttt{ReLU}  & \texttt{ReLU} \\\hline
\texttt{Layer} 10 & \texttt{DenseN(nf=4096)} & \texttt{DenseN(nf=4096)}\\\hline
\texttt{Layer} 11 & \texttt{ReLU}  & \texttt{ReLU} \\\hline
\texttt{Layer} 12 & \texttt{DenseN(nf=512)}  & \texttt{DenseN(nf=512)}\\\hline
\texttt{Layer} 13 & \texttt{ReLU}  & \texttt{ReLU} \\\hline
\texttt{Layer} 14 & \texttt{DenseN(nf=10)}  & \texttt{DenseN(nf=10)}\\\hline
\end{tabular}
\end{center}
\end{table}

\section{Inference across WHU test set}

\begin{figure*}
\centering
\includegraphics[clip, trim=7.cm 12.8cm 5.2cm 13.1cm,width=\textwidth]{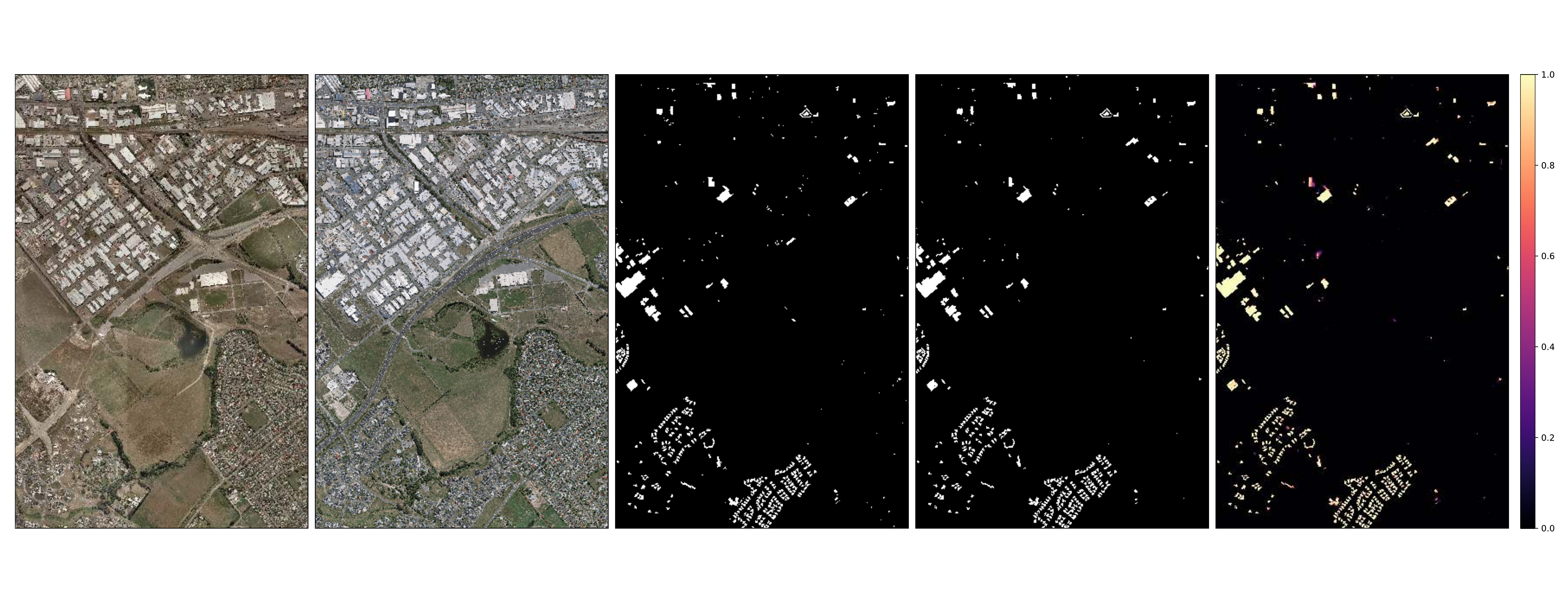}
\caption{Inference across the whole test area over NZBLDG CD Dataset using the \mantis{}\ceecnet V1 D6nf32 model. From left to right: 2011 input image, 2016 input image, ground truth, prediction (threshold 0.5) and confidence heat map.} 
\label{NZBLDGCD_TEST_AREA_show}
\end{figure*}

The inference for the best performing model, the \mantis{} \ceecnet V1 D6nf32 model can be seen on Fig. \ref{NZBLDGCD_TEST_AREA_show}. The predictions match very closely the ground truth. 

\section{Algorithms}
\label{section_algorithms}
Here we present with \textsc{mxnet} style pseudocode the implementation of the \FracTAL associated modules. In all the listings presented, \texttt{Conv2DN} is a sequential combination of a 2D convolution followed by a normalization layer. When the batch size is very small, due to GPU memory normalization (e.g. smaller than 4 datums per GPU), the normalization used was Group Normalization \cite{DBLP:journals/corr/abs-1803-08494}. Practically, in all \mantis{} \ceecnet{} realizations for change detection, we used GroupNorm. 

\subsection{Fractal Tanimoto Attention 2D module}

\begin{python}[emphstyle=\textcolor{magenta}, caption={\textsc{mxnet/gluon} style pseudo code for the fractal Tanimoto coefficient, predefined for spatial similarity.}, emph={FTanimoto,forward,__init__,inner_prod},label={FTanimotoCODE}]
from mxnet.gluon import nn
class FTanimoto(nn.Block):
    def __init__(self,depth=5, axis=[2,3],**kwards):
        super().__init__(**kwards)
        self.depth = depth
        self.axis=axis

    def inner_prod(self, prob, label):
        prdct = prob*label #dim:(B,C,H,W)
        prdct = prdct.sum(axis=self.axis,keepdims=True)
        return prdct #dim:(B,C,1,1)

    def forward(self, prob, label):
        a = 2.**self.depth
        b = -(2.*a-1.)

        tpl= self.inner_prod(prob,label)
        tpp= self.inner_prod(prob,prob)
        tll= self.inner_prod(label,label)
        
        denum = a*(tpp+tll)+b*tpl
        ftnmt = tpl/denum
        return ftnmt #dim:(B,C,1,1)
\end{python}

\begin{python}[caption={\textsc{mxnet/gluon} style pseudocode for the fractal Tanimoto Attention module},
emph={FTAttention2D,forward,__init__,inner_prod},emphstyle=\textcolor{magenta},label={FTAttentionCODE}]
from mxnet import nd as F
from mxnet.gluon import nn
class FTAttention2D(nn.Block):
    def __init__(self, nchannels, nheads, **kwards):
            super().__init__(**kwards)
        self.q = Conv2DN(nchannels,groups=nheads)
        self.k = Conv2DN(nchannels,groups=nheads)
        self.v = Conv2DN(nchannels,groups=nheads)
        # spatial/channel similarity
        self.SpatialSim = FTanimoto(axis=[2,3])
        self.ChannelSim = FTanimoto(axis=1)
        self.norm = nn.BatchNorm()

    def forward(self, qin, kin, vin):
        # query, key, value
        q = F.sigmoid(self.q(qin))#dim:(B,C,H,W)
        k = F.sigmoid(self.k(vin))#dim:(B,C,H,W) 
        v = F.sigmoid(self.v(kin))#dim:(B,C,H,W)

        att_spat = self.ChannelSim(q,k)#dim:(B,1,H,W)
        v_spat  =  att_spat*v #dim:(B,C,H,W)

        att_chan = self.SpatialSim(q,k)#dim:(B,C,1,1)
        v_chan   = att_chan*v #dim:(B,C,H,W)

        v_cspat =  0.5*(v_chan+v_spat) 
        v_cspat = self.norm(v_cspat)

        return v_cspat #dim:(B,C,H,W) 
\end{python}

\begin{python}[caption={\textsc{mxnet/gluon} style pseudocode for the Relative Attention Fusion module},
emph={Fusion,forward,__init__},emphstyle=\textcolor{magenta},label={FusionCODE}]
import mxnet as mx
from mxnet import nd as F
class Fusion(nn.Block):
    def __init__(self, nchannels, nheads, **kwards):
        super().__init__(**kwards)
        self.fuse = Conv2DN(nchannels,
                          kernel=3,
                          padding=1,
                          groups=nheads)
        self.att12 = FTAttention2D(nchannels,nheads)
        self.att21 = FTAttention2D(nchannels,nheads)

        self.gamma1  = self.params.get('gamma1',
                      shape=(1,),
                      init=mx.init.Zero())
        self.gamma2  = self.params.get('gamma2',
                      shape=(1,),
                      init=mx.init.Zero())

    def forward(self, input1,input2):
        ones = nd.ones_like(input1)          
    
        # Attention on 1, for k,v from 2        
        qin = input1
        kin = input2
        vin = input2
        att12 = self.att12(qin,kin,vin)
        out12 = input1*(ones+self.gamma1*att12)
        
        # Attention on 2, for k,v from 1
        qin = input2
        kin = input1
        vin = input1
        att21 = self.att21(qin,kin,vin)
    	out21 = input2*(ones+self.gamma2*att21)

    	out = nd.concat(out12,out21,dim=1)
    	out = self.fuse(out)    	    
        return out
\end{python}

\subsection{\FracTAL ResNet}

In this Listing, the \texttt{ResBlock} consists of the sequence of \texttt{BatchNorm}, \texttt{ReLU}, \texttt{Conv2D}, \texttt{BatchNorm}, \texttt{ReLU}, \texttt{Conv2D}. The normalization can change to \texttt{GroupNorm} for a small batch size. 

\begin{python}[caption={\textsc{mxnet/gluon} style pseudocode for the Residual Attention Fusion module},
emph={FTAttResUnit,forward,__init__},emphstyle=\textcolor{magenta},label={FTAttResUnitCODE}]
import mxnet as mx
from mxnet import nd as F
class FTAttResUnit(nn.Block):
    def __init__(self, nchannels, nheads, **kwards):
        super().__init__(**kwards)
        # Residual Block: sequence of
        # (BN,ReLU,Conv,BN,ReLU,Conv)
        self.ResBlock = ResBlock(nchannels,
                          kernel=3,
                          padding=1)
        self.att = FTAttention2D(nchannels,nheads)

        self.gamma  = self.params.get('gamma',
                      shape=(1,),
                      init=mx.init.Zero())

    def forward(self, input):
        out = self.ResBlock(input)#dim:(B,C,H,W)        
        qin = input
        vin = input
        kin = input
        att = self.attention(qin,vin,kin)#dim:(B,C,H,W)
        att = self.gamma * att
        out = (input + out)*(F.ones_like(out)+att)
        return out
\end{python}

\subsection{\ceecnet{} building blocks}

In this section we provide with pseudo-code the implementation of the \ceecnet V1 unit. 
\begin{python}[caption={\textsc{mxnet/gluon} style pseudocode for the \ceecnet V1 unit.},
emph={CEECNet_unit_V1,forward,__init__},emphstyle=\textcolor{magenta},label={CEECNetUnitCODE}]
import mxnet as mx
from mxnet import nd as F
class CEECNet_unit_V1(nn.Block):
    def __init__(self, nchannels, nheads, **kwards):
        super().__init__(**kwards)
        # Compress-Expand
        self.conv1= Conv2DN(nchannels/2)
        self.compr11= Conv2DN(nchannels,k=3,p=1,s=2)
        self.compr12= Conv2DN(nchannels,k=3,p=1,s=1)
        self.expand1= ExpandNComb(nchannels/2)
        
        # Expand Compress
        self.conv2= Conv2DN(nchannels/2)
        self.expand2= Expand(nchannels/4)
        self.compr21= Conv2DN(nchannels/2,k=3,p=1,s=2)
        self.compr22= Conv2DN(nchannels/2,k=3,p=1,s=1)
        
        self.collect= Conv2DN(nchannels,k=3,p=1,s=1)
        
        self.att= FTAttention2D(nchannels,nheads)
        self.ratt12= RelFTAttention2D(nchannels,nheads)
        self.ratt21= RelFTAttention2D(nchannels,nheads)


        self.gamma1  = self.params.get('gamma1',
                      shape=(1,),
                      init=mx.init.Zero())
        self.gamma2  = self.params.get('gamma2',
                      shape=(1,),
                      init=mx.init.Zero())
        self.gamma3  = self.params.get('gamma3',
                      shape=(1,),
                      init=mx.init.Zero())

    def forward(self, input):
        # Compress-Expand
    	out10 = self.conv1(input)
    	out1  = self.compr11(out10)
    	out1  = F.relu(out1)
    	out1  = self.compr12(out1)
    	out1  = F.relu(out1)
    	out1  = self.expand1(out1,out10)
    	out1  = F.relu(out1)
    
    	# Expand-Compress  
    	out20 = self.conv2(input)
    	out2  = self.expand2(out20)
    	out2  = F.relu(out2)
    	out2  = self.compr21(out2)
    	out2  = F.relu(out2)
    	out2  = F.concat([out2,out20],axis=1)
    	out2  = self.compr22(out2)    	
    	out2  = F.relu(out2)

        # attention    
    	att   = self.gamma1*self.att(input)
    	
    	# relative attention 122
    	qin   = out1
        kin   = out2
        vin   = out2
        ratt12 = self.gamma2*self.ratt12(qin,kin,vin)

        # relative attention 211
        qin   = out2
        kin   = out1
        vin   = out1	
        ratt21 = self.gamma3*self.ratt21(qin,kin,vin)      
    
    	ones1 = F.ones_like(out10)# nchannels/2
    
    	out122= out1*(ones1+ratt12)
    	out211= out2*(ones1+ratt21)
        out12 = F.concat([out122,out211],dim=1)
        out12 = self.collect(out12)
        out12 = F.relu(out12)
		
        # Final fusion
        ones2 = F.ones_like(input)
        out   = (input+out12)*(ones2+att)
		
        return out 
\end{python}

The layers \texttt{Expand} and \texttt{ExpandNCombine} are defined through Listings \ref{ExpandCODE} and \ref{ExpandNCombineCODE}. 

\begin{python}[caption={\textsc{mxnet/gluon} style pseudocode for the \texttt{Expand} layer used in the \ceecnet V1 unit.},
emph={Expand,forward,__init__},emphstyle=\textcolor{magenta},label={ExpandCODE}]
import mxnet as mx
from mxnet import nd as F
class Expand(nn.Block):
    def __init__(self, nchannels, nheads, **kwards):
        super().__init__(**kwards)
        self.conv1 = Conv2DN(nchannels,k=3, p=1,
                          groups=nheads)
        self.conv2 = Conv2DN(nchannels,k=3, p=1,
                          groups=nheads)

    def forward(self, input):
       
        out = F.BilinearResize2D(input,
                 scale_height=2,
                 scale_width=2)    
        out = self.conv1(out)
        out = F.relu(out)
        out = self.conv2(out)
        out = F.relu(out)
        return out 
\end{python}

\begin{python}[caption={\textsc{mxnet/gluon} style pseudocode for the \texttt{ExpandNCombine} layer used in the \ceecnet V1 unit.},
emph={ExpandNCombine,forward,__init__},emphstyle=\textcolor{magenta},label={ExpandNCombineCODE}]
import mxnet as mx
from mxnet import nd as F
class ExpandNCombine(nn.Block):
    def __init__(self, nchannels, nheads, **kwards):
        super().__init__(**kwards)
        self.conv1 = Conv2DN(nchannels,k=3, p=1,
                          groups=nheads)
        self.conv2 = Conv2DN(nchannels,k=3, p=1,
                          groups=nheads)

    def forward(self, input1,input2):
        # input1 has lower spatial dimensions
        out1 = F.BilinearResize2D(input1,
                 scale_height=2,
                 scale_width=2)    
        out1 = self.conv1(out1)
        out1 = F.relu(out1)
        
        out2= F.concat([out1,input2],dim=1)
        out2 = self.conv2(out2)
        out2 = F.relu(out2)
        return out2 
\end{python}

\section{Software implementation and training characteristics}
\label{resuneta_babysitting}
The networks 
\mantis{} \ceecnet{} and \FracTAL \texttt{ResNet}   were built and trained using the \textsc{mxnet} deep learning library \citep{chen2015mxnet}, under the \textsc{GLUON} API. Each of the models  was trained with a batch size of $\sim$256 on 16 nodes containing 4 NVIDIA Tesla P100 GPUs each in CSIRO HPC facilities. Due to the complexity of the network, the batch size in a single GPU iteration cannot be made larger than $\sim$4 (per GPU). The models were trained in a  distributed scheme, using the ring 
allreduce algorithm, and in particular it's implementation on \textsc{Horovod} \citep{sergeev2018horovod} for the \textsc{mxnet} \citep{chen2015mxnet} deep learning library. 
For all models , we used the Adam \citep{DBLP:journals/corr/KingmaB14} optimizer, with  momentum parameters $(\beta_1,\beta_2)=(0.9,0.999)$.  
The learning rate was reduced by an order of magnitude  whenever the validation loss stopped decreasing.  Overall we reduced the learning rate 3 times. The depth of the evolving loss function was increased every time the learning rate was reduced. The depths of the $\langle\ftnmt \rangle^d$ that we used were $d \in \{0,10,20,30\}$.

\end{document}